\documentclass[10pt,twocolumn,letterpaper]{article}

\usepackage[pagenumbers]{cvpr} 

\usepackage{graphicx}
\usepackage{amsmath}
\usepackage{amssymb}
\usepackage{booktabs}

\usepackage[pagebackref,breaklinks,colorlinks]{hyperref}

\usepackage[utf8]{inputenc} 
\usepackage[T1]{fontenc}    
\usepackage{url}            
\usepackage{booktabs}       
\usepackage{amsfonts}       
\usepackage{nicefrac}       
\usepackage{microtype}      
\usepackage{xcolor}         
\usepackage{multicol}
\usepackage{multirow}
\usepackage{color}
\usepackage{amsmath}
\usepackage{epstopdf}
\usepackage{amsfonts}
\usepackage{amssymb}
\usepackage{bbding}
\usepackage{bbm}
\usepackage{bm}
\usepackage{colortbl}
\usepackage{makecell}
\usepackage{enumitem}

\usepackage{graphicx}
\usepackage{amsmath}
\usepackage{amssymb}
\usepackage{booktabs}
\usepackage{times}
\usepackage{epsfig}
\usepackage{graphicx}
\usepackage{amsmath}
\usepackage{amssymb}
\usepackage{color,soul}
\usepackage{graphicx}
\usepackage{caption}
\usepackage{subcaption}
\usepackage{multirow}
\usepackage{makecell}
\usepackage[ruled,vlined]{algorithm2e}
\usepackage{silence}
\usepackage{multirow}
\usepackage{float}
\usepackage{makecell}
\usepackage{bbm}
\usepackage{amsmath, amssymb, mathtools}

\definecolor{lightblue}{rgb}{0.9255,0.9569,   1.0000}
\definecolor{revision}{rgb}{0,0,0}
\definecolor{revision2}{rgb}{0,0,0}
\definecolor{todo}{rgb}{1,0,0}

\usepackage[capitalize]{cleveref}
\crefname{section}{Sec.}{Secs.}
\Crefname{section}{Section}{Sections}
\Crefname{table}{Table}{Tables}
\crefname{table}{Tab.}{Tabs.}

\makeatletter
\newcommand{\printfnsymbol}[1]{%
  \textsuperscript{\@fnsymbol{#1}}%
}
\newcommand{\printfnsymbolnew}[2]{%
  \textsuperscript{\@fnsymbol{#1}}%
}

\def\cvprPaperID{7170} 
\def\confName{CVPR}
\def\confYear{2023}

\begin{document}

\title{SeasonDepth: Cross-Season Monocular Depth Prediction Dataset and Benchmark under Multiple Environments}

\author{Hanjiang Hu$^{1}$\thanks{equal contribution},  ~Baoquan Yang$^{2}$\printfnsymbol{1}, Zhijian Qiao$^{3}$\printfnsymbol{1}, Shiqi Liu$^{1}$\\ Jiacheng Zhu$^{1}$, Zuxin Liu$^{1}$, Wenhao Ding$^{1}$, Ding Zhao$^{1}$, Hesheng Wang$^{2}$\thanks{corresponding author}\\
\small $^1$Carnegie Mellon University,    $^2$Shanghai Jiao Tong University,  $^3$Hong Kong University of Science and Technology
\\ {\tt\small hanjianghu@cmu.edu, wanghesheng@sjtu.edu.cn}}
\maketitle

\begin{abstract}
Different environments pose a great challenge to the outdoor robust visual perception for long-term autonomous driving, and the generalization of learning-based algorithms on different environments is still an open problem. Although monocular depth prediction has been well studied recently, few works focus on the robustness of learning-based depth prediction across different environments, e.g. changing illumination and seasons, owing to the lack of such a multi-environment real-world dataset and benchmark. To this end, the first cross-season monocular depth prediction dataset and benchmark, SeasonDepth, is introduced to benchmark the depth estimation performance under different environments. We investigate several state-of-the-art representative open-source supervised and self-supervised depth prediction methods using newly-formulated metrics. Through extensive experimental evaluation on the proposed dataset and cross-dataset evaluation with current autonomous driving datasets, the performance and robustness against the influence of multiple environments are analyzed qualitatively and quantitatively. We show that long-term monocular depth prediction is still challenging and believe our work can boost further research on the long-term robustness and generalization for outdoor visual perception. 
The dataset is available on \url{https://seasondepth.github.io}, and benchmark toolkit is available on \url{https://github.com/SeasonDepth/SeasonDepth}.
\end{abstract}

\section{Introduction}
\label{sec:intro}
Perception and localization for autonomous driving and mobile robotics have made significant progress due to the boost of deep  neural networks  \cite{eigen2014depth,liu2015learning,laina2016deeper, xu2022opv2v, xu2022v2x} in recent years. However, since the outdoor environmental conditions are changing because of different seasons, weather and daytime  \cite{maddern20171,sattler2018benchmarking,liu2019lpd}, the pixel-level appearance is drastically affected, which casts a big challenge for robust long-term visual perception and localization. 
\textcolor{revision}{Monocular depth prediction plays a critical role in long-term visual perception and localization \cite{zhou2021patch2pix,larsson2019fine,jenicek2019no,hu2020dasgil,piasco2021improving,xu2022cobevt} and is also significant to  safe applications such as self-driving cars under different environmental conditions.}
 Although some depth prediction datasets  \cite{cordts2016cityscapes,ranftl2020towards,antequera2020mapillary} include different environments for diversity, it is still not clear what kind of algorithm is more robust to adverse conditions and how they influence depth prediction performance. \textcolor{revision}{Besides, the generalization of learning-based depth prediction methods on different weather and illumination effects is still an open problem.
 Therefore, building a new dataset and benchmark under multiple environments is needed to  study this problem systematically. To the best of our knowledge, we are the first to study the generalization of learning-based depth prediction under changing environments, which is essential and significant to both robust machine learning algorithms and practical applications like autonomous driving. }
 
\begin{figure*}[]
	\centering
	\centering
	\includegraphics[width=0.19\linewidth]{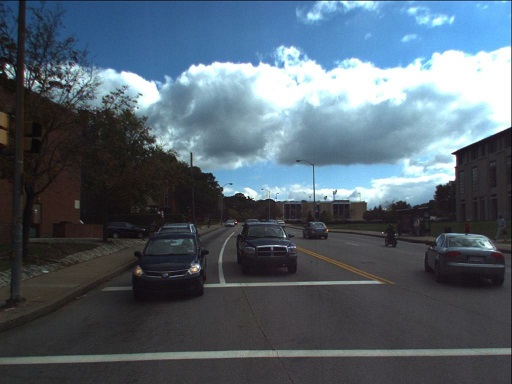} \includegraphics[width=0.19\linewidth]{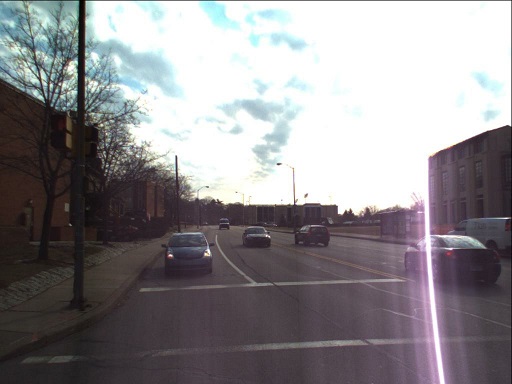}
	\includegraphics[width=0.19\linewidth]{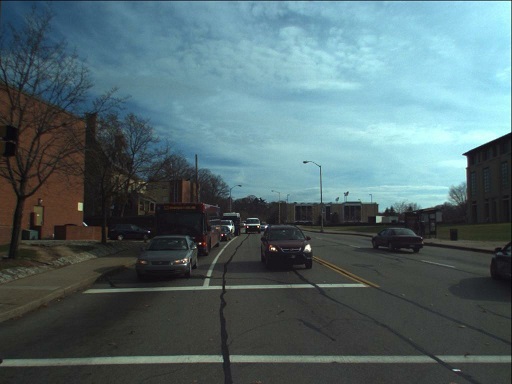}
	\includegraphics[width=0.19\linewidth]{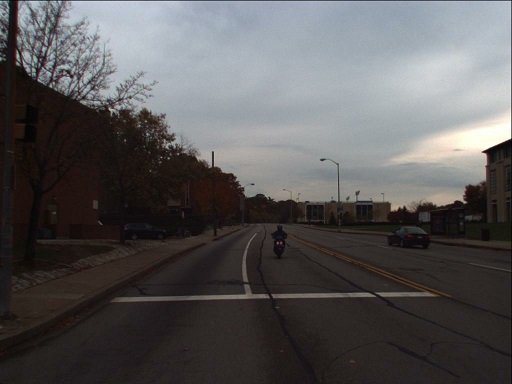}
	\includegraphics[width=0.19\linewidth]{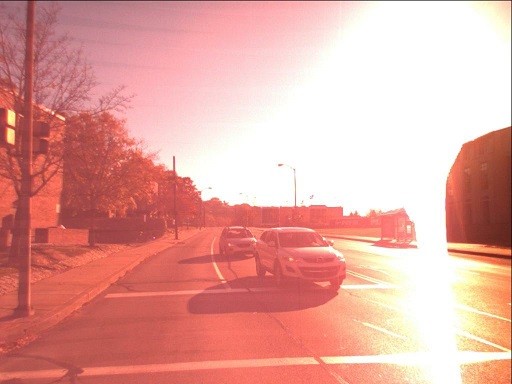} \\ \vspace{0.05cm}
	\includegraphics[width=0.19\linewidth]{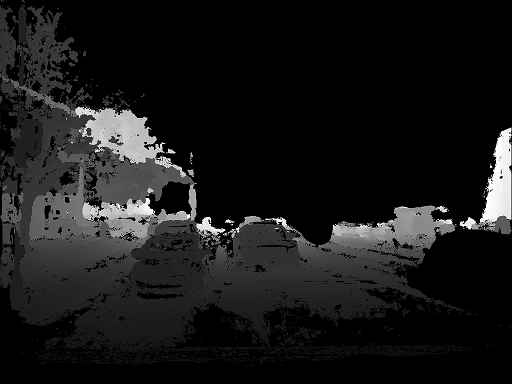} \includegraphics[width=0.19\linewidth]{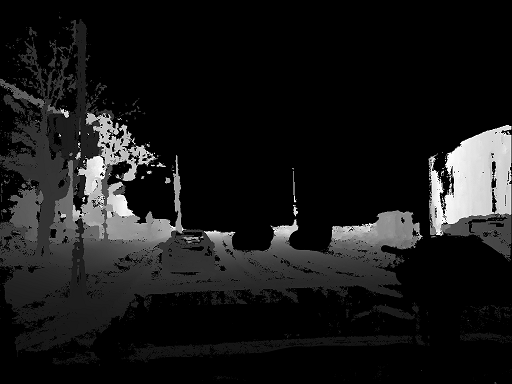}
	\includegraphics[width=0.19\linewidth]{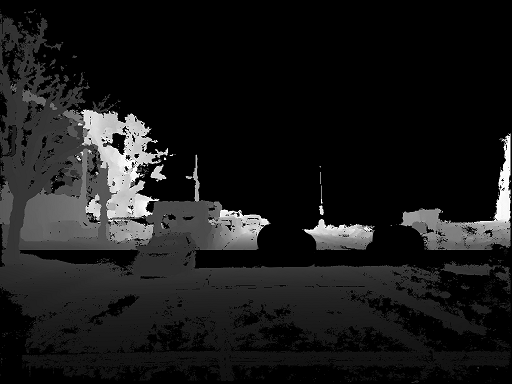}
	\includegraphics[width=0.19\linewidth]{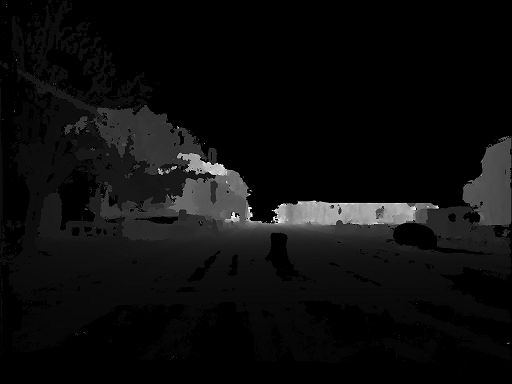}
	\includegraphics[width=0.19\linewidth]{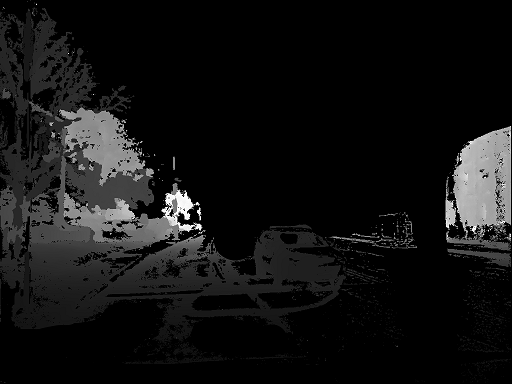}
	\vspace{-0.2cm}
	\caption{\textit{SeasonDepth} samples with depth maps under \textit{Cloudy + Foliage}, \textit{Low Sun + Foliage}, \textit{Cloudy + Mixed Foliage}, \textit{Overcast + Mixed Foliage} and \textit{Low Sun + Mixed Foliage}. }
		\vspace{-0.5cm}
	\label{example_dataset}
\end{figure*}

 The outdoor high-quality dense depth maps are not easy to obtain using LiDAR or laser scanner projection  \cite{geiger2012we,saxena2008make3d,antequera2020mapillary, diaz2022ithaca365}, or stereo matching  \cite{cordts2016cityscapes,Xian_2018_CVPR,xian2020structure}, let alone collections under multiple environments. We adopt Structure from Motion (SfM) and Multi-View Stereo (MVS) pipeline with RANSAC followed by careful manual post-processing to build a scaleless dense depth prediction dataset \textit{SeasonDepth} with multi-environment traverses based on the urban part of CMU Visual Localization dataset  \cite{sattler2018benchmarking,badino2011visual}. Some examples in the dataset are shown in Fig. \ref{example_dataset}.
 


For the benchmark on the proposed dataset, several statistical metrics are proposed for the experimental evaluation of the representative and state-of-the-art open-source methods from \textit{KITTI} leaderboard \cite{geiger2012we,Uhrig2017THREEDV}. The typical baselines we choose include supervised \cite{eigen2014depth, lee2019big,  yin2019enforcing,li2018megadepth,li2022depthformer, ranftl2021vision}, stereo training based self-supervised \cite{godard2017unsupervised,wong2019bilateral,tosi2019learning}, monocular video based self-supervised \cite{zhou2017unsupervised,guizilini20203d, godard2019digging,ranjan2019competitive,klingner2020self,zhou2021sub,watson2021temporal,xiang2022visual,jung2021fine,yan2021channel} and domain adaptation  \cite{atapour2018real, zheng2018t2net, zhao2019geometry} algorithms.
 \textcolor{revision}{Through thoroughly analyzing benchmark results, we find that most well-tuned methods cannot present satisfactory performance in terms of both mean and variance under multiple environments. Besides, through cross-dataset evaluations, current \textit{KITTI} pretrained models cannot  generalize well on our dataset while the models tuned on our dataset perform better on \textit{KITTI} \cite{geiger2012we} compared to models tuned on \textit{Cityscapes} \cite{cordts2016cityscapes}.   
 } Furthermore, the performance under each adverse environment is investigated both qualitatively and quantitatively to show hints to address robust perception against challenging environments.

\textcolor{revision2}{For the open problem of generalizability of learning-based depth prediction methods on different environmental conditions, our dataset is the first one that contains real-world RGB images with multiple environments under the same routes so that fair cross-environment evaluation and comparison can be conducted, giving hints to the future research on robust perception in changing environments.} In summary, our contributions in this work are listed as follows. 
\begin{itemize}
    \item A new monocular depth prediction dataset \textit{SeasonDepth} with the same multi-traverse routes under changing environments is introduced through SfM and MVS pipeline and is publicly available to the community.
    \item We benchmark best and representative open-sourced supervised and self-supervised prediction methods on \textit{SeasonDepth} using several new statistical metrics.
    \item From the extensive cross-environment and cross-dataset evaluation, \textcolor{revision}{we find that long-term robust depth prediction is still challenging and our dataset and benchmark can give future research direction by
    pointing out how adversary environments affect the performance with some promising hints to enhance robustness.}  
\end{itemize}

The rest of the paper is structured as follows. Sec. \ref{related work} analyzes the related work about depth prediction datasets and algorithms.  Sec. \ref{dataset} presents the process of building \textit{SeasonDepth}. Sec. \ref{benchmarksetup} introduces the metrics and benchmark setup. The experimental evaluation and analysis are shown in Sec. \ref{experiments}. Finally, in Sec. \ref{conclusion} we give the conclusions.

%

\begin{table*}[htbp]
        \centering
		\caption{Comparison between \textit{SeasonDepth} and Other Datasets}
		\label{datasets_compare}
		\resizebox{0.7\textwidth}{!}{
		\begin{tabular}{ccccccc}
			\toprule
			\textbf{Name} & \textbf{Scene}& \textbf{\textbf{\begin{tabular}[c]{@{}c@{}}Real or \\ Virtual\end{tabular}}} & \textbf{\textbf{\begin{tabular}[c]{@{}c@{}}Depth \\ Value\end{tabular}}} & \textbf{\textbf{\begin{tabular}[c]{@{}c@{}}Sparse or \\  Dense\end{tabular}}} & \textbf{\textbf{\begin{tabular}[c]{@{}c@{}}Multiple \\ Traverses\end{tabular}}} & \textbf{\textbf{\begin{tabular}[c]{@{}c@{}}Different \\ Environments\end{tabular}}} \\ \midrule
			NYUV2 \cite{silberman2012indoor}         & Indoor & Real                                                                & Absolute             & Dense                                                               & $\times$                                                               & $\times$                                                                        \\
			DIML \cite{kim2018deep}       & Indoor & Real                                                                   & Absolute             & Dense                                                               & $\times$                                                                & $\times$                                                                         \\
			iBims-1 \cite{Koch18_ECS}       & Indoor & Real                                                                   & Absolute             & Dense                                                               & $\times$                                                                & $\times$                                                                        \\
			Make3D \cite{saxena2008make3d}        & Outdoor \& Indoor & Real                                                                & Absolute             & Sparse                                                              & $\times$                                                     & $\times$                                                                       \\
			ReDWeb \cite{Xian_2018_CVPR}       & Outdoor \& Indoor & Real                                                                   & Relative             & Dense                                                               & $\times$                                                                & $\times$                                                                      \\
			WSVD \cite{wang2019web}       & Outdoor \& Indoor & Real                                                                   & Relative             & Dense                                                               & $\times$                                                                & $\times$                                                                        \\
			HR-WSI \cite{xian2020structure}       & Outdoor \& Indoor & Real                                                                   & Absolute             & Dense                                                               & $\times$                                                                & $\times$                                                                       \\
			DIODE \cite{vasiljevic2019diode}       & Outdoor \& Indoor & Real                                                                   & Absolute             & Dense                                                               & $\times$                                                                & $\times$                                                                      \\
			OASIS \cite{chen2020oasis}       & Outdoor \& Indoor & Real                                                                   & Relative             & Dense                                                               & $\times$                                                                & $\times$                                                                     \\
			3D Movies \cite{ranftl2020towards}       & Outdoor \& Indoor & Real                                                                   & Relative             & Dense                                                               & $\times$                                                                & $\checkmark$                                                                     \\
			KITTI \cite{geiger2012we}         & Outdoor & Real                                                                & Absolute             & Sparse                                                              & $\times$                                                               & $\times$                                                                        \\
			Cityscapes \cite{cordts2016cityscapes}    & Outdoor & Real                                                                & Absolute             & Dense                                                               & $\times$                                                                & $\checkmark$                                                                       \\
			
			DIW \cite{chen2016single}           & Outdoor & Real                                                                & Relative             & Sparse                                                              & $\times$                                                 & $\times$                                                                      \\
			MegaDepth \cite{li2018megadepth}       & Outdoor & Real                                                                   & Relative             & Dense                                                               & $\times$                                                                & $\times$                                                                       \\		
			DDAD \cite{guizilini20203d}       & Outdoor & Real                                                                   & Absolute             & Dense                                                               & $\times$                                                                & $\times$                                                                       \\
			MPSD \cite{antequera2020mapillary}       & Outdoor & Real                                                                   & Absolute             & Dense                                                               & $\times$                                                                & $\checkmark$                                                                       \\	
			V-KITTI \cite{gaidon2016virtual}       & Outdoor & Virtual                                                                   & Absolute             & Dense                                                               & $\checkmark$                                                                & $\checkmark$                                                                      \\
			SYNTHIA \cite{ros2016synthia}       & Outdoor & Virtual                                                                   & Absolute             & Dense                                                               & $\times$                                                                & $\times$                                                                       \\ 
			TartanAir \cite{tartanair}       & Outdoor \& Indoor & Virtual                                                                   & Absolute             & Dense                                                               & $\checkmark$                                                                & $\checkmark$                                                                       \\ 
			DeepGTAV \cite{gtav}       & Outdoor & Virtual                                                                   & Absolute             & Dense                                                               & $\checkmark$                                                                & $\checkmark$                                                                        \\ 
			
			\textbf{SeasonDepth}   & \textbf{Outdoor} & \textbf{Real}                                                                & \textbf{Relative}             & \textbf{Dense}                                                               & \boldmath$\checkmark$\unboldmath                                                 & \boldmath$\checkmark$\unboldmath    
			\\ \bottomrule
		\end{tabular}
	}
\end{table*} 
\section{Related Work}
\label{related work}
\subsection{Monocular Depth Prediction Datasets}
\label{related_datasets}
Depth prediction plays an  important role in the perception and localization of autonomous driving and other computer vision applications. Many indoor datasets are built through calibrated RGBD cameras \cite{silberman2012indoor,kim2018deep,Koch18_ECS}, expensive laser scanners \cite{saxena2008make3d,vasiljevic2019diode} and web stereo photos \cite{wang2019web,Xian_2018_CVPR,xian2020structure,ranftl2020towards,lai2019real}.
However, outdoor depth maps as ground truth are more complex to get, \textit{e.g.} projecting 3D point cloud data onto the image plane \cite{geiger2012we,saxena2008make3d,antequera2020mapillary} for sparse maps and using stereo matching to calculate inaccurate and limited-scope depth \cite{cordts2016cityscapes,ranftl2020towards,Xian_2018_CVPR}. 
 Another way to get the depth map is through SfM \cite{chen2016single,li2018megadepth,chen2020oasis,antequera2020mapillary} from monocular sequences.
 Although this method is time-consuming, it generates pretty accurate relatively-scaled dense depth maps
 , which is more general for depth prediction under different scenarios. 
 \textcolor{revision2}{For the long-term robust perception under changing environments,  though some real-world datasets \cite{cordts2016cityscapes,antequera2020mapillary,ranftl2020towards} include some environmental changes, there are still no multi-environment traverses with the same routes, which is essential and necessary for the fair evaluation of robustness across different environments. Since graphical rendering is becoming more and more realistic, some virtual synthetic datasets \cite{gaidon2016virtual,ros2016synthia,tartanair,gtav, sun2022shift} contain multi-environment traverses. But the rendered RGB images are still different from real-world ones due to the domain gap and cannot be used to benchmark real-world cross-environment performance. The details of the comparison between datasets are shown in Tab. \ref{datasets_compare} and Sec. \ref{comparison}.  The closest work to ours is the Ithaca365 \cite{diaz2022ithaca365}, where images and point clouds are collected from multiple environments for different perception tasks. But they do not involve the task of monocular depth prediction but only stereo disparity estimation with LiDAR points as ground truth. }
 


\subsection{Monocular Depth Prediction Algorithms}
\label{baselines}
The monocular depth prediction task aims to predict the dense depth map  in an active way given one single RGB image. Early studies including CRF \cite{xu2017multi, yuan2022new} and other graph models \cite{saxena2006learning,saxena2008make3d,liu2010single} largely depend on man-made descriptors, constraining the performance of depth prediction.
Afterward, studies based on CNNs  \cite{eigen2014depth,eigen2015predicting,laina2016deeper,song2021monocular, kim2022global} and Transformers \cite{ranftl2021vision, li2022depthformer, bhat2021adabins} have shown promising results for monocular depth estimation. 
Eigen \textit{et al.} \cite{eigen2014depth} first predict depth maps using CNN model, while \cite{laina2016deeper} introducing fully convolutional neural networks to regress the depth value.
 After that, supervised methods for monocular depth prediction have been well studied through normal estimation \cite{yin2019enforcing,Kusupati_2020_CVPR,patil2022p3depth}, the supervision of depth maps and stereo disparity ground truth \cite{li2018megadepth,fu2018deep,lee2019big,xian2020structure,vip_deeplab}.
However, since outdoor depth map ground truth is  expensive and time-consuming to obtain, self-supervised   depth estimation methods
have appeared using stereo geometric left-right consistency \cite{garg2016unsupervised,godard2017unsupervised,luo2018single,wong2019bilateral,tosi2019learning,gonzalezbello2020forget}, egomotion-pose constraint through monocular video \cite{zhou2017unsupervised,mahjourian2018unsupervised,casser2019depth,guizilini20203d,godard2019digging,ruhkamp2021attention,zhou2021r,masoumian2022gcndepth,guizilini2022multi,shu2020feature,he2022ra,zhao2020towards,spencer2020defeat,zhou2021sub,watson2021temporal} and multi-task learning with optical flow, motion and semantics segmentation \cite{yin2018geonet,zou2018df,ranjan2019competitive,klingner2020self,jung2021fine,lee2021learning} inside monocular video training pipeline as secondary supervisory signals. Furthermore, some problems posed by self-supervised learning strategies, such as dynamic objects\cite{saunders2022dyna,lee2021learning,chen2022self}, and scale consistency\cite{zhang2022towards,bian2019unsupervised,bian2021unsupervised,jiang2022detaching}, have also been well studied.
Besides, to avoid using expensive real-world depth ground truth, other algorithms are trained on synthetic virtual datasets \cite{gaidon2016virtual,ros2016synthia,tartanair,gtav} to leverage high-quality depth maps with zero cost. Such methods \cite{zheng2018t2net,atapour2018real,chen2019learning,zhao2019geometry,bozorgtabar2019syndemo,gurram2021monocular} confront the domain adaptation from synthetic  to real-world domain only supervised by virtual images for model training.

\section{SeasonDepth Dataset}
\label{dataset}
Our proposed dataset \textit{SeasonDepth} is derived from CMU Visual Localization dataset \cite{badino2011visual} through SfM algorithm. The original CMU Visual Localization dataset  covers over one year in Pittsburgh, USA, including 12 different environmental conditions. 
Images were collected from two identical cameras on the left and right of the vehicle along a route of 8.5 kilometers. 
And this dataset is also derived for long-term visual localization \cite{sattler2018benchmarking} by calculating the 6-DoF camera pose of images with more appropriate categories about the weather, vegetation and area. To be consistent with the content of driving scenes in other datasets like \textit{KITTI}, we adopt images from Urban areas categorized in \cite{sattler2018benchmarking} to build our dataset. 
More details about the dataset can be found in Appendix Sec. \ref{dataset_details}.

\subsection{Dense Reconstruction and Post-processing}
\label{dataset_buildup}
We reconstruct the dense model for each traversal under every environmental condition through SfM and MVS pipeline \cite{schonberger2016pixelwise}\textcolor{revision}{, which is commonly used for depth reconstruction \cite{guizilini20203d,li2018megadepth} and most suitable for multi-environment dense reconstruction for 3D mapping \cite{larsson2019cross,sattler2018benchmarking} and show advantage on the aspects of high dense quality despite of huge computational efforts compared to active sensing from LiDAR}. Specifically, similar to \textit{MegaDepth} \cite{li2018megadepth}, COLMAP \cite{schonberger2016structure,schonberger2016pixelwise} with SIFT descriptor \cite{lowe2004distinctive} is used to obtain the depth maps through photometric and geometric consistency from sequential images. 
\begin{figure}[h]
	\centering
		\begin{minipage}[t]{0.19\linewidth}
			\centering
			\includegraphics[width=\linewidth]{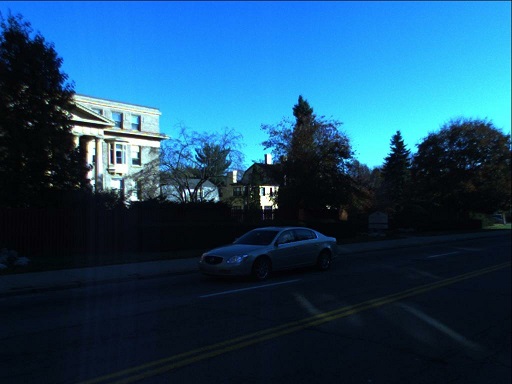}\\
			\vspace{0.05cm}
			\includegraphics[width=\linewidth]{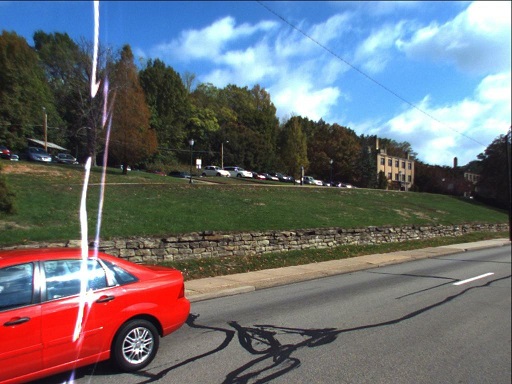}\\
			\vspace{0.05cm}
			\centerline{\scriptsize  RGB Images}
		\end{minipage}%
		\begin{minipage}[t]{0.19\linewidth}
			\centering
			\includegraphics[width=\linewidth]{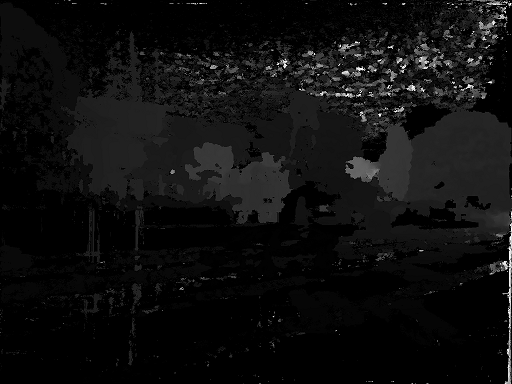}\\
			\vspace{0.05cm}
			\includegraphics[width=\linewidth]{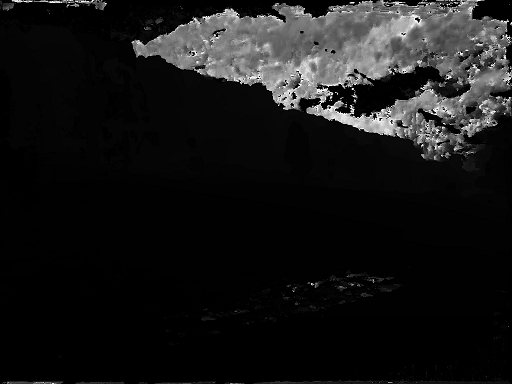}\\
			\vspace{0.05cm}
			\centerline{\scriptsize  After SfM}
		\end{minipage}%
	\begin{minipage}[t]{0.19\linewidth}
		\centering
		\includegraphics[width=\linewidth]{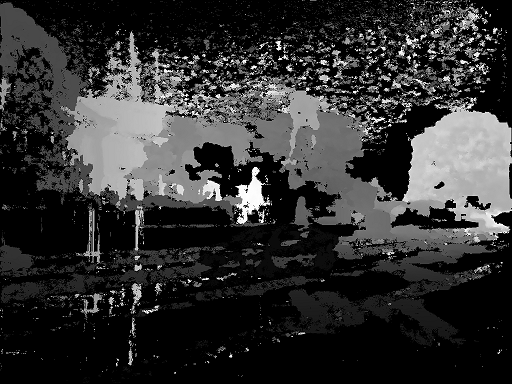}\\
		\vspace{0.05cm}
		\includegraphics[width=\linewidth]{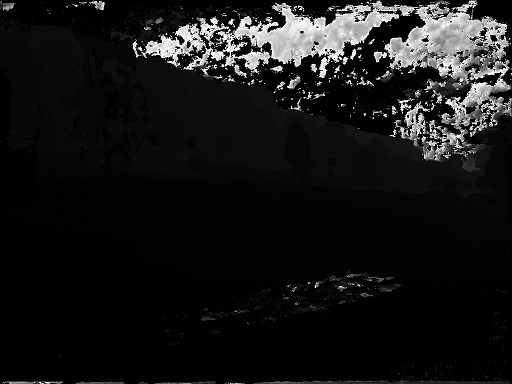}\\
		\vspace{0.05cm}
		\centerline{\scriptsize   Range Filtering}
	\end{minipage}%
	\begin{minipage}[t]{0.19\linewidth}
		\centering
		\includegraphics[width=\linewidth]{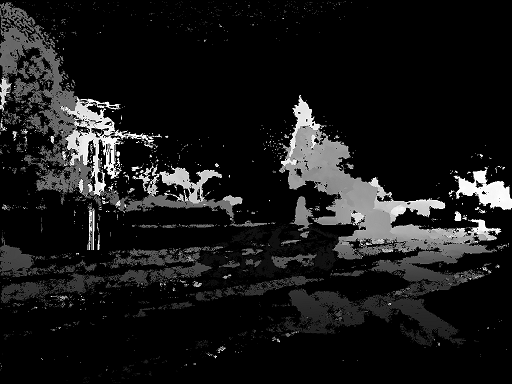}\\
		\vspace{0.05cm}
		\includegraphics[width=\linewidth]{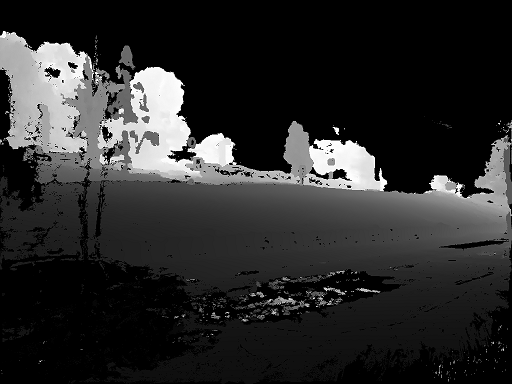}\\
		\vspace{0.05cm}
		\centerline{\scriptsize   HSV Filtering}
	\end{minipage}%
	\begin{minipage}[t]{0.19\linewidth}
		\centering
		\includegraphics[width=\linewidth]{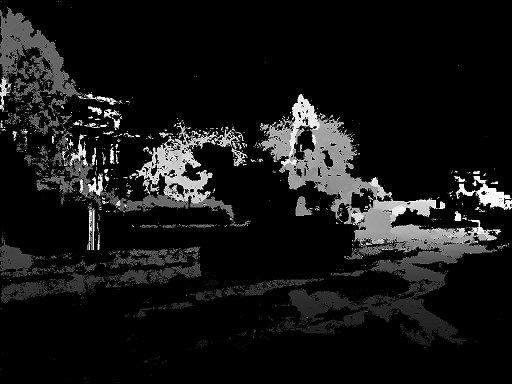}\\
		\vspace{0.05cm}
		\includegraphics[width=\linewidth]{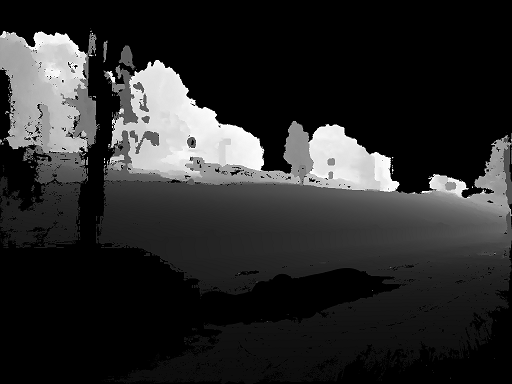}\\
		\vspace{0.05cm}
		\centerline{\scriptsize   Post-processing}
	\end{minipage}%
	\caption{The illustration of depth map processing.	}
	\label{ransac_filter}
\end{figure}
Furthermore, we adopt RANSAC algorithm in the SfM to remove the inaccurate values of dynamic objects in the images through  effective modification in SIFT matching triangulation based on the original COLMAP\textcolor{revision}{, where dynamic objects with additional motion besides relative camera motion  do not obey the multi-view geometry constraint and should be removed as noise via RANSAC in bundle adjustment optimization}. Besides, from our justification experiments in Sec. \ref{model_training}, it is validated that using relative depth values and removing dynamic noise will not significantly influence the training and the performance of depth prediction models.
Because the MVS algorithm generates the depth maps with error pixel values that are out of range or too close, like the cloud in the sky or noisy points on a very near road, we filter those outside the normal range of the depth map.


After the reconstruction, based on the observation of noise distribution in the HSV color space, \textit{e.g.} blue pixels always appear in the sky and dark pixels always appear in the shade of the low sun, which tend to be noise in most cases, we remove the noisy values in the HSV color space given some specific thresholds. 
Though outliers are set to be empty in RANSAC, instance segmentation is adopted through MaskRCNN \cite{he2017mask} to fully remove the noise of dynamic objects. However, since it is difficult to generate accurate segmentation maps only for dynamic objects under drastically changing environments, we leverage human annotation as the last step to finally check the depth map.   The data processing is shown in Fig.\ref{ransac_filter} with normalization after each step. \textcolor{revision2}{Since we are rigorous and serious to the quality of valid depth pixels which are used for benchmark, we set most noise to be invalid (which causes some “holes” on the boundary from appearance) to avoid any possible pollution to the following benchmark, ensuring the reliable evaluation and benchmark results. } More details can be found in Appendix Sec. \ref{building_details}.




\begin{figure*}[]
	\centering
	\includegraphics[width=0.19\linewidth]{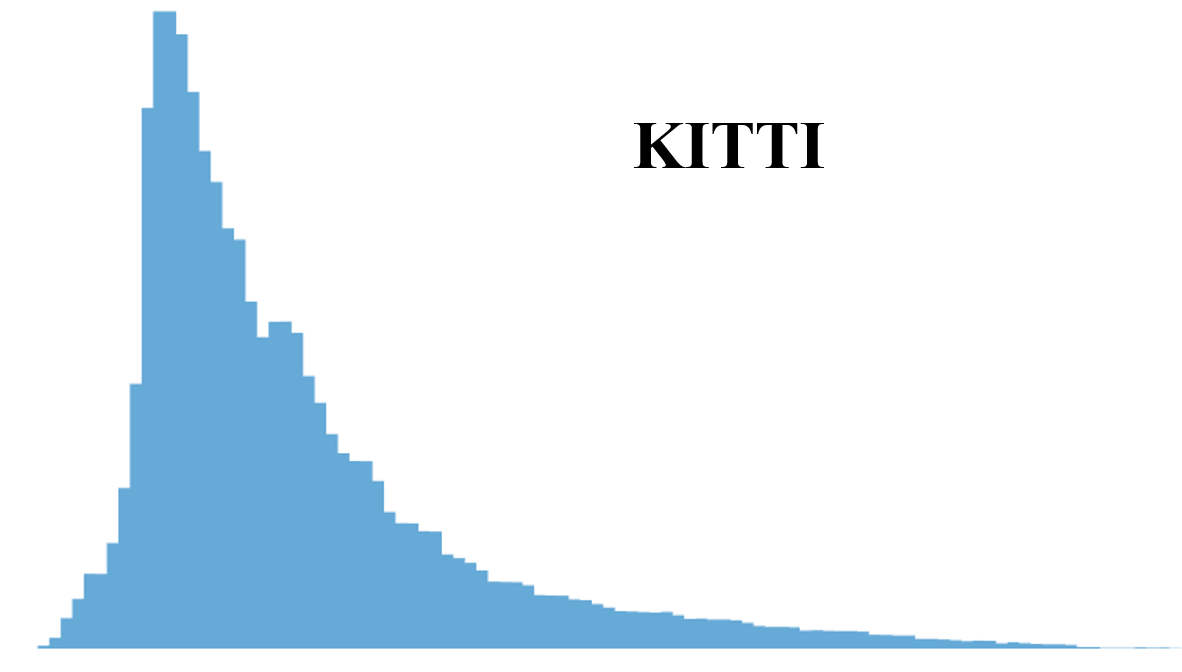} \includegraphics[width=0.19\linewidth]{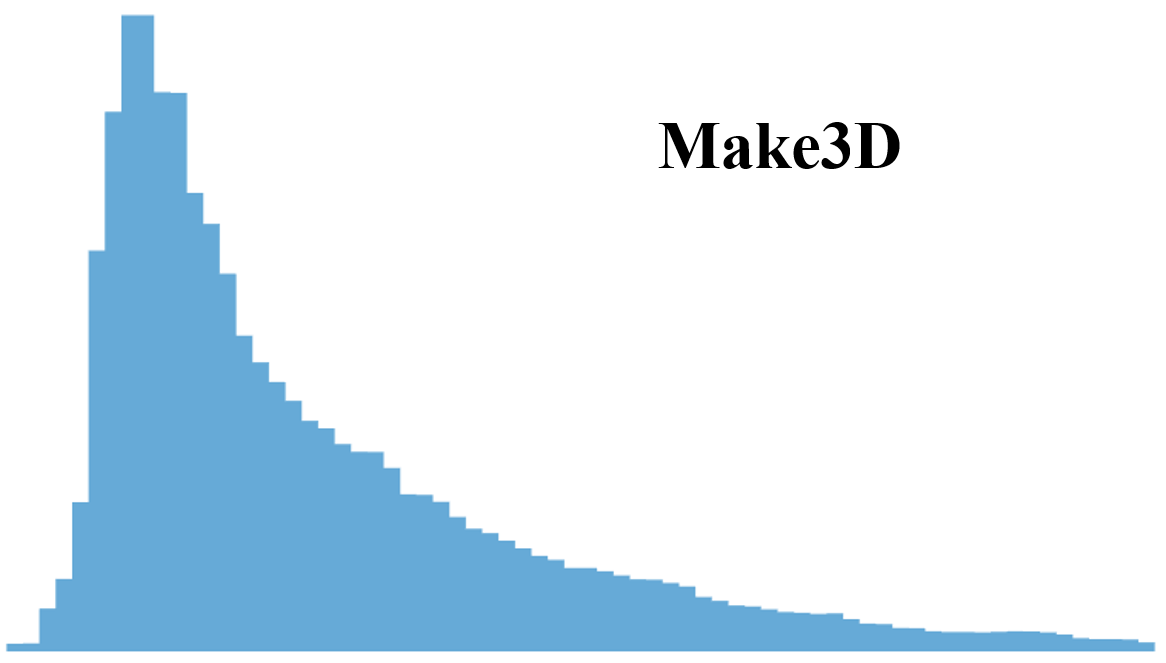}
	\includegraphics[width=0.19\linewidth]{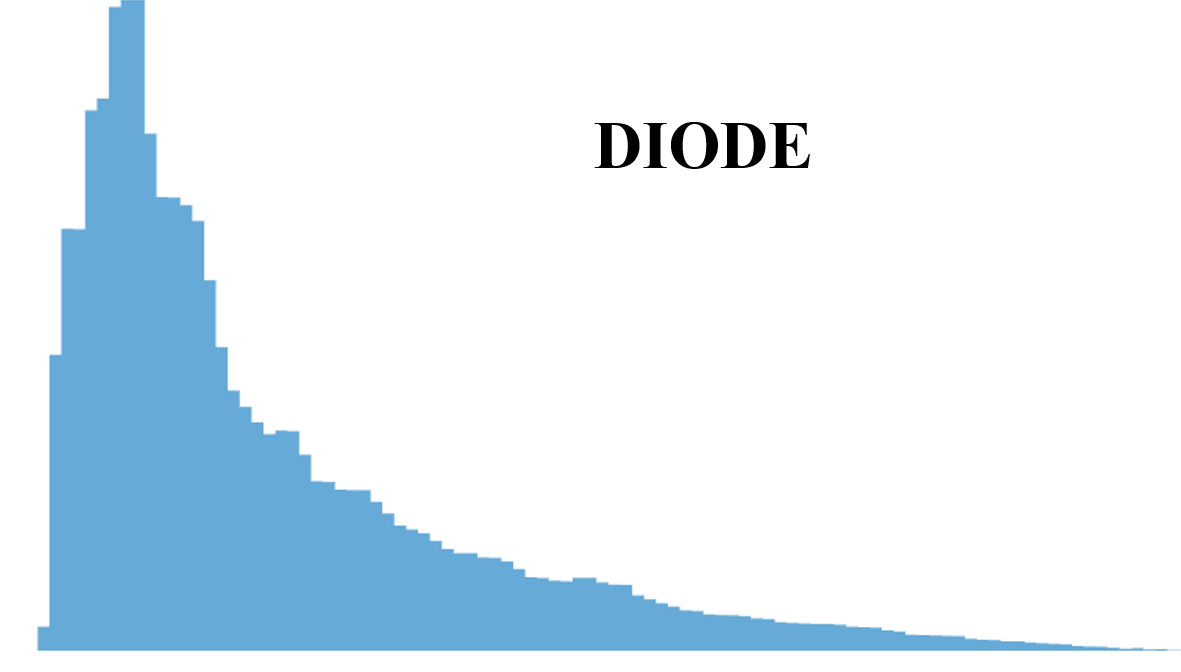}
	\includegraphics[width=0.19\linewidth]{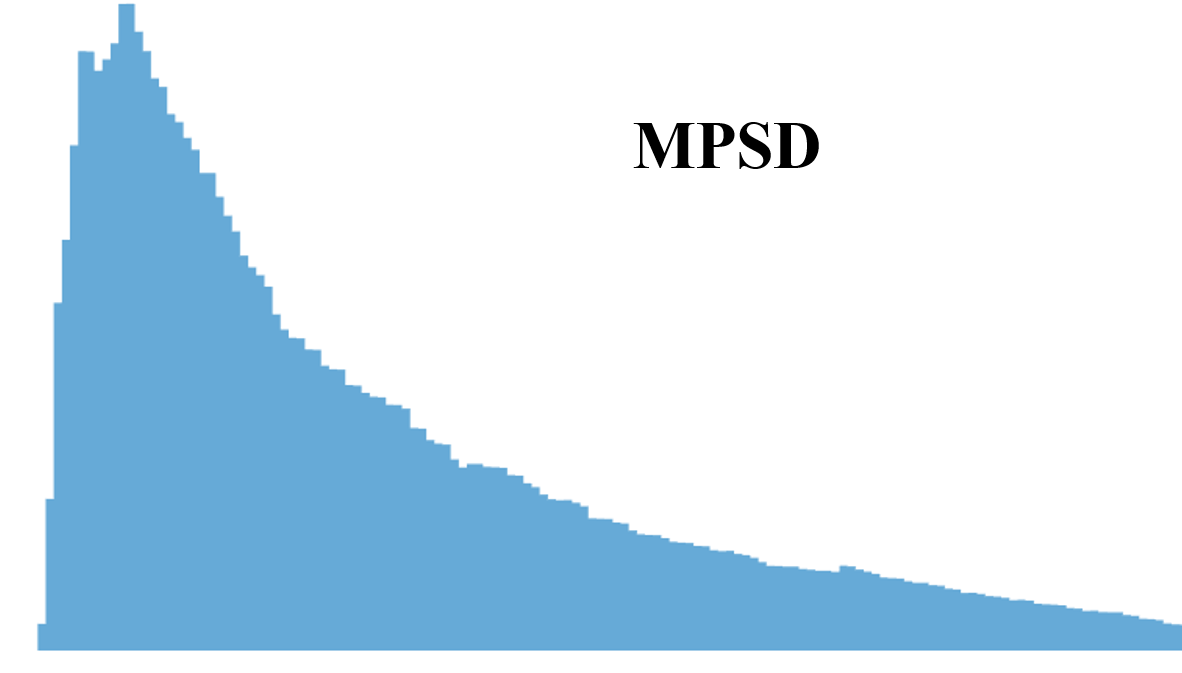}
	\includegraphics[width=0.19\linewidth]{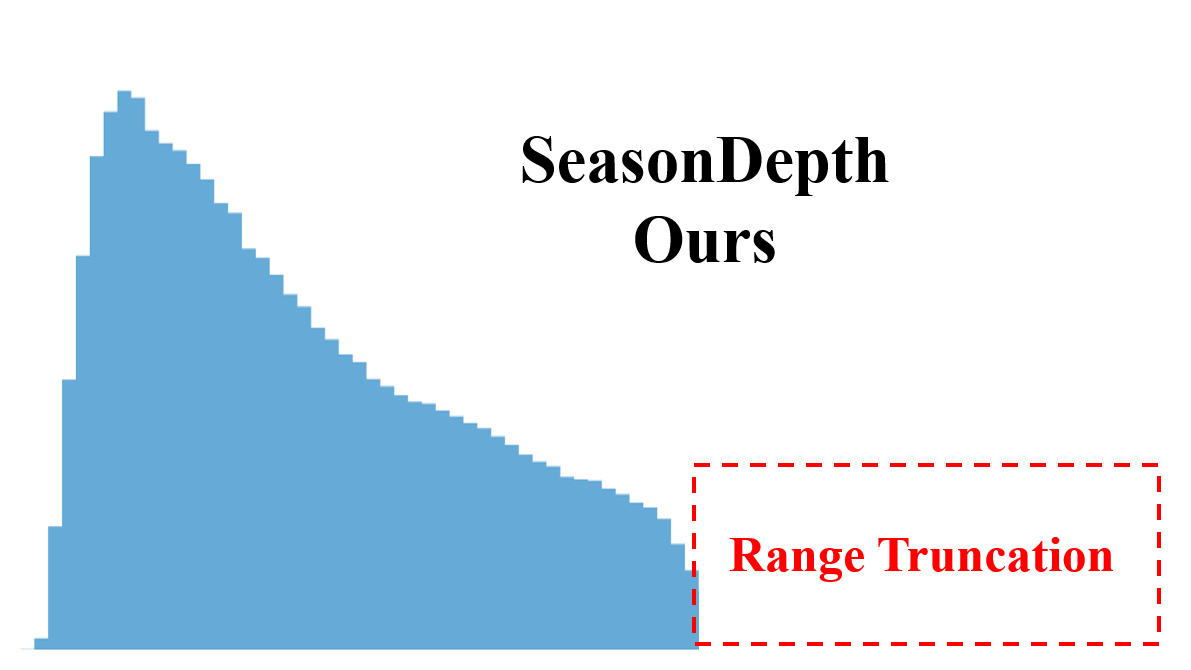}
	\caption{Comparison of relative depth distributions of several datasets. 
	}
	\label{distribution_map}
\end{figure*}

\subsection{Comparison with Other Datasets}
\label{comparison}
The current datasets are introduced in Sec. \ref{related_datasets}. The comparison between \textit{SeasonDepth} and current datasets is shown in Tab. \ref{datasets_compare}. \textcolor{revision2}{The distinctive feature of the proposed dataset is that \textit{SeasonDepth} contains comprehensive outdoor real-world multi-environment  sequences with repeated scenes, just like  virtual synthetic datasets \cite{gaidon2016virtual,gtav,tartanair,sun2022shift} but they are rendered from computer graphics and suffer from the huge domain gap.} Though real-word datasets \cite{antequera2020mapillary,ranftl2020towards,cordts2016cityscapes,sun2020scalability} include different environments, they lack the same-route traverses under different conditions so they are not able to fairly evaluate the performance across changing environments.
Similar to outdoor datasets \cite{chen2016single,li2018megadepth,chen2020oasis}, the depth maps of ours are scaleless with relative depth values, where the metrics should be designed for evaluation as the following section shows. The depth map ground truth from SfM is dense compared to LiDAR-based sparse depth maps. 
Besides, the comparison of depth value distribution is shown in Fig. \ref{distribution_map}. Note that the values of our dataset are scaleless and relative, so the x-axes of other datasets are also omitted for a fair comparison. 
We normalize the depth values for all the environments to mitigate the influence of the aggregation from relative depth  distributions under different environments  to get the final distribution map. The details of implementation can be found in Appendix Sec. \ref{stat_dataset}. From Fig. \ref{distribution_map}, it can be seen that our dataset also follows the long-tail distribution \cite{jiao2018look} which is the same as other datasets, with a difference of missing large-depth part due to range truncation during the building process in Sec. \ref{dataset_buildup}.

\section{Benchmark Setup}
\label{benchmarksetup}

\subsection{Evaluation Metrics}
\label{eval_measures}
The challenge for the design of evaluation metrics lies in two folds. One is to cope with scaleless and partially-valid dense depth map ground truth, and the other is to fully measure the depth prediction average performance and the stability or robustness across different environments. \textcolor{revision}{Due to the scaleless ground truth of relative depth value, some common metrics \cite{Uhrig2017THREEDV} cannot be used for evaluation directly. Since the focal lengths of two cameras are close enough to generate similarly-distributed depth values, unlike \cite{zhou2017unsupervised,li2018megadepth,chen2020oasis}, we align the distribution of depth prediction to depth ground truth via mean value and variance for a fair evaluation. The other key point for multi-environment evaluation lies in the reflection of robustness to changing environments for same-route sequences, which has not been studied in the previous work to the best of our knowledge. We formulate our metrics below.}


 First, for each pair of predicted and ground truth depth maps, the valid pixels $ D^{i,j}_{valid_{predicted}} $ of the predicted depth map $ D_{valid_{predicted}} $ are determined by non-empty valid pixels $ D^{i,j}_{valid_{GT}} $ of the depth map ground truth. And then the valid mean and variance of both $ D_{valid_{GT}} $ and $ D_{valid_{predicted}} $ are calculated as $ Avg_{GT} $,$ Avg_{pred} $ and $ Var_{GT} $,$ Var_{pred} $.
Then we  adjust the predicted depth map $ D_{adj} $ to get the same distribution with $ D_{valid_{GT}} $,
$${D_{adj}} = ({D_{pred}} - Av{g_{pred}}) \times \sqrt {{{Va{r_{GT}}} \mathord{\left/
			{\vphantom {{Va{r_{GT}}} {Va{r_{pred}}}}} \right.
			\kern-\nulldelimiterspace} {Va{r_{pred}}}}}  + Av{g_{GT}}$$

\begin{figure}[htbp]
	\centering
		\begin{minipage}[t]{0.24\linewidth}
			\centering
			\includegraphics[width=\linewidth]{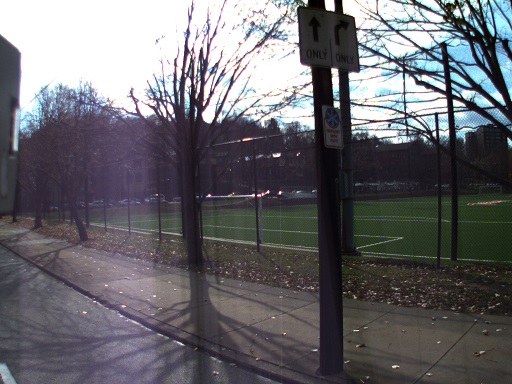}\\
			\vspace{0.05cm}
			\includegraphics[width=\linewidth]{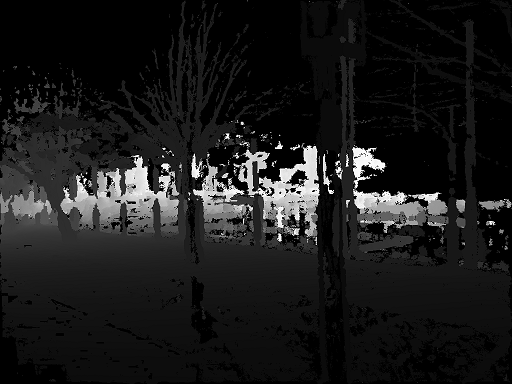}\\
			\vspace{0.05cm}
			\centerline{\scriptsize RGB and ground truth }
		\end{minipage}%
		\begin{minipage}[t]{0.24\linewidth}
			\centering
			\includegraphics[width=\linewidth]{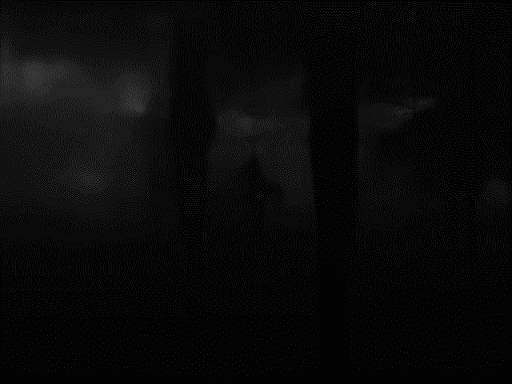}\\
			\vspace{0.05cm}
			\includegraphics[width=\linewidth]{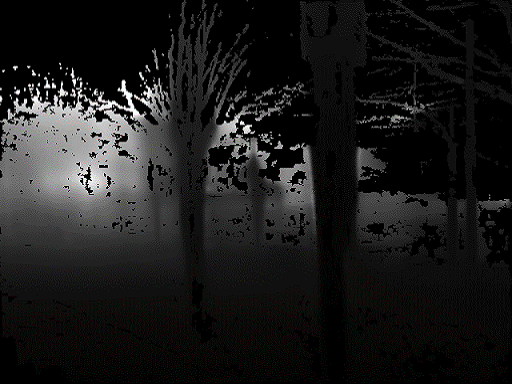}\\
			\vspace{0.05cm}
			\centerline{\scriptsize\textit{PackNet} \cite{guizilini20203d}}
		\end{minipage}%
		\begin{minipage}[t]{0.24\linewidth}
			\centering
			\includegraphics[width=\linewidth]{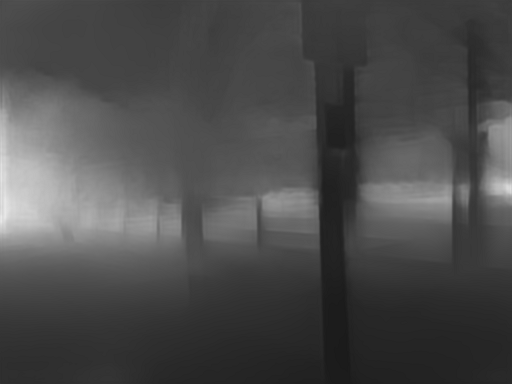}\\
			\vspace{0.05cm}
			\includegraphics[width=\linewidth]{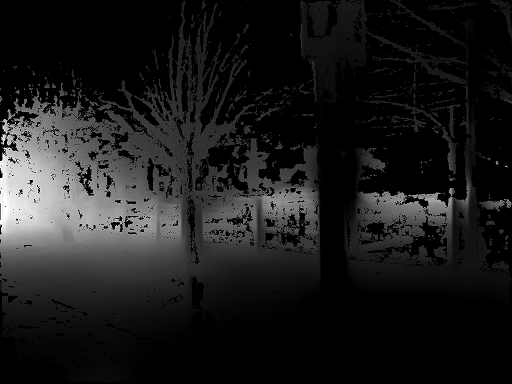}\\
			\vspace{0.05cm}
			\centerline{\scriptsize\textit{VNL} \cite{yin2019enforcing}}
		\end{minipage}%
		\begin{minipage}[t]{0.24\linewidth}
			\centering
			\includegraphics[width=\linewidth]{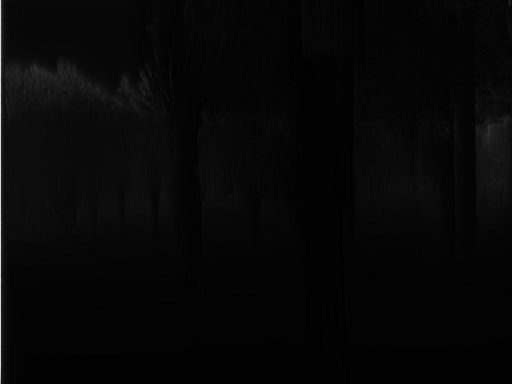}\\
			\vspace{0.05cm}
			\includegraphics[width=\linewidth]{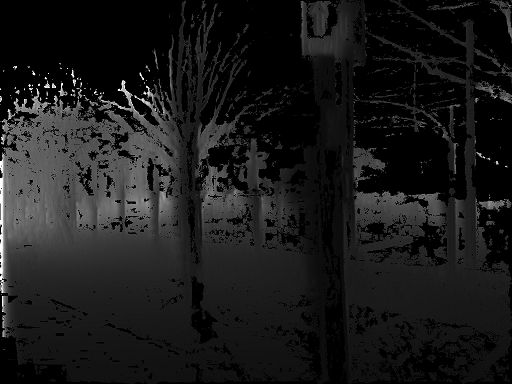}\\
			\vspace{0.05cm}
			\centerline{\scriptsize\textit{GASDA} \cite{zhao2019geometry}}
		\end{minipage}%
	\caption{The examples of depth adjustment (from the first to second row) for prediction results.}
	\label{adjusted_results}
\end{figure}
The examples of adjusted depth prediction are shown in Fig. \ref{adjusted_results}. After this operation, we can eliminate scale difference for  depth prediction across datasets, which makes this zero-shot evaluation on \textit{SeasonDepth} reliable and applicable  to all the models even though they predict absolute depth values, showing generalization ability on new datasets and robustness across different environments. 
Denote the adjusted valid depth prediction $ D_{adj} $  as $ D_{P} $ in the following formulation. 
To measure the depth prediction performance, we choose the most distinguishable metrics under multiple environments from commonly-used metrics in  \cite{Uhrig2017THREEDV}, \textit{AbsRel} and $\delta < 1.25 $ ($ a_1 $). 
For environment $ k $,  we have,
$$AbsRel^{k} = \frac{1}{n} \sum\limits_{i,j}^n {{{\left| {{D_P}{{^k}_{i,j}} - {D_{GT}}{{^k}_{i,j}}} \right|} \mathord{\left/
			{\vphantom {{\left| {{D_P}{{^k}_{i,j}} - {D_{GT}}{{^k}_{i,j}}} \right|} {{D_{GT}}{{^k}_{i,j}}}}} \right.
			\kern-\nulldelimiterspace} {{D_{GT}}{{^k}_{i,j}}}}}$$
$$	a_{1}^{k}  =\frac{1}{n}\sum\limits_{i,j}^n {\mathbbm{1}(max\{ \frac{{D_P}{{^k}_{i,j}}}{{D_{GT}}{{^k}_{i,j}}},\frac{{D_{GT}}{{^k}_{i,j}}}{{D_{P}}{{^k}_{i,j}}}\}  < 1.25} )$$
For the evaluation under different environments, 6 secondary  metrics are derived based on original metrics, 
$$AbsRel^{avg} = \frac{1}{m}\sum\limits_k {{AbsRel^k}}, a_1^{avg} = \frac{1}{m}\sum\limits_k {{a_1^k}}$$
$$AbsRel^{var} = \frac{1}{m}{\sum\limits_k {\left| {{AbsRel^k} - \frac{1}{m}\sum\limits_k {{AbsRel^k}} } \right|} ^2} $$
$$a_1^{var} = \frac{1}{m}{\sum\limits_k {\left| {{a_1^k} - \frac{1}{m}\sum\limits_k {{a_1^k}} } \right|} ^2}$$


where $ avg $ terms $ AbsRel^{avg} $, $ a_1^{avg} $ and $ var $ terms $ AbsRel^{var} $, $ a_1^{var} $ come from  \textit{Mean} and \textit{Variance} in statistics, indicating the average performance and the fluctuation around the mean value across multiple environments. 

Considering the depth prediction applications, it should be more rigorous to prevent better results fluctuation than worse results under changing conditions. Therefore, we use the \textit{Relative Range} terms $ AbsRel^{relRng} $, $ a_1^{relRng} $ to calculate the relative difference of maximum and minimum for all the environments. 
$$AbsRe{l^{relRng}} = \frac{\max\{ AbsRe{l^k}\}  - \min\{ AbsRe{l^k}\}}{\frac{1}{m}\sum\limits_k {AbsRe{l^k}}}$$
$$a_1^{relRng} = \frac{\max\{ 1 - a_1^k\}  - \min\{ 1 - a_1^k\}}{\frac{1}{m}\sum\limits_k {(1 - a_1^k)}}$$

\textit{Relative Range} terms for \textit{AbsRel} and $ 1-a_1 $ are more strict than the \textit{Variance} terms $ AbsRel^{var} $, $ a_1^{var} $ and note that $ 1 - a_1 $ instead of $ a_1 $ is used to calculate $ a_1^{relRng} $ to make relative range fluctuation more distinguishable for better methods.


\subsection{Benchmark Design and Algorithms} 
In the experiment, we aim to first benchmark the well-tuned performance on \textit{SeasonDepth} using state-of-the-art algorithms and then present the cross-dataset performance with other datasets using representative baselines of each category.  More details can be found in Appendix Sec. \ref{benchmark_details} and \ref{comparison_cityscape}.

We first split the split training set, validation set, and test set with 11407, 17225 and 3944 images respectively. Note that the detailed analysis for each environment is based on the validation set which requires more images. For the benchmark on \textit{SeasonDepth}, though there is no limit to other datasets or pre-trained models to obtain the best performance, since \textit{SeasonDepth} only has monocular images as the training set, we categorize the state-of-the-art evaluated algorithms as supervised methods and self-supervised methods with monocular video training. Specifically, DepthFormer \cite{li2022depthformer}, BTS \cite{lee2019big} are DPT \cite{ranftl2021vision} are supervised baselines, while SUB-Depth	 \cite{zhou2021sub},\textit{VADepth} \cite{xiang2022visual}, Monodepth2 \cite{godard2019digging}, SfMLearner \cite{zhou2017unsupervised} and ManyDepth   \cite{watson2021temporal} are self-supervised baselines. 

\begin{table*}[]
		\caption{\textit{SeasonDepth} Benchmark (\color{red}\bm{$\downarrow$}\color{black}: Lower Better, \color{blue}\bm{$\uparrow$}\color{black}: Higher Better, \textbf{Best} for each category)}
		\label{best_evalation_results}
		\centering
		\resizebox{0.8\textwidth}{!}{
	\begin{tabular}{cccccccc}
		\toprule
		\multicolumn{2}{c}{}                                                                                                                       & \multicolumn{2}{c}{\textbf{Average}}                        & \multicolumn{2}{c}{ \textbf{Variance}$ (10^{-2}) $}                        & \multicolumn{2}{c}{ \textbf{Relative Range}}                 \\
\multirow{-2}{*}{Category} &\multirow{-2}{*}{Method}
	   & \cellcolor[HTML]{FFCCC9}$ AbsRel $ \color{red}\bm{$\downarrow$} & \cellcolor[HTML]{CBCEFB}$ a_1 $ \color{blue}\bm{$\uparrow$}    & \cellcolor[HTML]{FFCCC9}$ AbsRel  $ \color{red}\bm{$\downarrow$} & \cellcolor[HTML]{FFCCC9}$ a_1 $ \color{red}\bm{$\downarrow$}     & \cellcolor[HTML]{FFCCC9}$ AbsRel  $ \color{red}\bm{$\downarrow$} & \cellcolor[HTML]{FFCCC9}$ 1-a_1 $ \color{red}\bm{$\downarrow$}    \\ \midrule
		& DepthFormer	 \cite{li2022depthformer}                                                  & \textbf{0.135} &	\textbf{0.835} &	\textbf{0.0210} &	0.120 &	0.294 &	0.576\\
		& BTS \cite{lee2019big}                                   & 0.242 &	0.587 &	0.0222 &	\textbf{0.0632}	 & \textbf{0.220} &	\textbf{0.220 }                \\
	\multirow{-3}{*}{Supervised} & DPT   \cite{ranftl2021vision}                           &     0.152 &	0.790 &	0.0286 &	0.1574 &	0.364 &	0.637               \\ \midrule
	& SUB-Depth	 \cite{zhou2021sub}                                                  & \textbf{0.095} &	\textbf{0.920} &	0.008 &	\textbf{0.015} &	0.398 &	0.668\\ 
		& VADepth \cite{xiang2022visual}                                   & 0.131 &	0.852 &	\textbf{0.006} &	0.024	 & \textbf{0.247} &	0.397                 \\
			& Monodepth2 \cite{godard2019digging}                                   & 0.144 &	0.824 &	0.011 &	0.046 &	0.305 &	0.502                 \\
				& SfMLearner \cite{zhou2017unsupervised}                                   & 0.325 &	0.482 &	0.107 &	0.155 &	0.298 &	\textbf{0.236}                 \\
	\multirow{-5}{*}{\begin{tabular}[c]{@{}c@{}}Self-supervised\\ Monocular\\ Video Training\end{tabular}} & ManyDepth   \cite{watson2021temporal}                    &     0.227 &	0.649 &	0.080 &	0.262 &	0.486 &	0.549               \\ 
		\bottomrule
	\end{tabular}
}
\end{table*}

\begin{table*}[]
		\caption{Cross-dataset Generalization from \textit{KITTI} to \textit{SeasonDepth}  (\color{red}\bm{$\downarrow$}\color{black}: Lower Better, \color{blue}\bm{$\uparrow$}\color{black}: Higher Better, \textbf{Best} for each category)}
		\label{evalation_results}
		\centering
		\resizebox{0.9\textwidth}{!}{
	\begin{tabular}{cccccccccc}
		\toprule
		\multicolumn{2}{c}{}                                                                                                                 & \multicolumn{2}{c}{\textbf{\textit{KITTI} Eigen Split}}                                      & \multicolumn{2}{c}{\textbf{\textit{SeasonDepth}:} \textbf{Average}}                        & \multicolumn{2}{c}{ \textbf{Variance}$ (10^{-2}) $}                        & \multicolumn{2}{c}{ \textbf{Relative Range}}                 \\
\multirow{-2}{*}{Category} &\multirow{-2}{*}{Method}
		& \cellcolor[HTML]{FFCCC9}$ AbsRel$ \color{red}\bm{$\downarrow$} & \cellcolor[HTML]{CBCEFB}$ a_1 $ \color{blue}\bm{$\uparrow$}    & \cellcolor[HTML]{FFCCC9}$ AbsRel $ \color{red}\bm{$\downarrow$} & \cellcolor[HTML]{CBCEFB}$ a_1 $ \color{blue}\bm{$\uparrow$}    & \cellcolor[HTML]{FFCCC9}$ AbsRel  $ \color{red}\bm{$\downarrow$} & \cellcolor[HTML]{FFCCC9}$ a_1 $ \color{red}\bm{$\downarrow$}     & \cellcolor[HTML]{FFCCC9}$ AbsRel  $ \color{red}\bm{$\downarrow$} & \cellcolor[HTML]{FFCCC9}$ 1-a_1 $ \color{red}\bm{$\downarrow$}    \\ \midrule
		& Eigen \textit{et al.} \cite{eigen2014depth}                         & 0.203                          & 0.702                         & 1.093                          & 0.340                         & 0.346                          & \textbf{0.0170}                         & 0.206                          & \textbf{0.0746}                         \\
		& BTS \cite{lee2019big}                                  & \textbf{0.060}   &  \textbf{0.955}  & 0.676                          & 0.209                         & 0.545                         & 0.0650                         & 0.405                          & 0.129                        \\
		& MegaDepth \cite{li2018megadepth}                            & 0.220                          & 0.632                         & 0.515                          & 0.417                         & \textbf{0.0874}                         & 0.0285                         & \textbf{0.200}                          & 0.107                         \\                                                \multirow{-5}{*}{Supervised}    & VNL \cite{yin2019enforcing}                                  & {0.072}                          & {0.938}                         & \textbf{0.306}                          & \textbf{0.527}                         & 0.126                          & 0.166                          & 0.400                          & 0.290                        \\              
		\midrule
		& Monodepth \cite{godard2017unsupervised}    & 0.148  & 0.803 & \textbf{0.436}  & \textbf{0.455} & \textbf{0.0475} & \textbf{0.0213} & 0.198  & 0.104 \\
		& adareg \cite{wong2019bilateral}       & 0.126  & 0.840 & 0.507  & 0.405 & 0.0630 & 0.0474 & \textbf{0.178}  & \textbf{0.0137 }\\
		\multirow{-3}{*}{\begin{tabular}[c]{@{}c@{}}Self-supervised\\ Stereo Training\end{tabular}}   & monoResMatch \cite{tosi2019learning} & \textbf{0.096}  & \textbf{0.890} & 0.487  & 0.389 & 0.286  & 0.0871 & 0.414  & 0.160 \\ 
\midrule
		& SfMLearner \cite{zhou2017unsupervised}                           & 0.181                          & 0.733                         &0.360                          & 0.495                        & 0.0801                         & 0.0628                         & 0.269                         & 0.182                         \\
		& PackNet \cite{guizilini20203d}                              & 0.116                          & 0.865                         & 0.722                          & 0.421                         & 0.187                          & 0.0705                         & \textbf{0.186 }                         & \textbf{0.155}                        \\
		& Monodepth2 \cite{godard2019digging}   & 0.106  & 0.874 & {0.256}  & {0.624} & {0.0311} & 0.0532 & 0.235  & 0.229 \\
		& CC \cite{ranjan2019competitive}                                   & 0.140                          & 0.826                         & 0.648                          & 0.479                         & 0.223                          & 0.0881                         & 0.280                          & 0.241                         \\ & SGDepth \cite{klingner2020self}                              & 0.113                          & 0.879                         & 0.648                          & 0.480                       & 0.0987                         & 0.0498                         & 0.197                          & 0.169                   \\
			& FSRE-Depth \cite{jung2021fine}                                   & 0.105 & 0.886 & 0.256 & 0.624 & 0.0288 & 0.0283 & 0.227& 0.158                         \\ 
				& CADepth-Net \cite{yan2021channel}                                   & 0.105 & \textbf{0.892} & 0.257 & 0.625 & 0.0447 & 0.0725 & 0.265 & 0.278                         \\
		 \multirow{-6}{*}{\begin{tabular}[c]{@{}c@{}}Self-supervised\\ Monocular\\ Video Training\end{tabular}} & VADepth \cite{xiang2022visual}                                   & \textbf{0.104} & \textbf{0.892} & \textbf{0.230} & \textbf{0.667} & \textbf{0.0158} & \textbf{0.0215} & 0.205 & 0.179                         \\
\midrule
		& Atapour \textit{et al.} \cite{atapour2018real}                       & \textbf{0.110}                          & \textbf{0.923}                         & 0.687                          & 0.300                         & 0.224                          & \textbf{0.0220}                        & \textbf{0.231}                          & \textbf{0.0622}                         \\
		& T2Net \cite{zheng2018t2net}                                & 0.169                          & 0.769                         & 0.827                          & 0.391                         & 0.399                          & 0.0799                         & 0.286                          & 0.146                         \\
		  \multirow{-3}{*}{\begin{tabular}[c]{@{}c@{}}Syn-to-real\\ Domain\\ Adaptation\end{tabular}}   & GASDA \cite{zhao2019geometry}        & 0.143  & 0.836 & \textbf{0.438}  & \textbf{0.411} & \textbf{0.121}  & 0.0665 & 0.271  & 0.145 \\
		\bottomrule
	\end{tabular}
}
\end{table*}
For the cross-dataset performance with other datasets, we choose the other two popular autonomous driving datasets  \textit{KITTI} and \textit{Cityscapes} together with \textit{SeasonDepth}. To analyze the performance under each environment, we report the results on the validation set of \textit{SeasonDepth}. We first present generalization performance from \textit{KITTI} to \textit{SeasonDepth}. Following the category introduced in Sec. \ref{baselines}, some representative baseline models  on \textit{KITTI}  leaderboard \cite{Uhrig2017THREEDV} are chosen to evaluate the performance on the \textit{SeasonDepth} dataset without fine-tuning. These methods include supervised methods (Eigen \textit{et al.} \cite{eigen2014depth}, \textit{BTS} \cite{lee2019big}, \textit{MegaDepth} \cite{li2018megadepth} and \textit{VNL} \cite{yin2019enforcing}), self-supervised methods with stereo training  (\textit{Monodepth} \cite{godard2017unsupervised}, \textit{adareg} \cite{wong2019bilateral}, \textit{monoResMatch} \cite{tosi2019learning}), self-supervised methods with monocular video training (\textit{SfMLearner} \cite{zhou2017unsupervised},  \textit{Monodepth2} \cite{godard2019digging}, \textit{PackNet} \cite{guizilini20203d},  \textit{CC} \cite{ranjan2019competitive}, \textit{SGDepth} \cite{klingner2020self}, \textit{FSRE-Depth} \cite{jung2021fine} \textit{CADepth-Net} \cite{yan2021channel}  \textit{VADepth} \cite{xiang2022visual}), and  domain adaptation methods (Atapour \textit{et al.} \cite{atapour2018real}, \textit{T2Net} \cite{zheng2018t2net}, \textit{GASDA} \cite{zhao2019geometry})  trained on the virtual dataset with multiple environments.

We then introduce cross-dataset comparison evaluation 
to justify that the depth accuracy and the ground truth are good enough for the dataset usage of autonomous driving for model training in spite of the lack of dynamic objects. Specifically, inspired by cross-dataset transfer degradation evaluation in \cite{ranftl2020towards}, we compare our dataset with the  stereo depth dataset \textit{Cityscapes} \cite{cordts2016cityscapes} in terms of the degraded performance on \textit{KITTI} dataset after cross-dataset fine-tuning. Based on the pre-trained models on \textit{KITTI},  we fine-tune BTS \cite{lee2019big} and SfMLearner \cite{zhou2017unsupervised} models on \textit{SeasonDepth} and \textit{Cityscapes} dataset with the same amount of images for 50 epochs, and evaluate the  depth prediction on \textit{KITTI} validation set  using the metrics of $MAE$, $absErrorRel$, $iMAE$, $iRMSE$, $sqErrorRel$ from \cite{Uhrig2017THREEDV} and report the mean and standard deviation from the last 10 training epochs.


\section{Experimental Evaluation Results}
\label{experiments}
\subsection{SeasonDepth Benchmark Results}
\label{seasondepth_bench}
In this section, we present the evaluation results on the test set of \textit{SeasonDepth} in Tab. \ref{best_evalation_results}. The models are well tuned on \textit{SeasonDepth} training set and have impressive performance on the test set, especially for $Average$ performance. We can see that self-supervised methods do not perform worse than supervised ones after well-tuning. It can be found that  DepthFormer \cite{li2022depthformer} and SUB-Depth\cite{zhou2021sub} perform the best on $Average$ but not satisfactory on $Variance$ or $Relative Range$, showing that even the well-tuned models cannot perform well consistently across different environments. Therefore, there is still a long way to go even for the state-of-the-art methods towards long-term robust depth estimation.


\begin{figure*}[htbp]
	\centering
		\begin{minipage}[t]{0.084\linewidth}
			\centering
			\includegraphics[width=\linewidth]{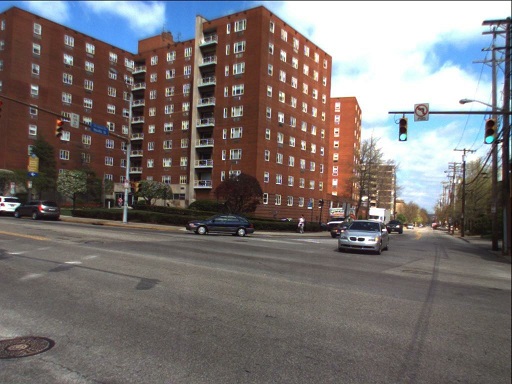}\\
			\includegraphics[width=\linewidth]{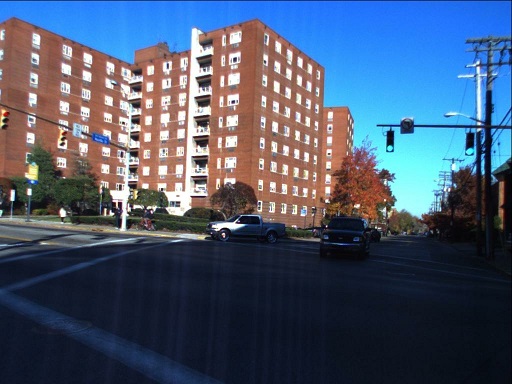}\\
			\includegraphics[width=\linewidth]{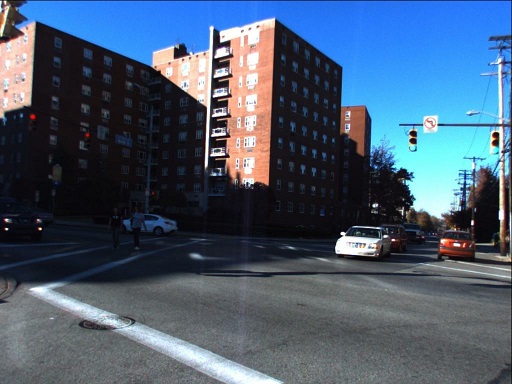}\\
			\centerline{\tiny RGB}
		\end{minipage}%
		\begin{minipage}[t]{0.084\linewidth}
			\centering
			\includegraphics[width=\linewidth]{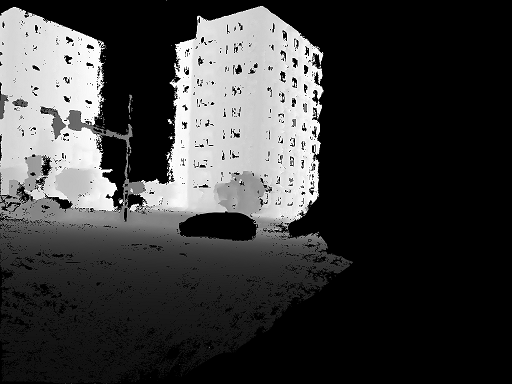}\\
			\includegraphics[width=\linewidth]{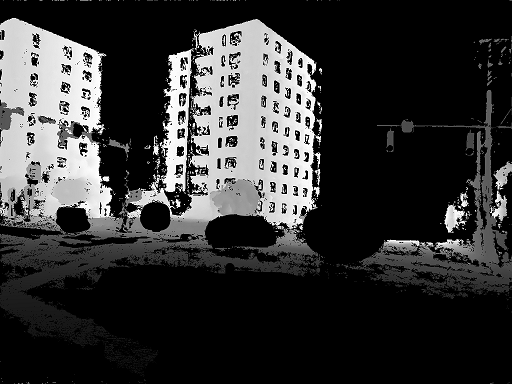}\\
			\includegraphics[width=\linewidth]{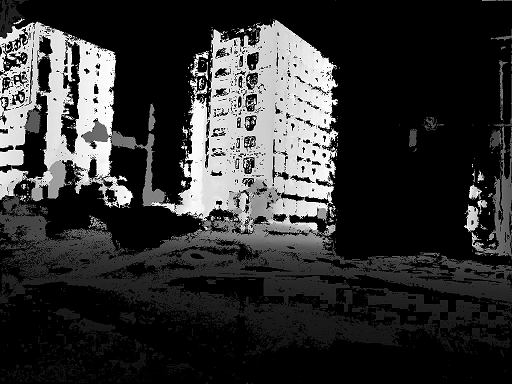}\\
			\centerline{\tiny Ground Truth}
		\end{minipage}%
		\begin{minipage}[t]{0.084\linewidth}
			\centering
			\includegraphics[width=\linewidth]{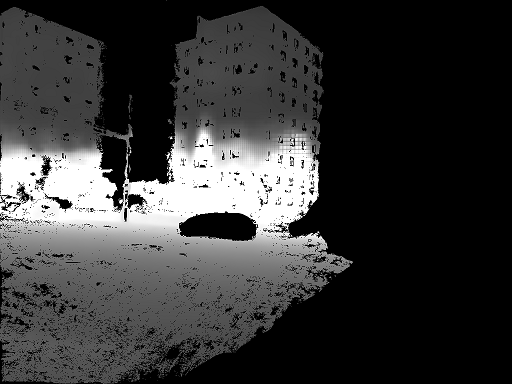}\\
			\includegraphics[width=\linewidth]{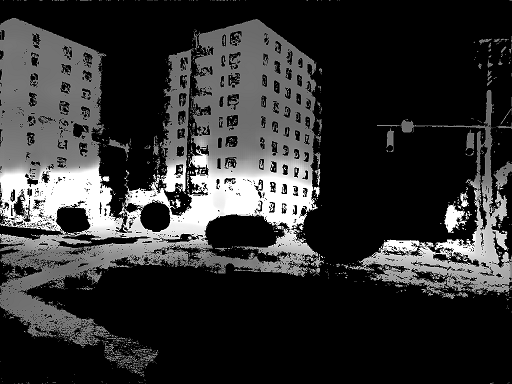}\\
			\includegraphics[width=\linewidth]{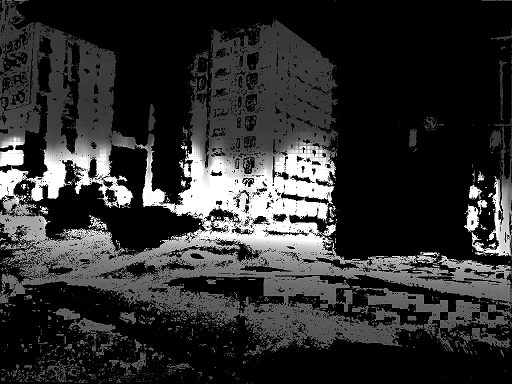}\\
			\begin{center}
             \tiny BTS\cite{lee2019big}\\
Supervised \\
\end{center}
		\end{minipage}%
		\begin{minipage}[t]{0.084\linewidth}
			\centering
			\includegraphics[width=\linewidth]{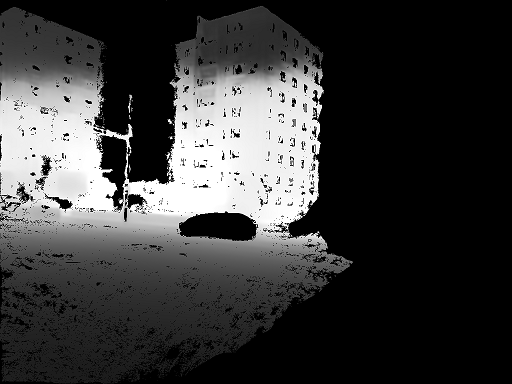}\\
			\includegraphics[width=\linewidth]{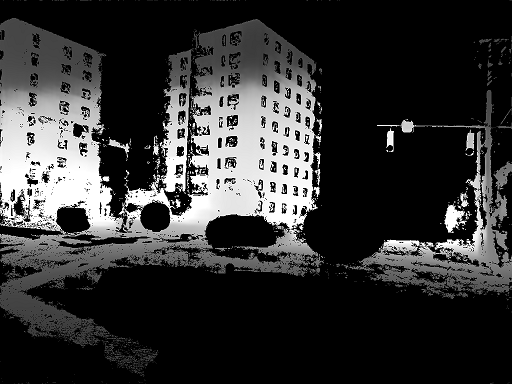}\\
			\includegraphics[width=\linewidth]{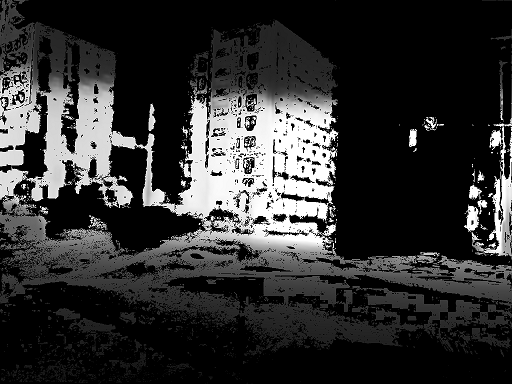}\\
			\begin{center}
             \tiny VNL\cite{yin2019enforcing}\\
Supervised \\
\end{center}
			
		\end{minipage}%
		\begin{minipage}[t]{0.084\linewidth}
			\centering
			\includegraphics[width=\linewidth]{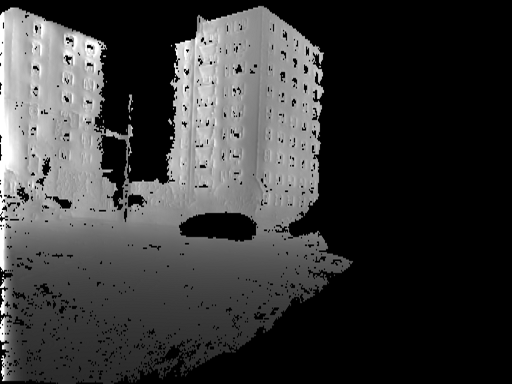}\\
			\includegraphics[width=\linewidth]{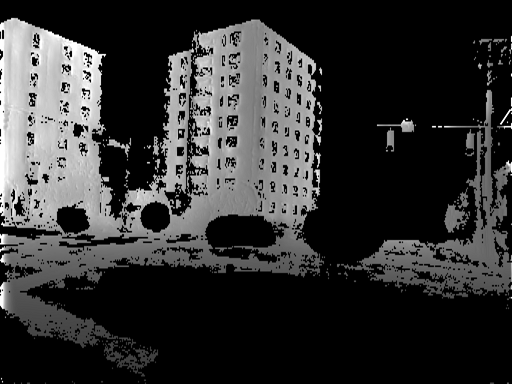}\\
			\includegraphics[width=\linewidth]{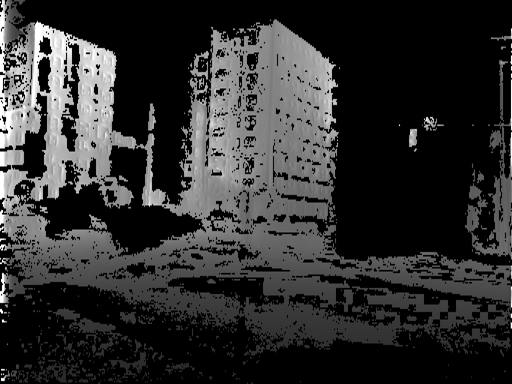}\\
			\begin{center}
             \tiny {Monodepth}\\
                S-Sup-S \cite{godard2017unsupervised}\\
\end{center}
		\end{minipage}%
		\begin{minipage}[t]{0.084\linewidth}
			\centering
			\includegraphics[width=\linewidth]{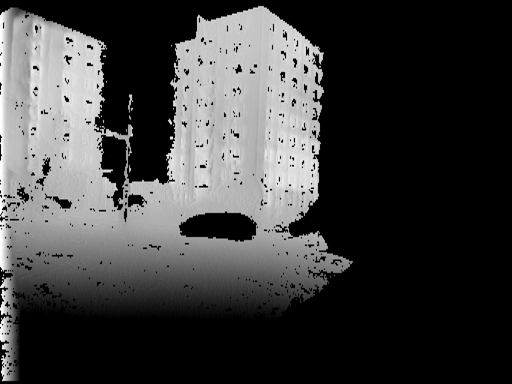}\\
			\includegraphics[width=\linewidth]{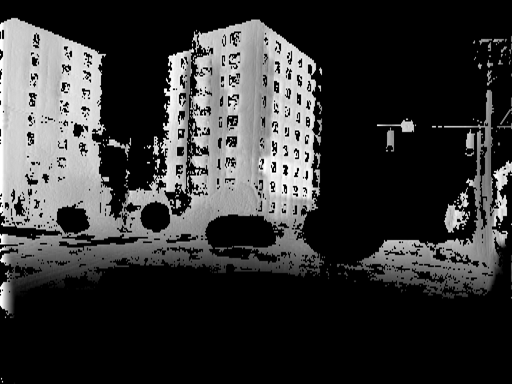}\\
			\includegraphics[width=\linewidth]{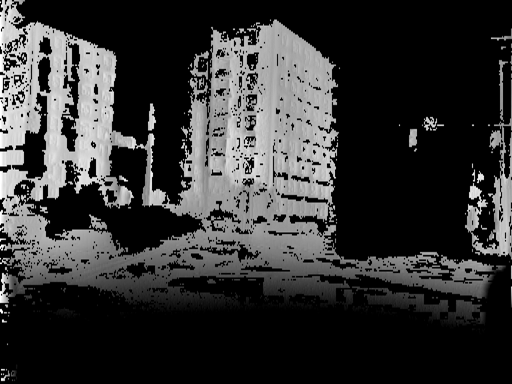}\\
			\begin{center}
             \tiny {adareg}\cite{wong2019bilateral}\\
S-Sup-S \\
\end{center}
		\end{minipage}%
		\begin{minipage}[t]{0.084\linewidth}
			\centering
			\includegraphics[width=\linewidth]{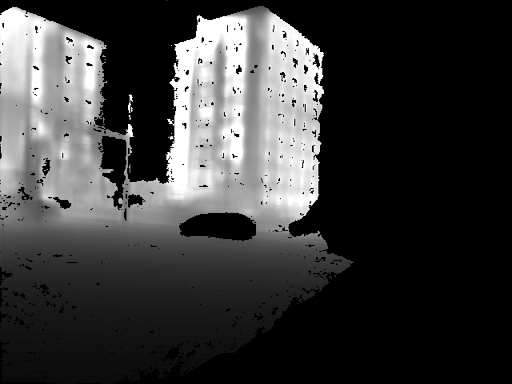}\\
			\includegraphics[width=\linewidth]{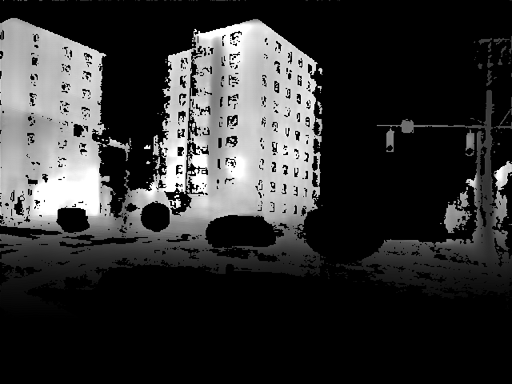}\\
			\includegraphics[width=\linewidth]{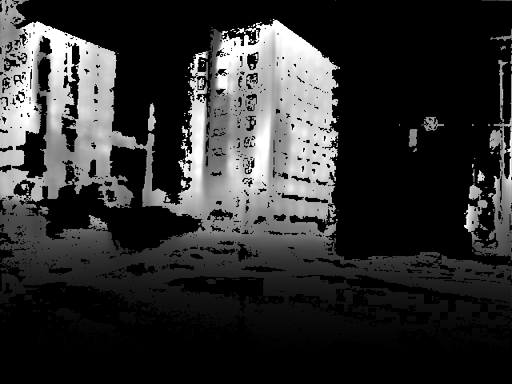}\\
			\begin{center}
             \tiny PackNet\cite{guizilini20203d}\\
S-Sup-M \\
\end{center}
		\end{minipage}%
		\begin{minipage}[t]{0.084\linewidth}
			\centering
			\includegraphics[width=\linewidth]{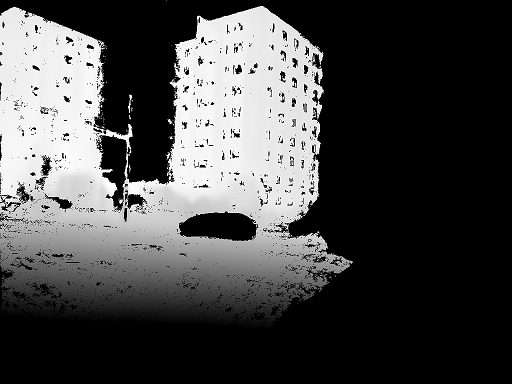}\\
			\includegraphics[width=\linewidth]{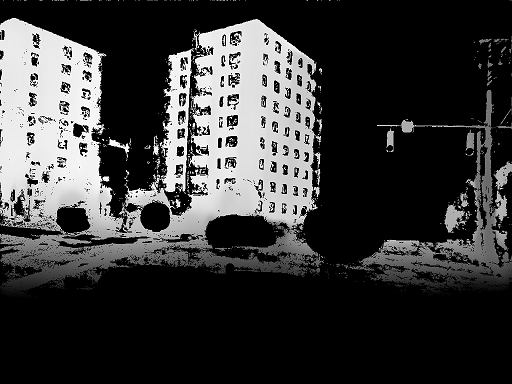}\\
			\includegraphics[width=\linewidth]{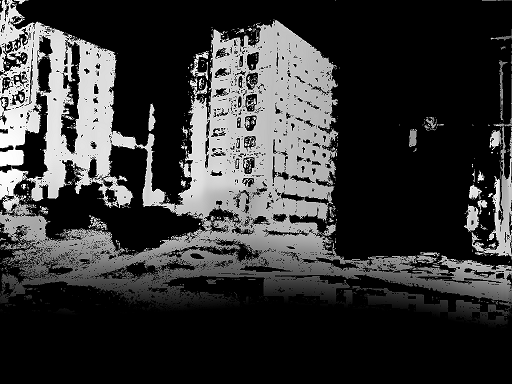}\\
			\begin{center}
        \tiny
         {Monodepth2}\\
S-Sup-M \cite{godard2019digging}\\
\end{center}
		\end{minipage}%
		\begin{minipage}[t]{0.084\linewidth}
			\centering
			\includegraphics[width=\linewidth]{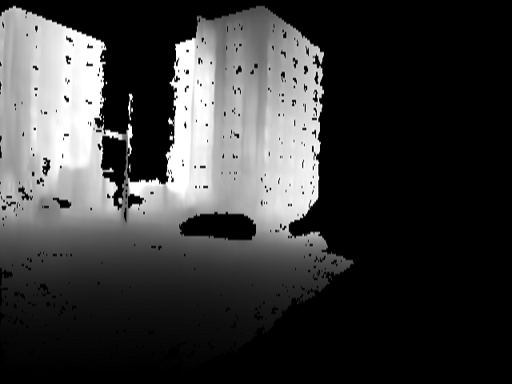}\\
			\includegraphics[width=\linewidth]{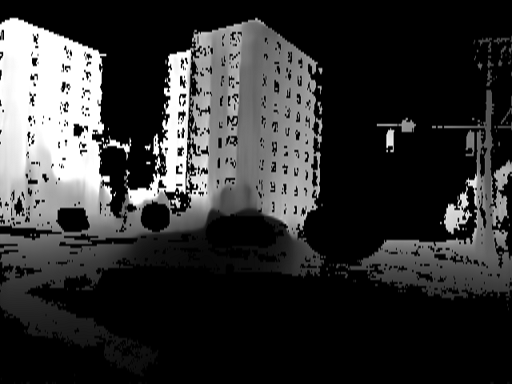}\\
			\includegraphics[width=\linewidth]{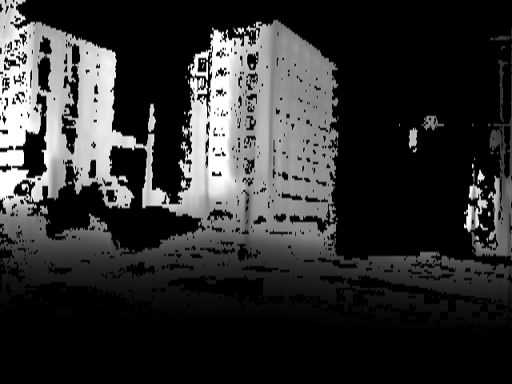}\\
			\begin{center}
             \tiny SGDepth\cite{klingner2020self}\\
S-Sup-M \\
\end{center}
		\end{minipage}%
		\begin{minipage}[t]{0.084\linewidth}
			\centering
			\includegraphics[width=\linewidth]{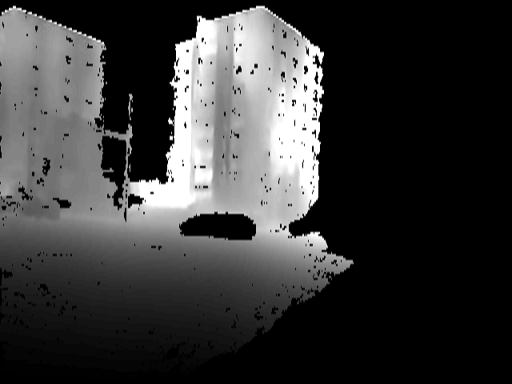}\\
			\includegraphics[width=\linewidth]{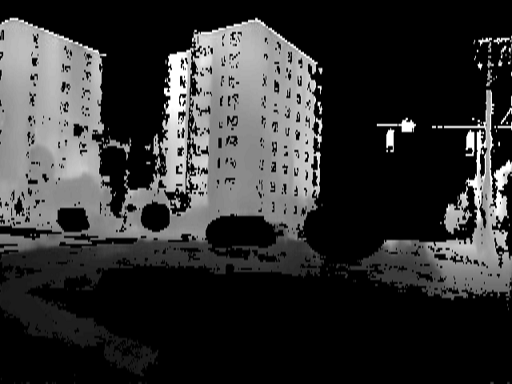}\\
			\includegraphics[width=\linewidth]{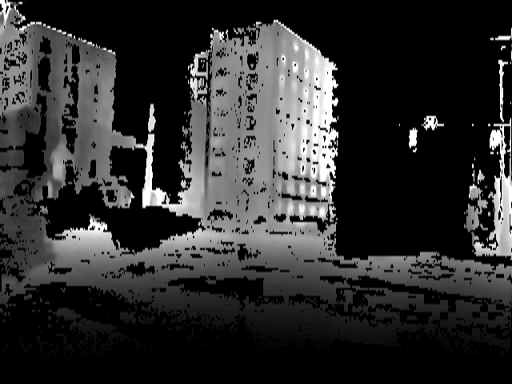}\\
			\begin{center}
             \tiny T2Net\cite{zheng2018t2net}\\
Syn-to-Real\\
\end{center}
		\end{minipage}%
		\begin{minipage}[t]{0.084\linewidth}
			\centering
			\includegraphics[width=\linewidth]{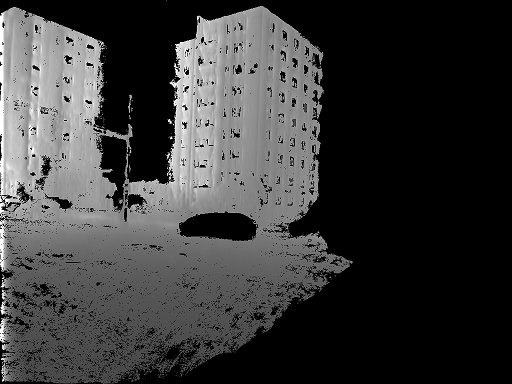}\\
			\includegraphics[width=\linewidth]{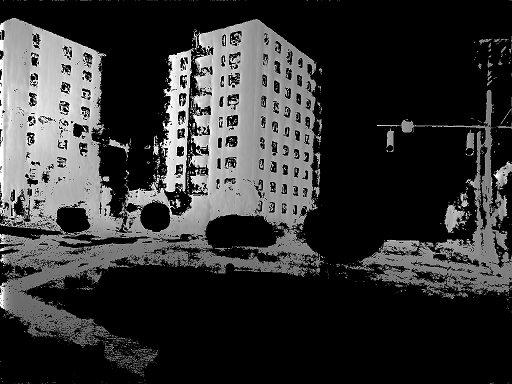}\\
			\includegraphics[width=\linewidth]{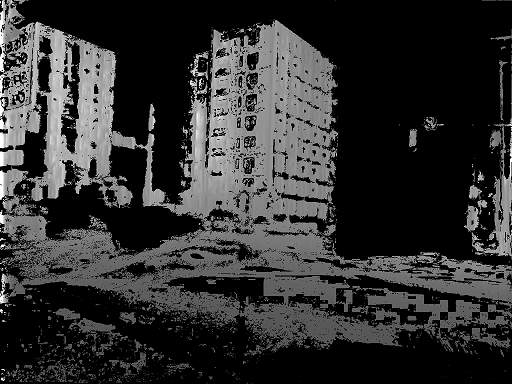}\\
			\begin{center}
             \tiny            
     {GASDA}\cite{zheng2018t2net}\\
Syn-to-Real\\
\end{center}
		\end{minipage}%
	\caption{\textcolor{revision}{Comparison among supervised, self-supervised stereo based (S-Sup-S), self-supervised monocular video based (S-Sup-M) and domain adaptation (Syn-to-Real) methods.}
	}
	\label{vis_diff_methods}
\end{figure*}

\begin{figure*}[]
	\centering
		\begin{minipage}[t]{0.49\linewidth}
			\centering
			\includegraphics[width=\linewidth]{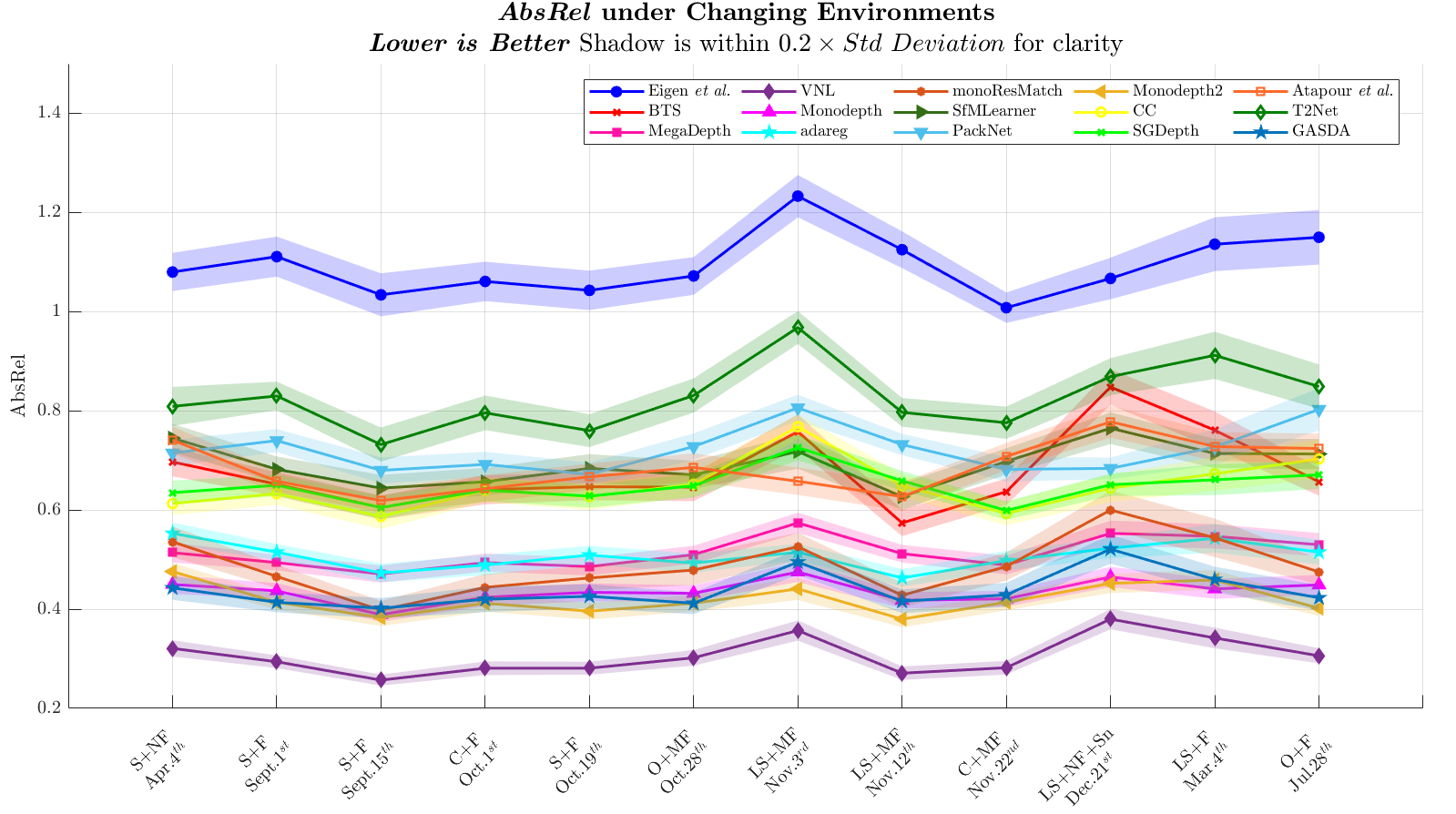}
		\end{minipage}%
		\begin{minipage}[t]{0.49\linewidth}
			\centering
			\includegraphics[width=\linewidth]{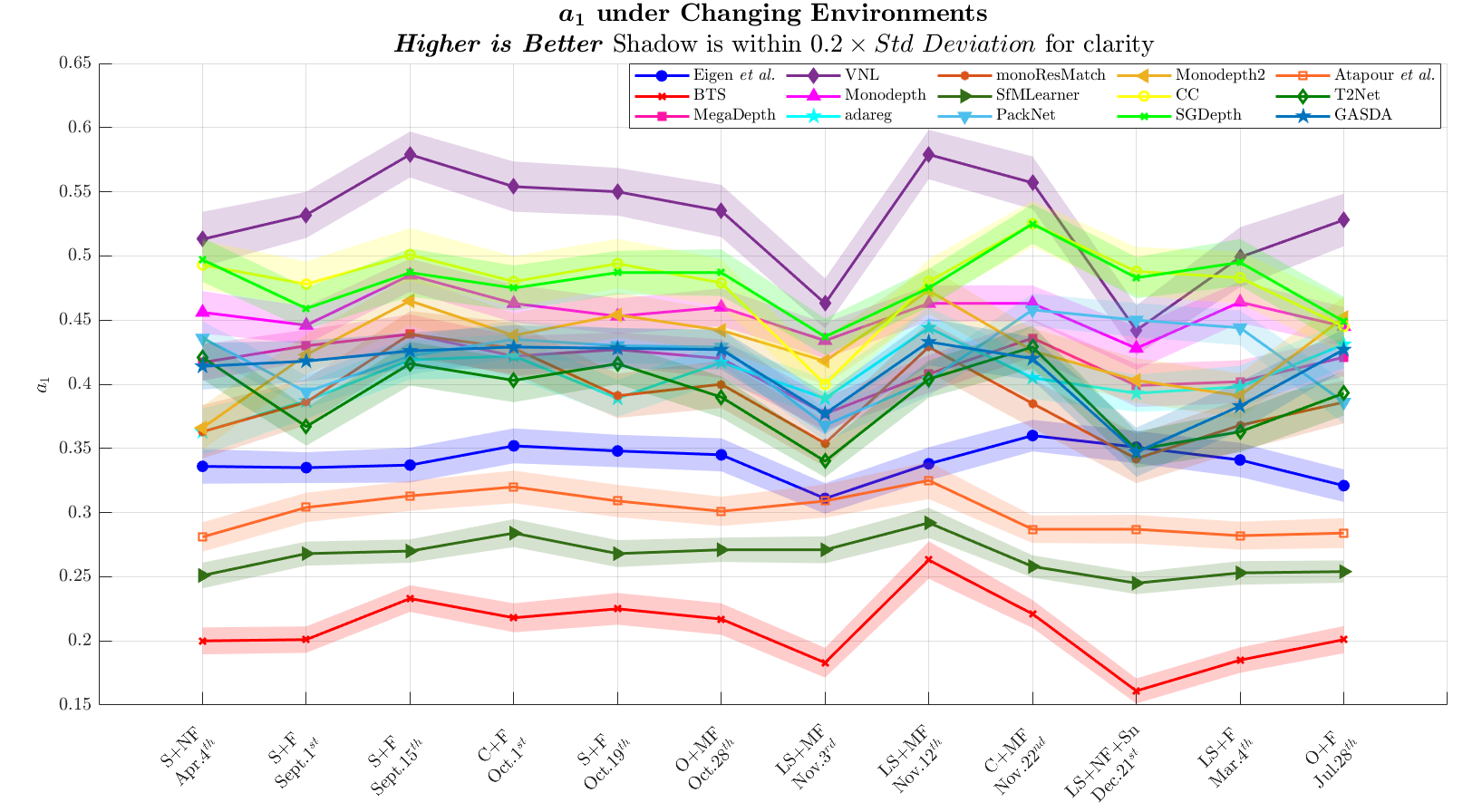}
		\end{minipage}%
	\centering
	\caption{Results on \textit{SeasonDepth} dataset under 12 different environments with dates. The shadows indicate error bars around mean values with $0.2\times Standard$ $ Deviation $ for more clarity.}
	\label{diagrams_absrelanda1}
\end{figure*}

\begin{table}[]
		\caption{Influence of fine-tuning from \textit{KITTI} to \textit{SeasonDepth} \\(\color{red}\bm{$\downarrow$}\color{black}: Lower Better, \color{blue}\bm{$\uparrow$}\color{black}: Higher Better)}
		\label{finetune_evalation_results}
		\centering
		\resizebox{0.5\textwidth}{!}{
	\begin{tabular}{ccccccc}
		\toprule
		\multicolumn{1}{c}{}                                                                                                                       & \multicolumn{2}{c}{\textbf{Average}}                        & \multicolumn{2}{c}{ \textbf{Variance}$ (10^{-2}) $}                        & \multicolumn{2}{c}{ \textbf{Relative Range}}                 \\
\multirow{-2}{*}{Method}
	   & \cellcolor[HTML]{FFCCC9}$ AbsRel $ \color{red}\bm{$\downarrow$} & \cellcolor[HTML]{CBCEFB}$ a_1 $ \color{blue}\bm{$\uparrow$}    & \cellcolor[HTML]{FFCCC9}$ AbsRel  $ \color{red}\bm{$\downarrow$} & \cellcolor[HTML]{FFCCC9}$ a_1 $ \color{red}\bm{$\downarrow$}     & \cellcolor[HTML]{FFCCC9}$ AbsRel  $ \color{red}\bm{$\downarrow$} & \cellcolor[HTML]{FFCCC9}$ 1-a_1 $ \color{red}\bm{$\downarrow$}    \\ \midrule
		  BTS \cite{lee2019big}                                   & 0.676                          & 0.209                         & 0.545                         & 0.0650                         & {0.405}                          & 0.129                        \\
		\makecell[c]{BTS \cite{lee2019big} \\ (fine-tuned)}                                        & {0.339} &	0.479 &	{0.0425} & {0.0389}	&	{0.203} &	{0.117}                       \\ \midrule
  SfMLearner \cite{zhou2017unsupervised}                                                  &{0.360}                          & {0.495}                       & 0.0801                         & 0.0628                         & 0.269                         & 0.182                         \\
		 \makecell[c]{SfMLearner \cite{zhou2017unsupervised}   \\ (fine-tuned)  }                           & 0.413 &	0.440 &	{0.0502} &	{0.0290}&	{0.178} &	{0.100}        \\
		\bottomrule
	\end{tabular}
}
\end{table}

\subsection{Cross-dataset Generalization Results}

\label{comparison_results}

In this section, we show the generalization performance from \textit{KITTI} to \textit{SeasonDepth} in Tab. \ref{evalation_results}. First we can see that in the zero-shot cross-dataset generalization setting, self-supervised methods show more robustness to different environments than supervised ones, which suffer from large values of $Variance$ and $Relative Range$ and more sensitive.  Also, the gap between \textit{KITTI} results and \textit{SeasonDepth} $Average$ results is clear, showing that the generalization without fine-tuning is challenging especially in different environments. 
Interestingly,  supervised  methods with good $ Variance $ performance are not consistent with those with good $ Average $ performance, which indicates that algorithms tend to work well in specific environments instead of being robust to all conditions, validating the significance of the cross-environment study with \textit{SeasonDepth} dataset.

\begin{figure}[htbp]
	\centering
		\begin{minipage}[t]{0.19\linewidth}
			\centering
			\includegraphics[width=\linewidth]{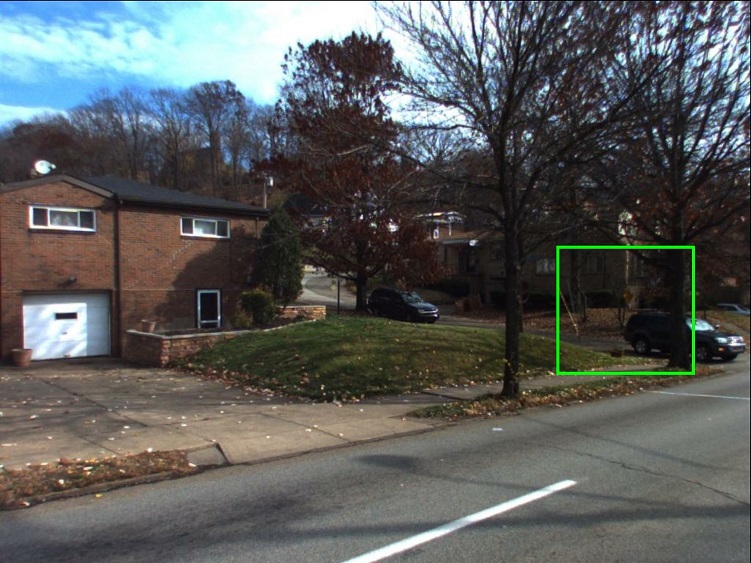}\\
			\includegraphics[width=\linewidth]{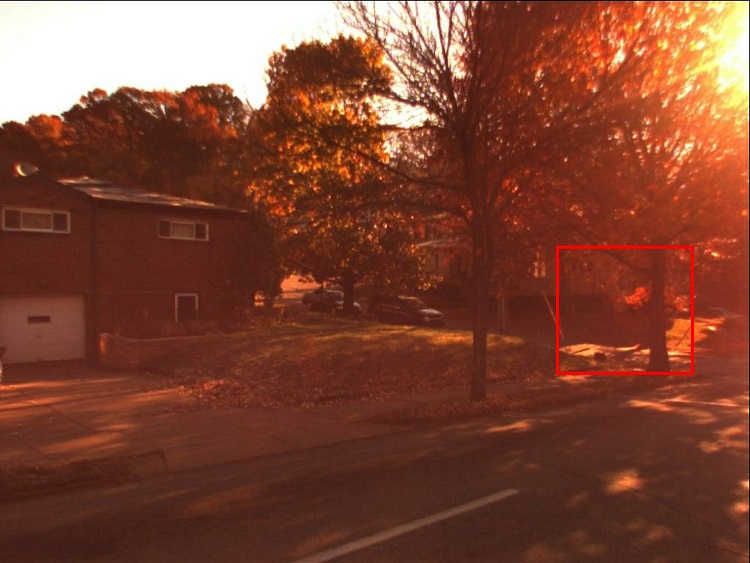}\\
			\vspace{0.05cm}
			\includegraphics[width=\linewidth]{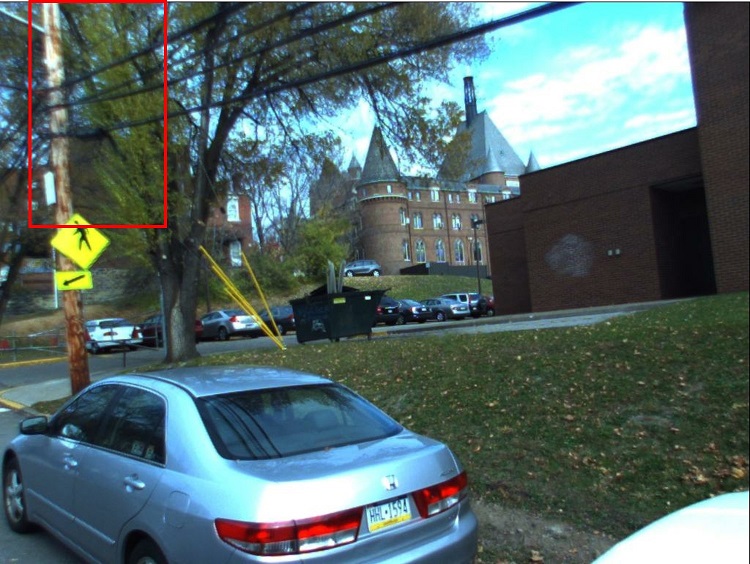}\\
			\includegraphics[width=\linewidth]{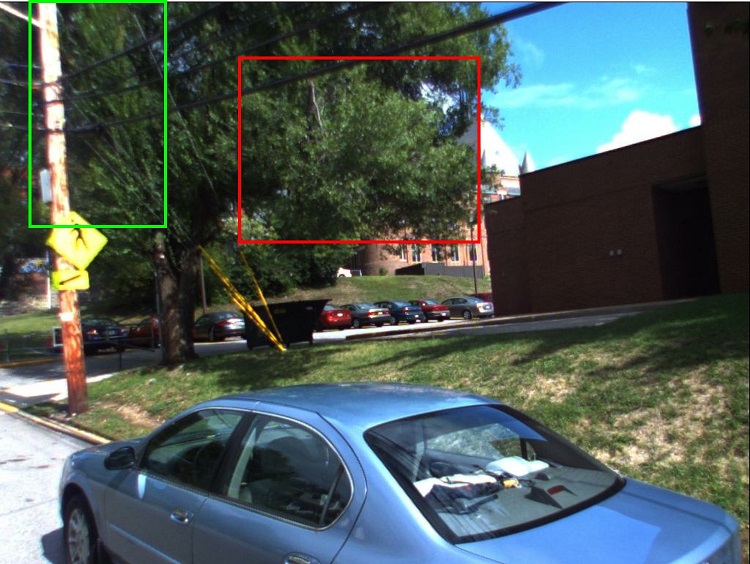}\\
			\centerline{\scriptsize RGB}
		\end{minipage}%
		\begin{minipage}[t]{0.19\linewidth}
			\centering
			\includegraphics[width=\linewidth]{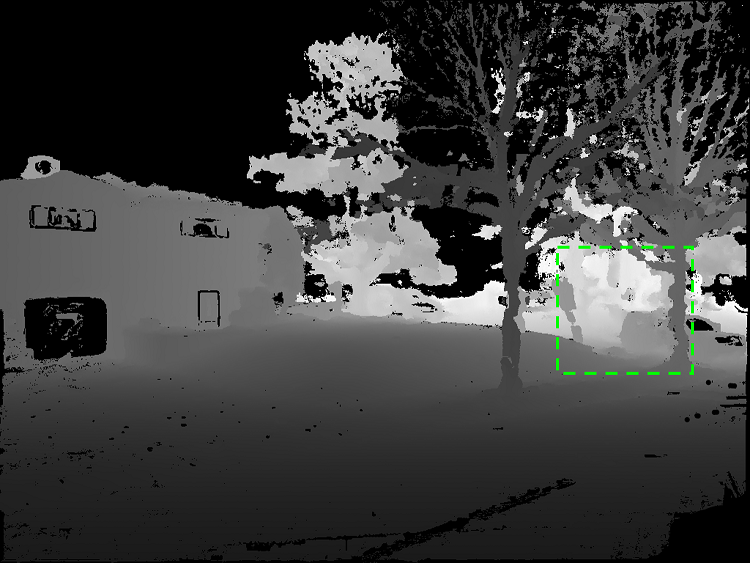}\\
			\includegraphics[width=\linewidth]{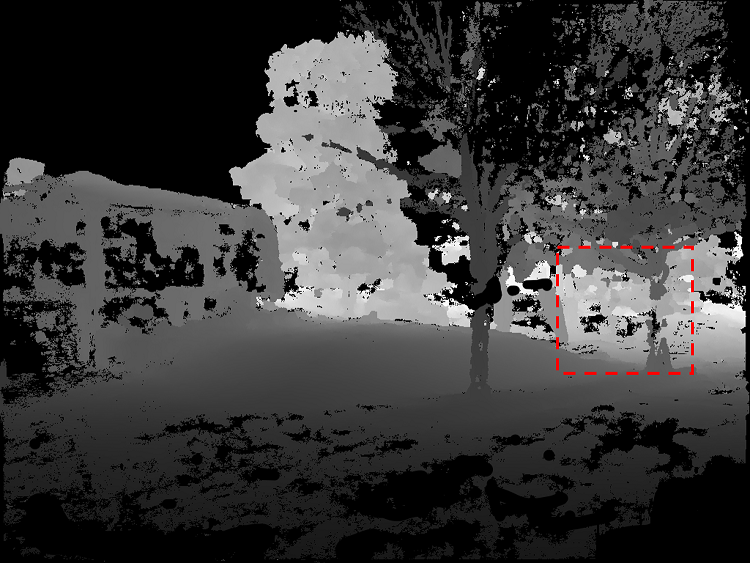}\\
			\vspace{0.05cm}
			\includegraphics[width=\linewidth]{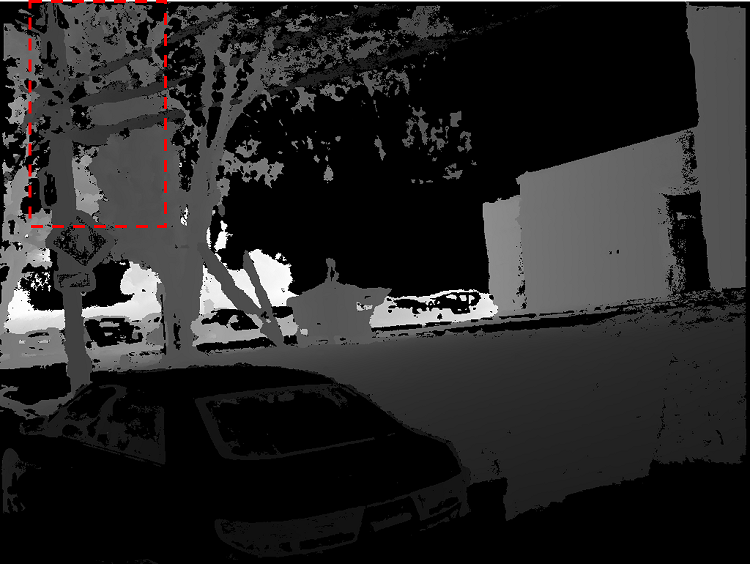}\\
			\includegraphics[width=\linewidth]{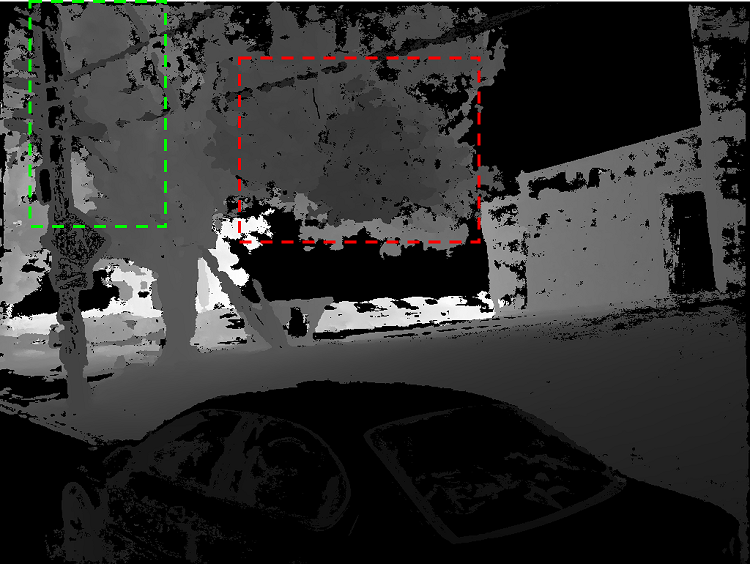}\\
			\centerline{\scriptsize Ground  Truth}
		\end{minipage}%
		\begin{minipage}[t]{0.19\linewidth}
			\centering
			\includegraphics[width=\linewidth]{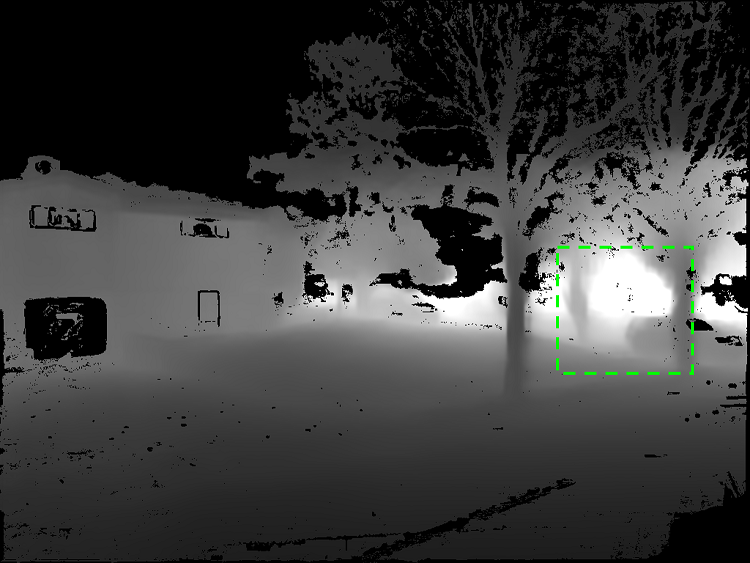}\\
			\includegraphics[width=\linewidth]{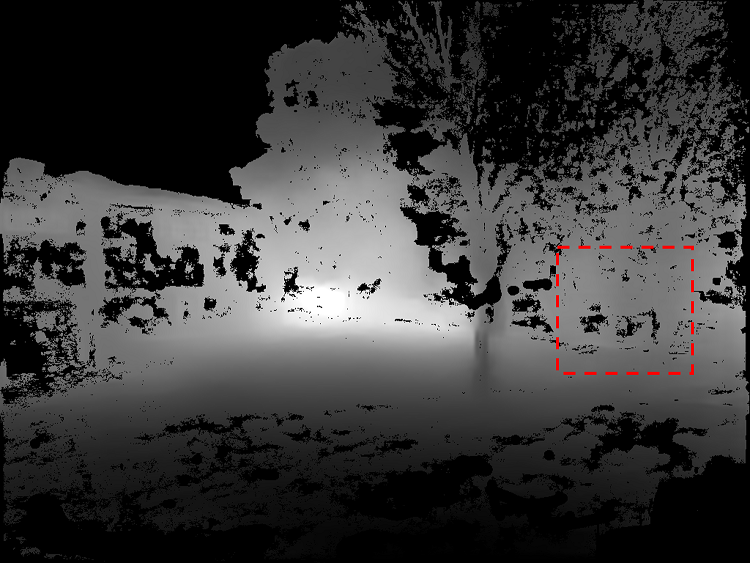}\\
			\vspace{0.05cm}
			\includegraphics[width=\linewidth]{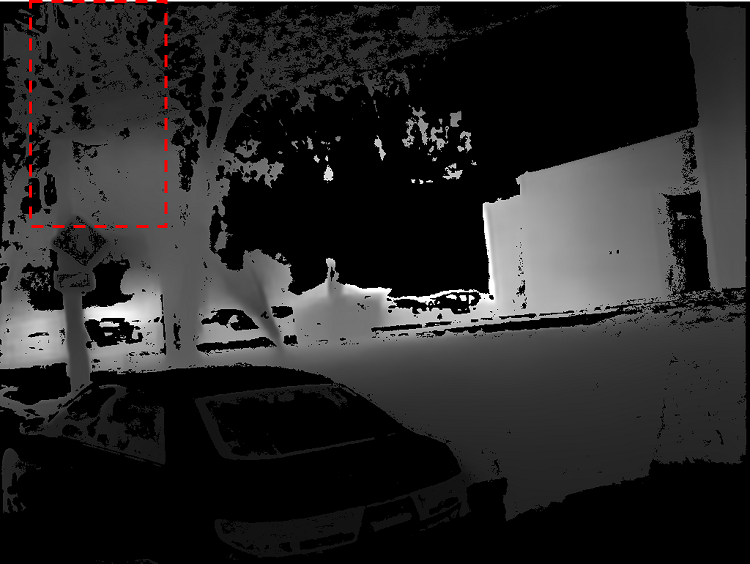}\\
			\includegraphics[width=\linewidth]{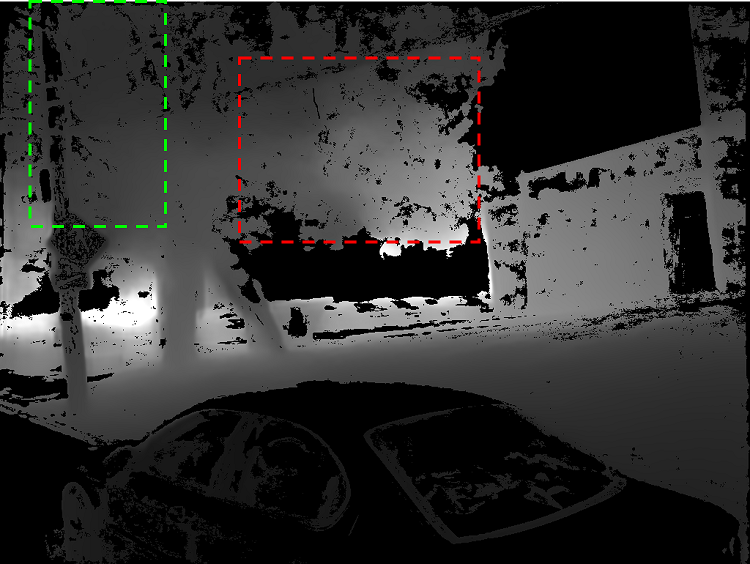}\\
			\centerline{\scriptsize  \textit{VNL}  \cite{yin2019enforcing}}
		\end{minipage}%
		\begin{minipage}[t]{0.19\linewidth}
			\centering
			\includegraphics[width=\linewidth]{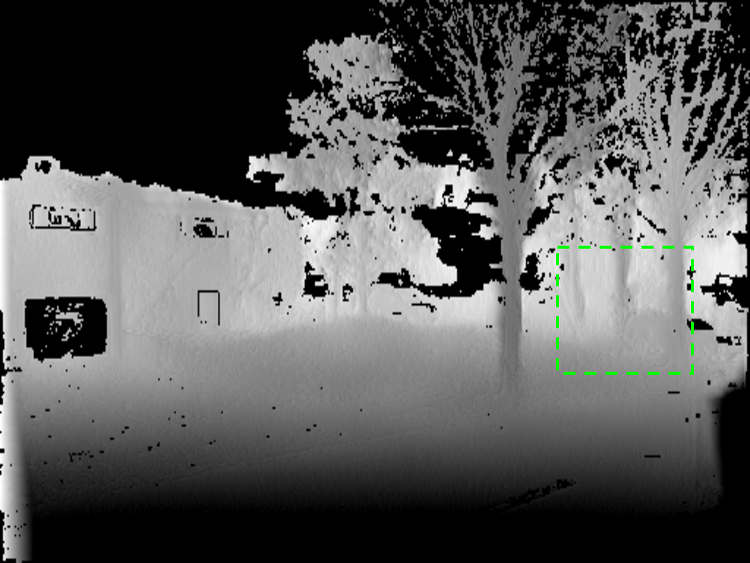}\\
			\includegraphics[width=\linewidth]{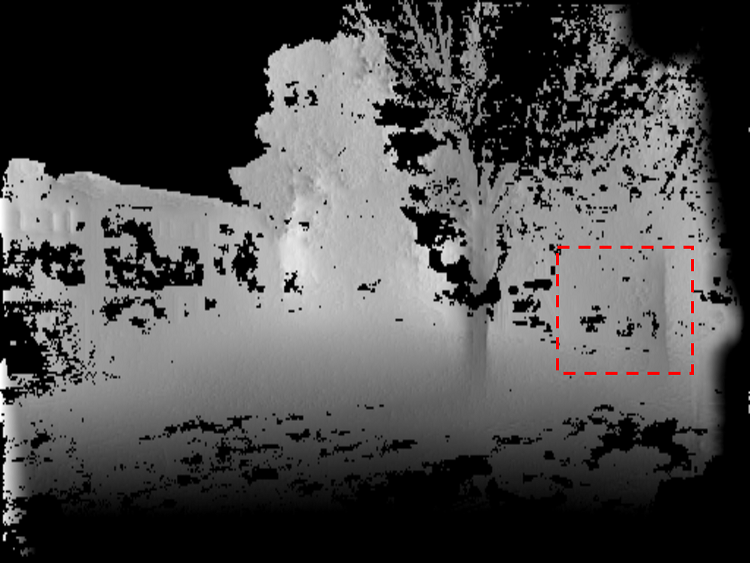}\\
			\vspace{0.05cm}
			\includegraphics[width=\linewidth]{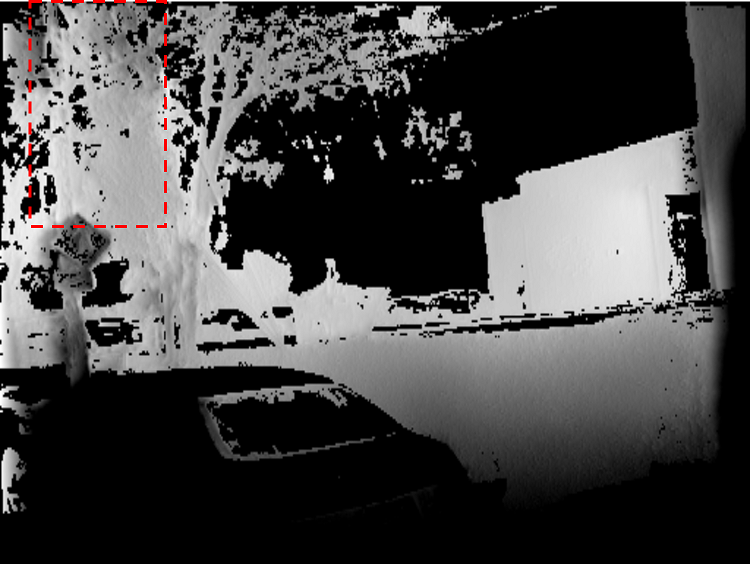}\\
			\includegraphics[width=\linewidth]{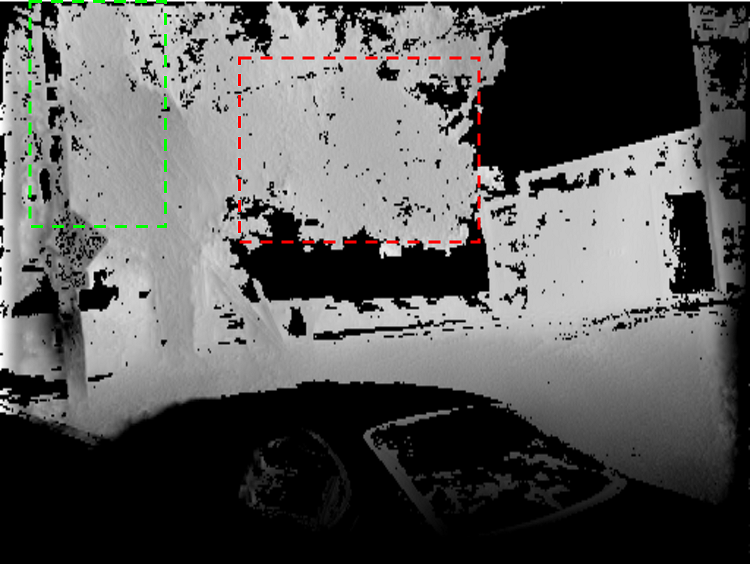}\\
			\centerline{\scriptsize \textit{adareg} \cite{wong2019bilateral}}
		\end{minipage}%
		\begin{minipage}[t]{0.19\linewidth}
			\centering
			\includegraphics[width=\linewidth]{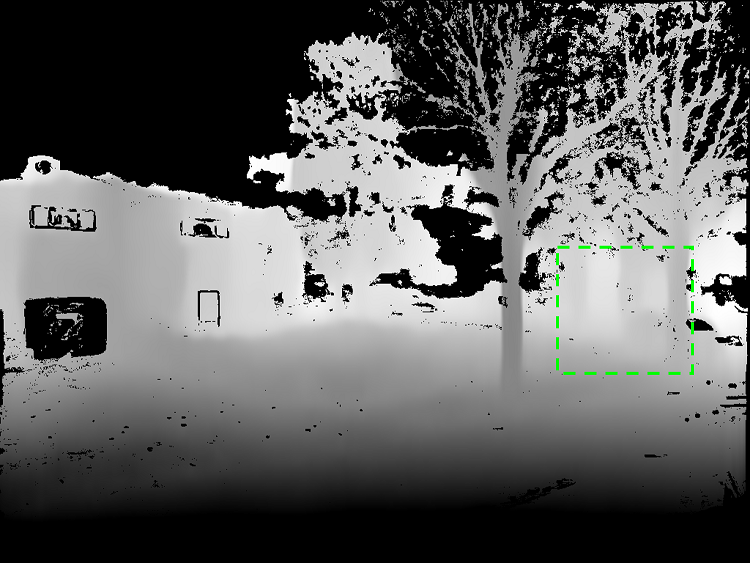}\\
			\includegraphics[width=\linewidth]{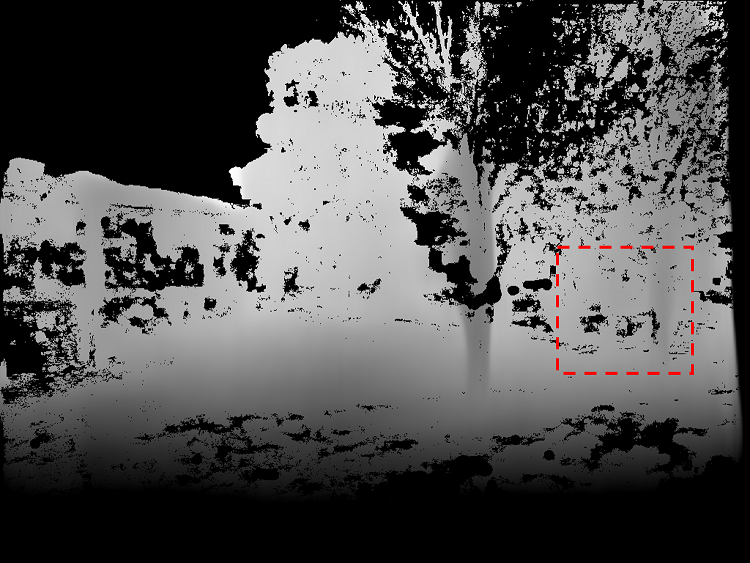}\\
			\vspace{0.05cm}
			\includegraphics[width=\linewidth]{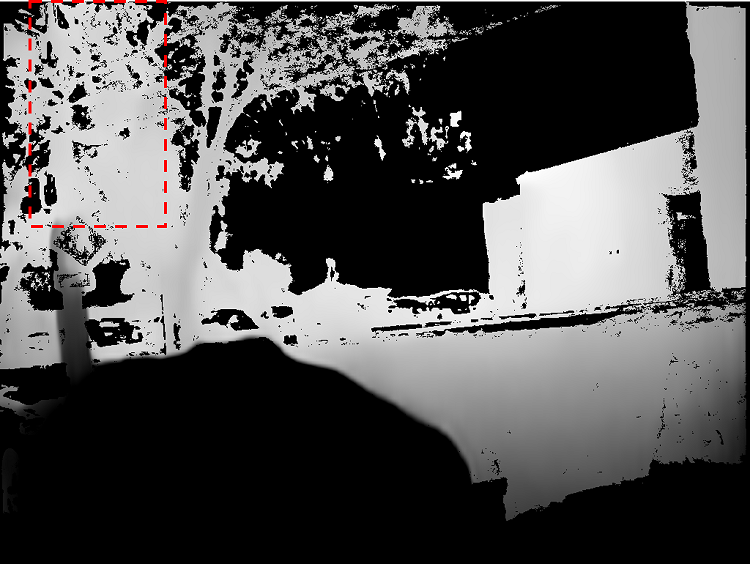}\\
			\includegraphics[width=\linewidth]{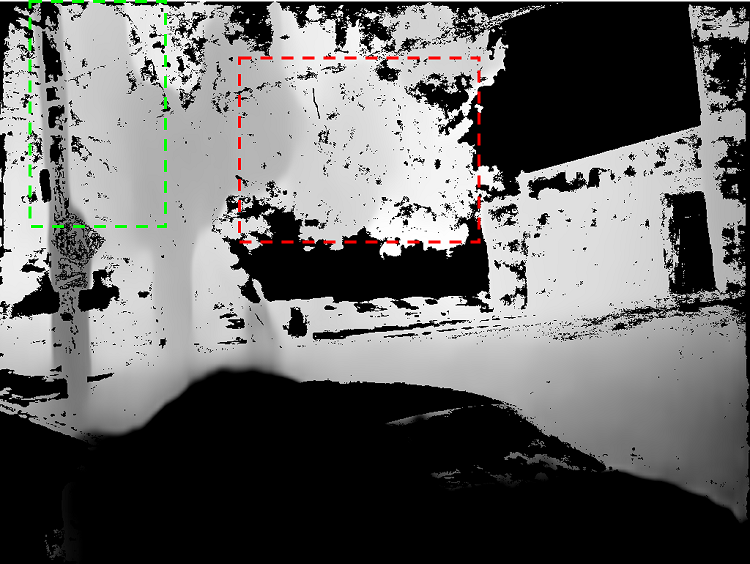}\\
			\centerline{\scriptsize \textit{Monodepth2} \cite{godard2019digging}}
		\end{minipage}%
	\caption{Qualitative comparison results with illumination or vegetation changes. The conditions from top to down are \textit{C+MF, Nov. $ 22^{nd} $},  \textit{LS+MF, Nov. $ 3^{rd} $}, \textit{C+MF, Nov. $ 22^{nd} $} and \textit{C+F, Oct. $ 1^{st} $}. Green blocks indicate good performance while red blocks are for bad results.
	}
	\label{benchmark_results}
\end{figure}	

Then we investigate the fine-tuned models from \textit{KITTI} to \textit{SeasonDepth} and compare it with generalization without fine-tuning in Tab. \ref{finetune_evalation_results}. It can be seen that although most metrics are improved through fine-tuning, the improvement is still limited compared to other zero-shot evaluation results in Tab. \ref{evalation_results}, indicating that solely increasing the variability of training data cannot address the challenge of environmental changes. Qualitative results for different types of baselines are shown in Fig. \ref{vis_diff_methods}. It can be seen that supervised methods \textit{BTS} \cite{lee2019big} and \textit{VNL} \cite{yin2019enforcing} suffer from overfitting through the predicted pattern where the top and bottom areas are dark while the central areas are light, even for buildings. 

\begin{table*}[]
    \centering
    \caption{Cross-dataset performance on \textit{KITTI}  \cite{Uhrig2017THREEDV} with models  fine-tuned on \textit{SeasonDepth} and \textit{Cityscapes}
    \cite{cordts2016cityscapes}. } 
    	\resizebox{\textwidth}{!}{
    \begin{tabular}{cccccc}
    \hline
        Method & MAE \color{red}\bm{$\downarrow$}\color{black} & absErrorRel \color{red}\bm{$\downarrow$}\color{black} & iMAE \color{red}\bm{$\downarrow$}\color{black} & iRMSE \color{red}\bm{$\downarrow$}\color{black} & sqErrorRel \color{red}\bm{$\downarrow$}\color{black} \\ \hline
        BTS \cite{lee2019big} tuned on Cityscapes \cite{cordts2016cityscapes} & \textbf{4.21}±0.411 & \textbf{0.29}±0.030 & 0.032±0.003 & 0.048±0.005 & 0.20±0.051\\ 
        BTS \cite{lee2019big} tuned on SeasonDepth (ous)  & 5.36±0.200& 0.32±0.019 & \textbf{0.030}±0.004 & \textbf{0.037}±0.005 & \textbf{0.19}±0.022 \\ \hline
       SfMLearner \cite{zhou2017unsupervised} tuned on Cityscapes \cite{cordts2016cityscapes} & 6.40±0.202 & 0.42±0.019 & 0.045±0.003 & 0.060±0.004 & 0.38±0.05 \\ 
        SfMLearner \cite{zhou2017unsupervised} tuned on SeasonDepth (ous)  & \textbf{6.31}±0.270 & \textbf{0.38}±0.023 & \textbf{0.032}±0.003 & \textbf{0.041}±0.003 & \textbf{0.30}±0.0338 \\ \hline
    \end{tabular}}
	\label{justification_quan2}
\end{table*}

 \begin{figure*}[htbp]
	\centering
		\begin{minipage}[t]{0.16\linewidth}
			\centering
			\includegraphics[width=\linewidth]{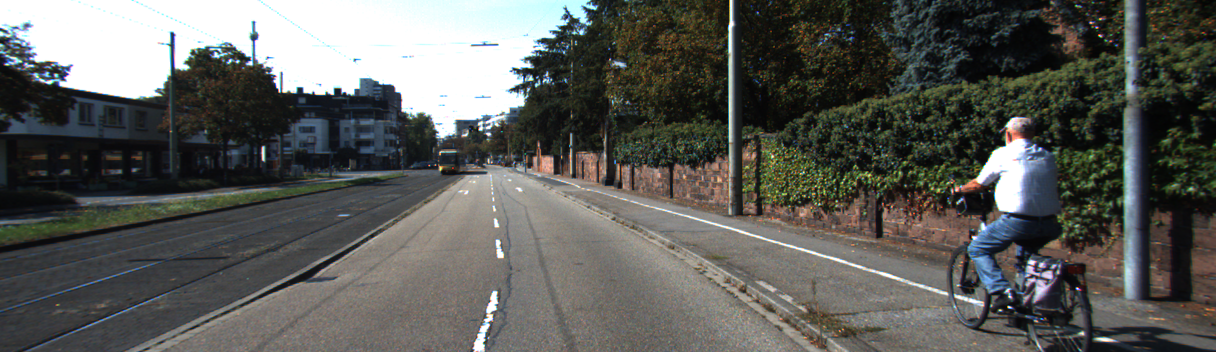}\\
			\vspace{0.05cm}
			\includegraphics[width=\linewidth]{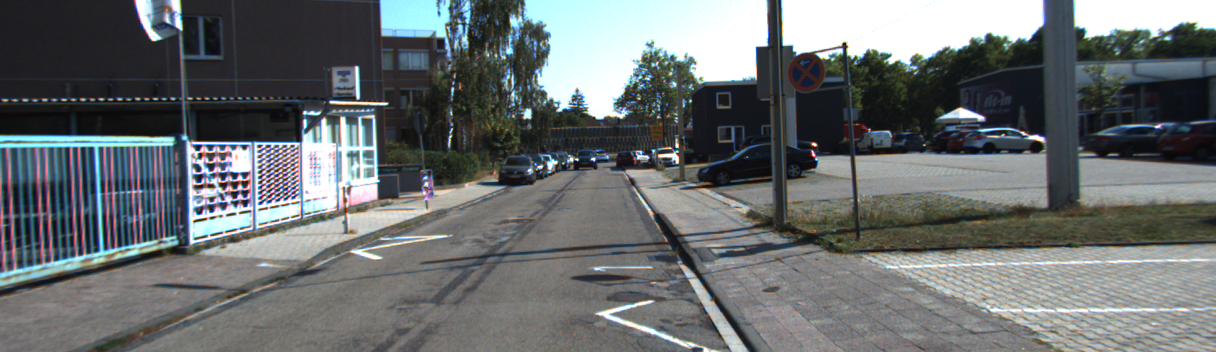}\\
			\vspace{0.05cm}
			\includegraphics[width=\linewidth]{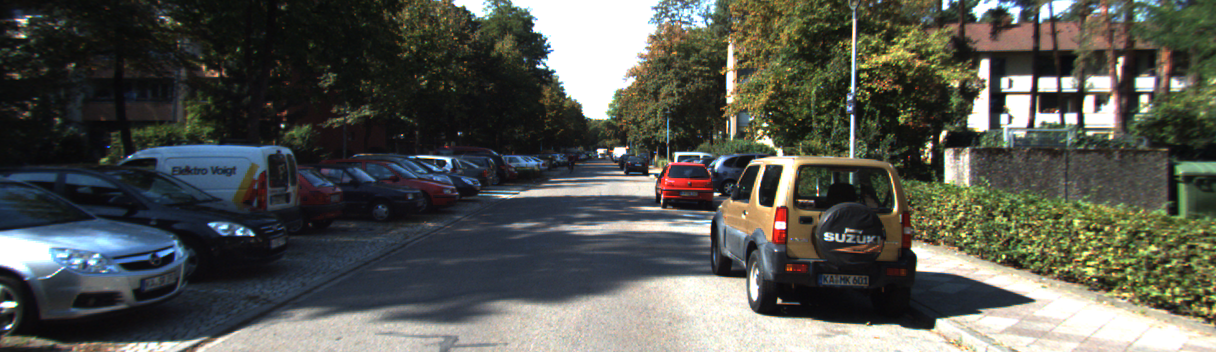}\\
			\vspace{0.05cm}
			\centerline{\scriptsize RGB}
		\end{minipage}%
		\begin{minipage}[t]{0.16\linewidth}
			\centering
			\includegraphics[width=\linewidth]{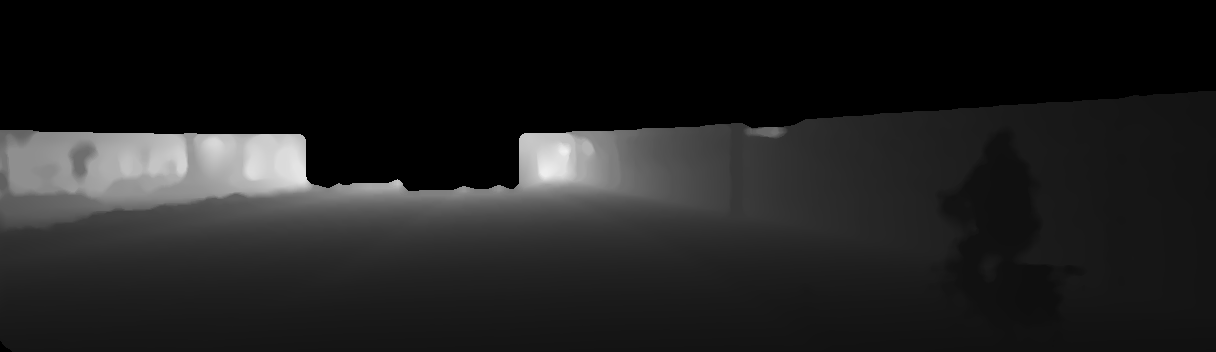}\\
			\vspace{0.05cm}
			\includegraphics[width=\linewidth]{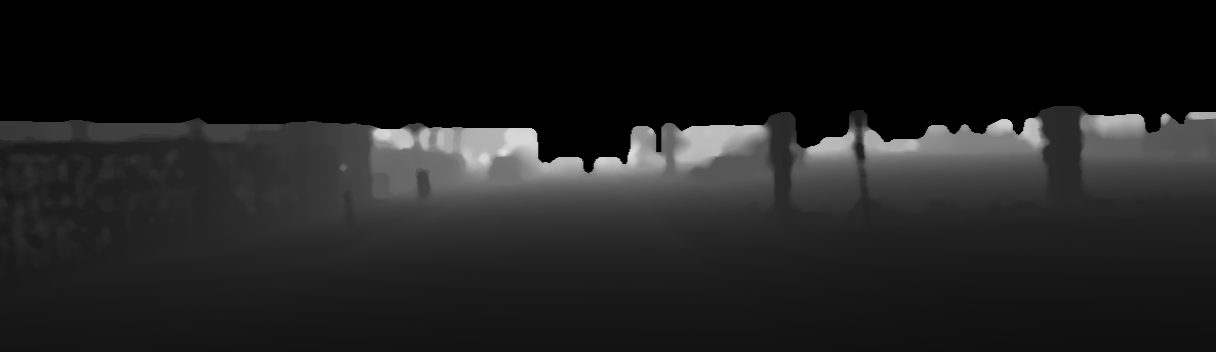}\\
			\vspace{0.05cm}
			\includegraphics[width=\linewidth]{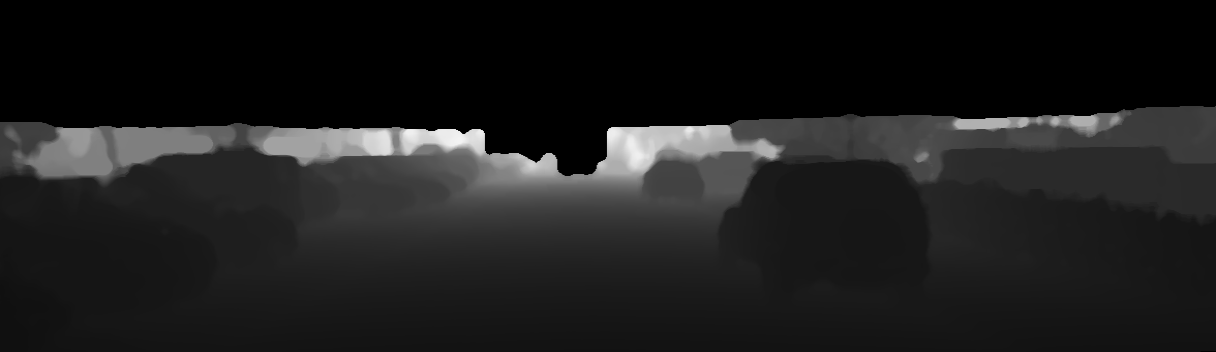}\\
			\vspace{0.05cm}
			\centerline{ \scriptsize Ground Truth}
		\end{minipage}%
		\begin{minipage}[t]{0.16\linewidth}
			\centering
			\includegraphics[width=\linewidth]{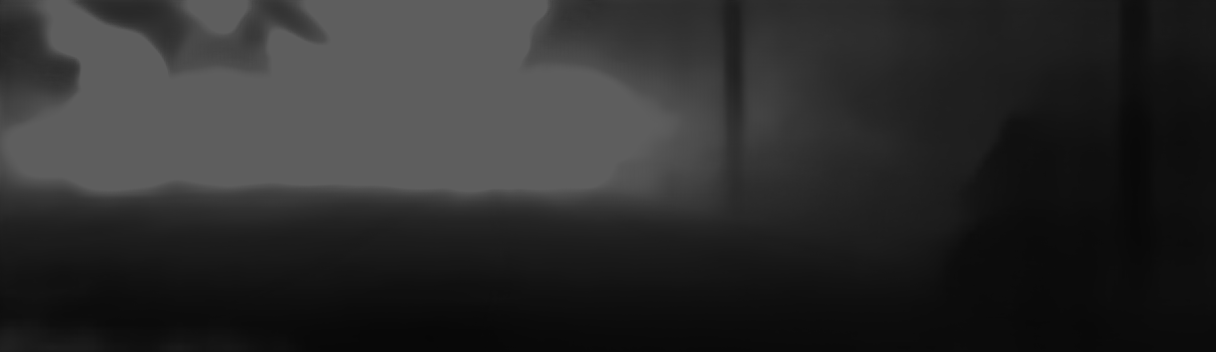}\\
			\vspace{0.05cm}
			\includegraphics[width=\linewidth]{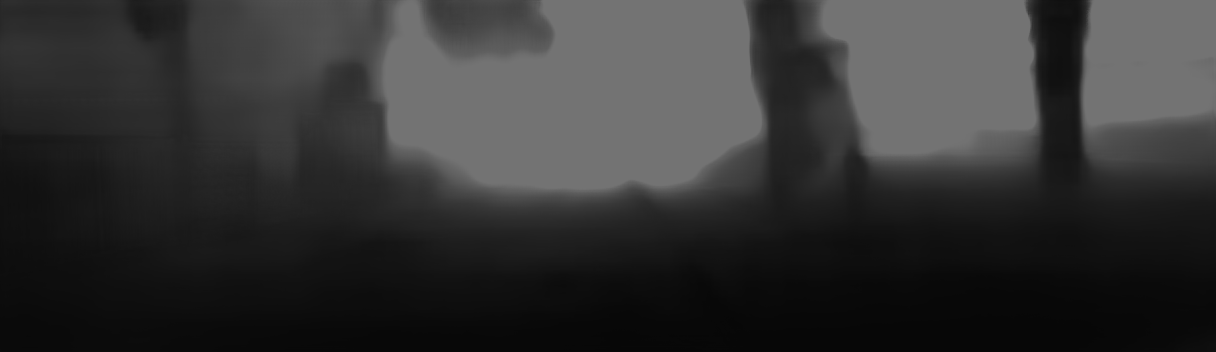}\\
			\vspace{0.05cm}
			\includegraphics[width=\linewidth]{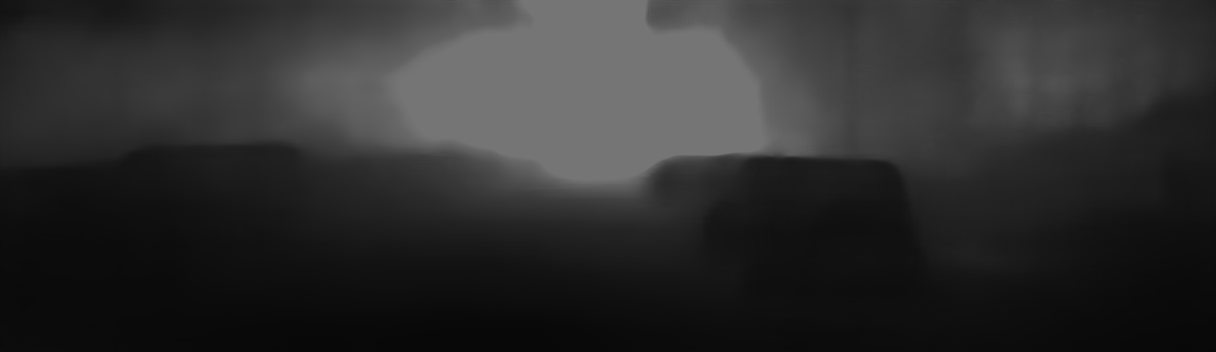}\\
			\vspace{-0.05cm}
            \begin{center}
            \scriptsize \textit{BTS \cite{lee2019big}} fine-tuned \\ on \textit{Cityscapes \cite{cordts2016cityscapes}}
\end{center}
		\end{minipage}%
		\begin{minipage}[t]{0.16\linewidth}
			\centering
			\includegraphics[width=\linewidth]{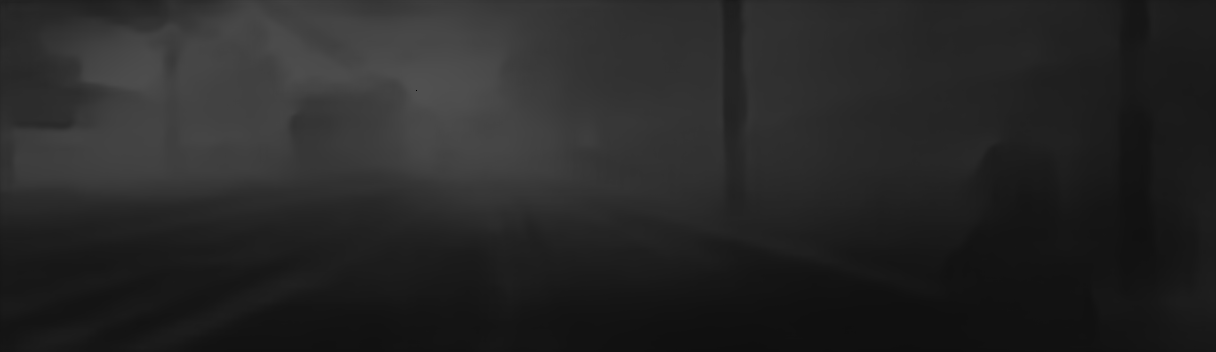}\\
			\vspace{0.05cm}
			\includegraphics[width=\linewidth]{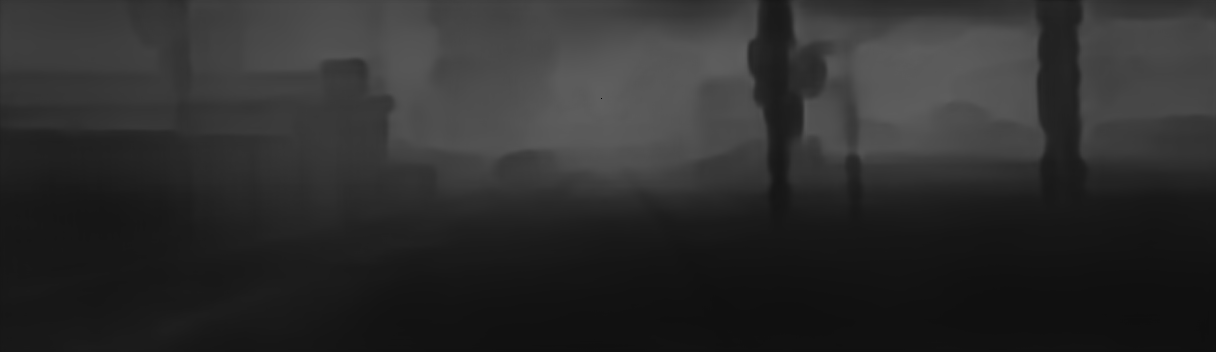}\\
			\vspace{0.05cm}
			\includegraphics[width=\linewidth]{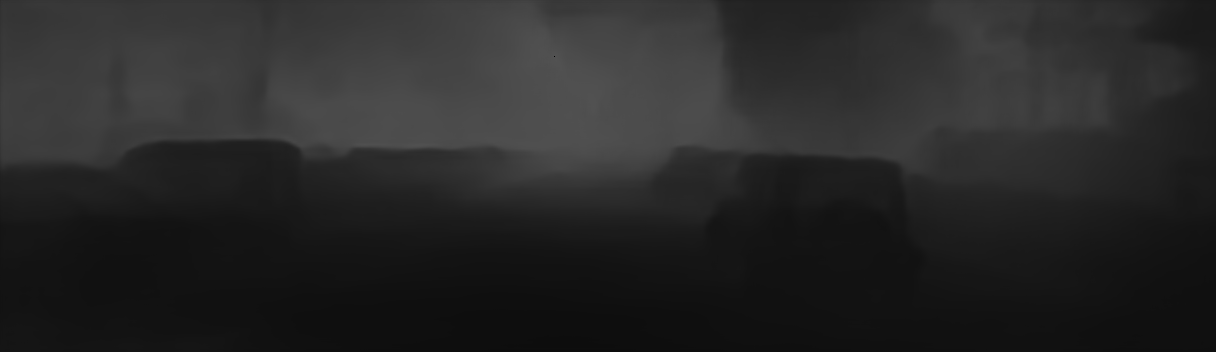}\\
			\vspace{-0.05cm}
			\begin{center}
            \scriptsize \textit{BTS \cite{lee2019big}} fine-tuned \\ on \textit{SeasonDepth}
            \end{center}
		\end{minipage}%
		\begin{minipage}[t]{0.16\linewidth}
			\centering
			\includegraphics[width=\linewidth]{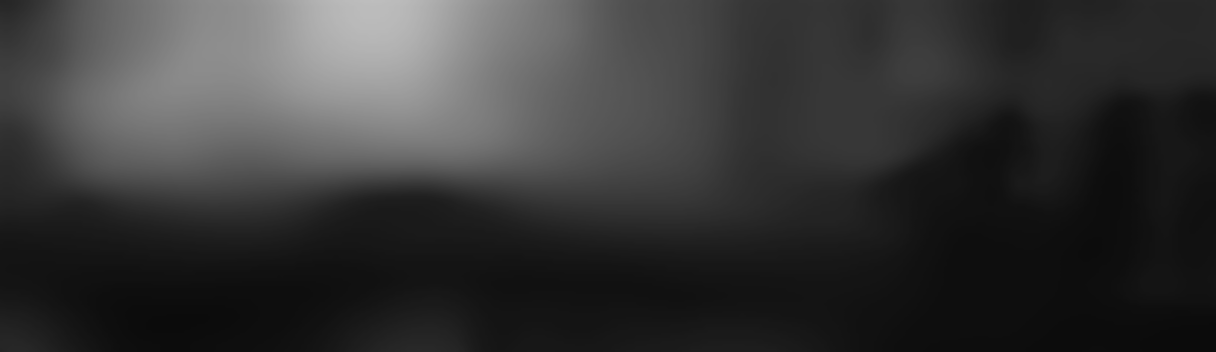}\\
			\vspace{0.05cm}
			\includegraphics[width=\linewidth]{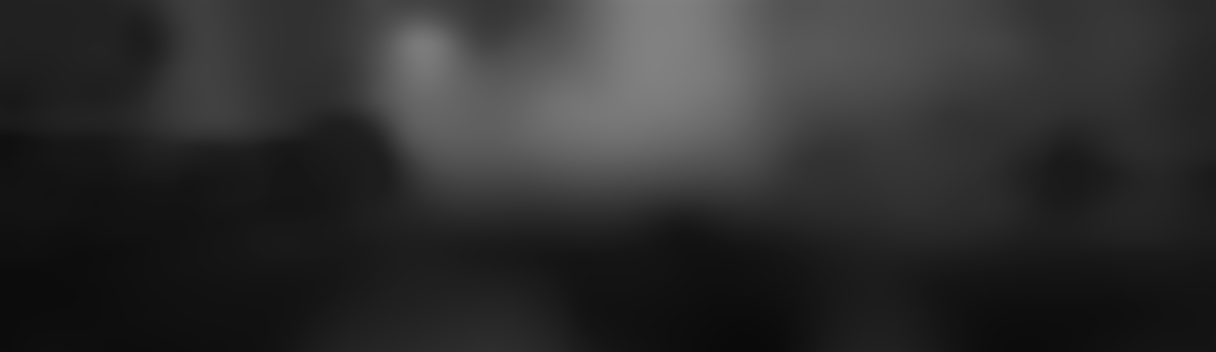}\\
			\vspace{0.05cm}
			\includegraphics[width=\linewidth]{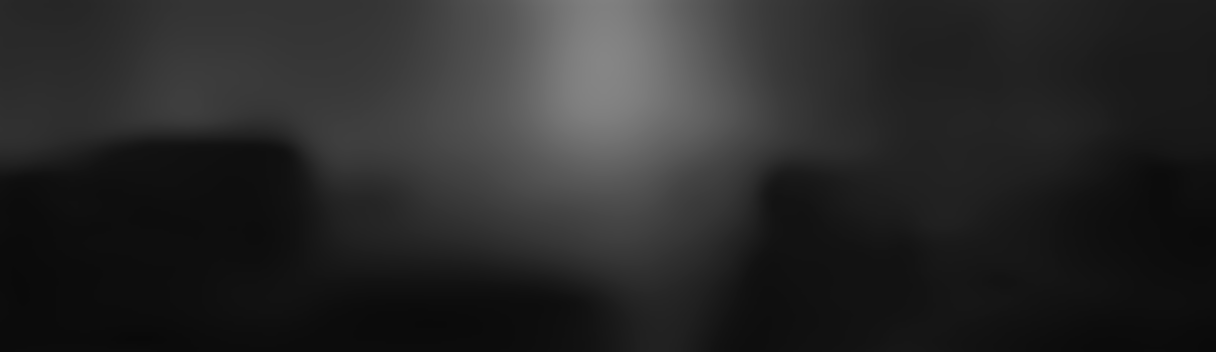}\\
			\vspace{-0.05cm}
			\begin{center}
            \scriptsize  \textit{SfMLearner \cite{zhou2017unsupervised}} fine-tuned on \textit{Cityscapes} \cite{cordts2016cityscapes}
            \end{center}
		\end{minipage}%
		\begin{minipage}[t]{0.16\linewidth}
			\centering
			\includegraphics[width=\linewidth]{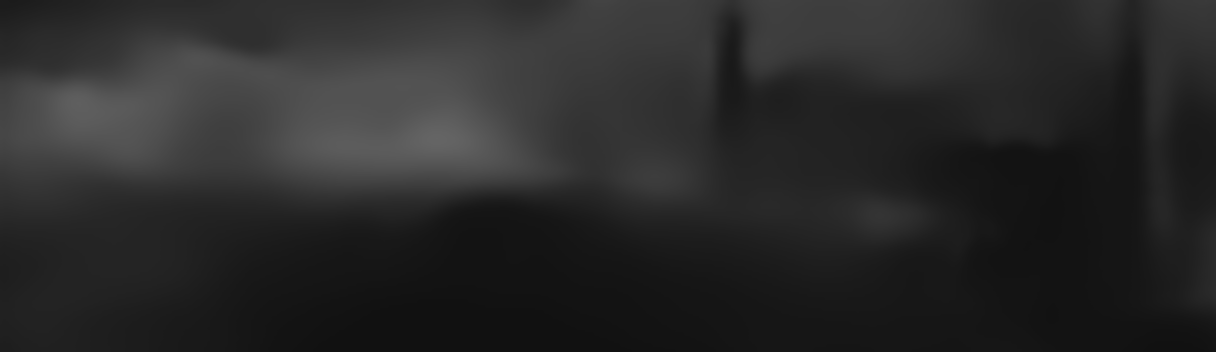}\\
			\vspace{0.05cm}
			\includegraphics[width=\linewidth]{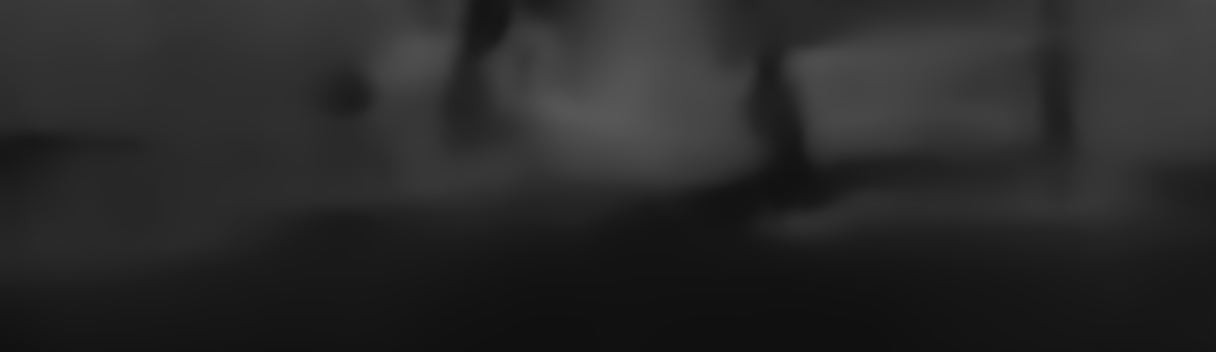}\\
			\vspace{0.05cm}
			\includegraphics[width=\linewidth]{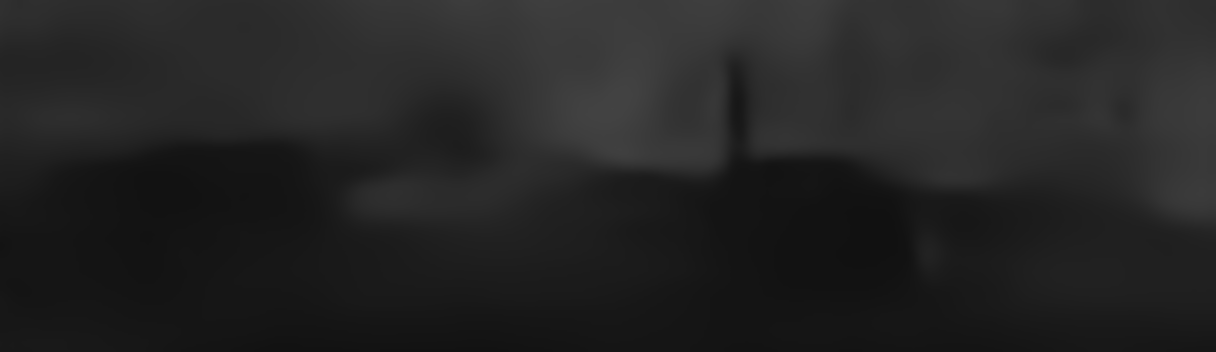}\\
			\vspace{-0.05cm}
			\begin{center}
            \scriptsize  \textit{SfMLearner \cite{zhou2017unsupervised}} fine-tuned on \textit{SeasonDepth}
            \end{center}
		\end{minipage}%
	\caption{Qualitative comparison on \textit{KITTI}  \cite{Uhrig2017THREEDV} with depth prediction models fine-tuned on \textit{SeasonDepth} and \textit{Cityscapes} \cite{cordts2016cityscapes}
	}
	\label{justification_qual}
\end{figure*}

\subsection{Influence of Challenging Environments}

In this section, we further investigate which environment is more difficult to the current depth prediction models.
 The abbreviations of environments in Fig. \ref{diagrams_absrelanda1}  are \textit{S} for \textit{Sunny}, \textit{C} for \textit{Cloudy}, \textit{O} for \textit{Overcast}, \textit{LS} for \textit{Low Sun}, \textit{Sn} for \textit{Snow},  \textit{F} for \textit{Foliage}, \textit{NF} for \textit{No Foliage}, and \textit{MF} for \textit{Mixed Foliage}.
 From Fig. \ref{diagrams_absrelanda1}, we can see that 
dusk scenes in \textit{LS+MF, Nov. $ 3^{rd} $} and snowy scenes in \textit{LS+NF+Sn, Dec. $ 21^{st} $} pose great challenge for most algorithms\textcolor{revision}{, which points out directions for future research and safe applications}. Besides, the consistent error bar in Fig. \ref{diagrams_absrelanda1} shows such adverse environments always result in large deviations for all algorithms.

Under these adverse environmental conditions, promising algorithms can also be found. For the dusk or snowy scenes,
some domain adaptation methods \cite{atapour2018real,zheng2018t2net} present impressive robustness under adverse scenes due to the various appearances of synthetic images. For the snowy scenes, self-supervised models
are less influenced compared to supervised methods. 
 Qualitative experimental results in Fig. \ref{benchmark_results} show how extreme illumination or vegetation changes affect depth prediction.
 From the top two rows, it can be seen that the illumination change of low sun makes the depth prediction of tree trunks less clear under the same vegetation condition as green and red blocks show. Also,  no foliage tends to make telephone poles and tree trunks less distinguishable by comparing red and green blocks from the last two rows, while the depth prediction of heavy vegetation is difficult as red blocks show on the fourth row given the same illumination and weather condition. More  results  can be found in Appendix Sec. \ref{more_qualitative_results}.

 \subsection{Cross-dataset Comparison with Cityscapes}
 \label{model_training} 
 
 From the quantitative results in Tab. \ref{justification_quan2}, the \textit{KITTI} performance from models fine-tuned on \textit{SeasonDepth} is mostly better than models fine-tuned on \textit{Cityscapes} with similar fluctuation of model performance.
 Based on the qualitative performance in Fig. \ref{justification_qual}, we can find that models fine-tuned on \textit{SeasonDepth} perform better than  those fine-tuned on \textit{Cityscapes} on the unseen \textit{KITTI} dataset.
 Consequently, although the depth maps of \textit{SeasonDepth} are reconstructed from structure from motion and do not contain dynamic objects,  the ground truth accuracy is eligible  to be used for model training compared to the stereo depth dataset \textit{Cityscapes}, justifying our ground truth accuracy is adequate to be beneficial to cross-dataset generalization ability.
 

\subsection{ \textcolor{revision2}{Further Discussion}}
 \textcolor{revision2}{In this section, we will discuss how to improve the robustness across multiple environments to boost more research on long-term robust visual perception. The key problem of long-term robust perception is the real-world out-of-distribution robustness of machine learning models \cite{hendrycks2021many}, where images from changing environments act as samples from different distribution with respect to the training distribution. Since real-world environments are very hard to quantify using specific distribution distance, the robust perception is very challenging. Empirically, research about long-term performance under changing environments stems from visual place recognition and localization. Most of deep learning based methods leverage environmentally-insensitive perceptual auxiliary information like semantic \cite{xu2018unsupervised,benbihi2020image,9296559}, geometric \cite{piasco2019learning,piasco2020improving},or learn the domain-invariant representation \cite{hu2019retrieval,zhou2020cross,tang2020adversarial} or image translation \cite{jenicek2019no, zheng_2020_ECCV} in multi-domain setting to deal with changing environments. Viewing the monocular depth prediction as pixel-level regression, we believe \textit{SeasonDepth} will facilitate future research  theoretically and empirically.}    

\section{Conclusion}
\label{conclusion}
In this paper, a new dataset \textit{SeasonDepth} is built for monocular depth prediction under different environments, and supervised and self-supervised state-of-the-art open-source algorithms are evaluated. From the experimental results, we find that there is still a long way to go to achieve robustness for long-term  depth prediction \textcolor{revision}{ and several promising avenues are given,
    pointing out self-supervised methods are more robust to changing environments.
Through studying how adverse environments influence, our findings via this dataset and benchmark will impact the research on long-term robust perception and related applications.} 

{\small
\bibliographystyle{ieee_fullname}
\bibliography{arxiv.bib}
}

\clearpage
\appendix
\section{Building SeasonDepth Dataset}
\label{dataset_details}

In this section, we present more details about the process of building \textit{SeasonDepth} dataset and statistical analysis of depth maps in each environment.

\subsection{Details in Building Dataset}
\label{building_details}
We adopt the categorized slices of the Urban part according to  \cite{sattler2018benchmarking} as original images after rectification through the camera intrinsic file. Specifically, we use \texttt{slice2, slice3, slice7, slice8} as the split validation slices for cross-dataset evaluation and benchmark, slices \texttt{slice4, slice5, slice6} are intended to treat as training sets, and \texttt{slice9} is used as the test set for the evaluation of well-tuned methods. Note that since not all images from the original dataset are appropriate for depth prediction due to huge noise, \textit{e.g.}, a moving truck covering almost all the pixels, we remove such images in the final version.
		The numbers of images under all the environments for all slices in training, validation, and test set are shown in Tab. \ref{num_dataset}. The abbreviations of environments are \texttt{S} for \texttt{Sunny}, \texttt{C} for \texttt{Cloudy}, \texttt{O} for \texttt{Overcast}, \texttt{LS} for \texttt{Low Sun}, \texttt{Sn} for \texttt{Snow},  \texttt{F} for \texttt{Foliage}, \texttt{NF} for \texttt{No Foliage}, and \texttt{MF} for \texttt{Mixed Foliage}. It could be seen that the total number of validation set is larger than that of the training set with more different slices, which helps to make the benchmark results more accurate and reliable. Also, the training set can be used to fine-tune pre-trained models, which do not need too many images. Images from the left and right cameras are merged together in the same slice for calculation.
		\begin{table*}[htbp]
		\begin{center}
			\caption{ Numbers of Images under All the Environments for All Slices}
			\label{num_dataset}
	\begin{tabular}{ccccccccccc}
		\toprule
		\multirow{2}{*}{Environments}                                & \multicolumn{4}{c}{Training Set}                                                 & \multicolumn{5}{c}{Validation Set}    & Test Set                                                           \\ \cmidrule(r){2-5}  \cmidrule(r){6-10} \cmidrule(r){11-11} 
		& slice4 & slice5 & slice6 & \begin{tabular}[c]{@{}c@{}}All \\ Slices\end{tabular} & slice2 & slice3 & slice7 & slice8 & \begin{tabular}[c]{@{}c@{}}All \\ Slices\end{tabular} & slice9 \\ \midrule
		\begin{tabular}[c]{@{}c@{}}S+NF\\ Apr. 4th\end{tabular}      & 221    & 129    & 543    & 893                                                  & 382    & 450    & 190    & 449    & 1471      & 380                                            \\\cmidrule(r){1-1}
		\begin{tabular}[c]{@{}c@{}}S+F\\ Sept. 1st\end{tabular}      & 116    & 230    & 190    & 536                                                  & 385    & 464    & 249    & 490    & 1588    & 334                                               \\\cmidrule(r){1-1}
		\begin{tabular}[c]{@{}c@{}}S+F\\ Sept. 15th\end{tabular}     & 202    & 213    & 526    & 941                                                  & 335    & 329    & 462    & 457    & 1583    & 283                                              \\\cmidrule(r){1-1}
		\begin{tabular}[c]{@{}c@{}}C+F\\ Oct. 1st\end{tabular}       & 406    & 205    & 626    & 1237                                                  & 347    & 438    & 350    & 244    & 1379    & 338                                              \\\cmidrule(r){1-1}
		\begin{tabular}[c]{@{}c@{}}S+F\\ Oct. 19th\end{tabular}      & 288    & 192    & 558    & 1038                                                  & 301    & 439    & 412    & 230    & 1382      & 166                                            \\\cmidrule(r){1-1}
		\begin{tabular}[c]{@{}c@{}}O+MF\\ Oct. 28th\end{tabular}     & 394    & 194    & 536    & 1124                                                  & 333    & 418    & 362    & 442    & 1555     & 338                                              \\\cmidrule(r){1-1}
		\begin{tabular}[c]{@{}c@{}}LS+MF\\ Nov. 3rd\end{tabular}     & 445    & 198    & 399    & 1042                                                  & 335    & 447    & 203    & 416    & 1401    & 351                                               \\\cmidrule(r){1-1}
		\begin{tabular}[c]{@{}c@{}}LS+MF\\ Nov. 12th\end{tabular}    & 0    & 221    & 552    & 762                                                  & 352    & 500    & 357    & 501    & 1710    & 366                                               \\\cmidrule(r){1-1}
		\begin{tabular}[c]{@{}c@{}}C+MF\\ Nov. 22nd\end{tabular}     & 323    & 163    & 578    & 1064                                                  & 298    & 436    & 380    & 423    & 1537   & 321                                                \\\cmidrule(r){1-1}
		\begin{tabular}[c]{@{}c@{}}LS+NF+Sn\\ Dec. 21st\end{tabular} & 241    & 14    & 592    & 847                                                  & 284    & 512    & 56     & 147    & 999   & 346                                                 \\\cmidrule(r){1-1}
		\begin{tabular}[c]{@{}c@{}}LS+F\\ Mar. 4th\end{tabular}      & 175    & 19    & 498    & 692                                                  & 354    & 222    & 0      & 512    & 1088      & 382                                             \\\cmidrule(r){1-1}
		\begin{tabular}[c]{@{}c@{}}O+F\\ Jul. 28th\end{tabular}       & 458    & 212    & 560    & 1230                                                  & 256    & 425    & 384    & 467    & 1532    & 309                                               \\\cmidrule(r){1-1}
		\begin{tabular}[c]{@{}c@{}}All \\ Environments\end{tabular}  & 3269   & 1980   & 6158   & 11407                                                 & 3962   & 5080   & 3405   & 4778   & 17225   & 3944                                               \\ \bottomrule
	\end{tabular}
\end{center}
\end{table*}
	
\begin{figure*}[htbp]
	\centering
		\begin{minipage}[t]{0.19\linewidth}
			\centering
			\includegraphics[width=\linewidth]{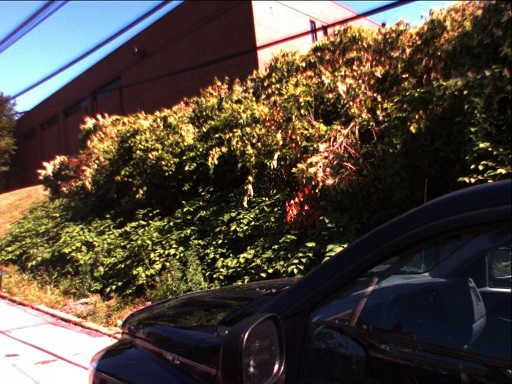}\\
			\vspace{0.05cm}
			\includegraphics[width=\linewidth]{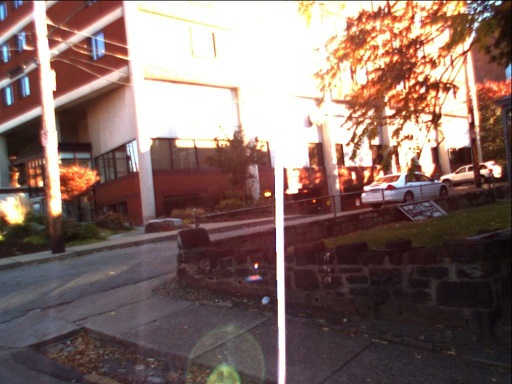}\\
			\vspace{0.05cm}
			\includegraphics[width=\linewidth]{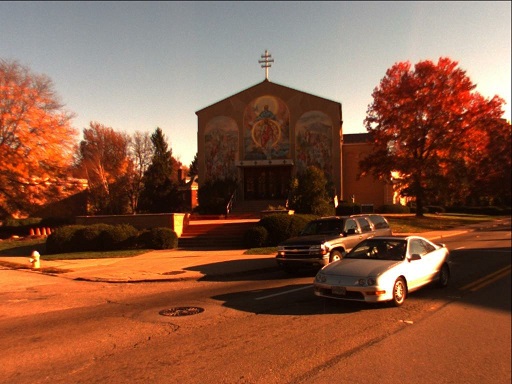}\\
			\vspace{0.05cm}
			\includegraphics[width=\linewidth]{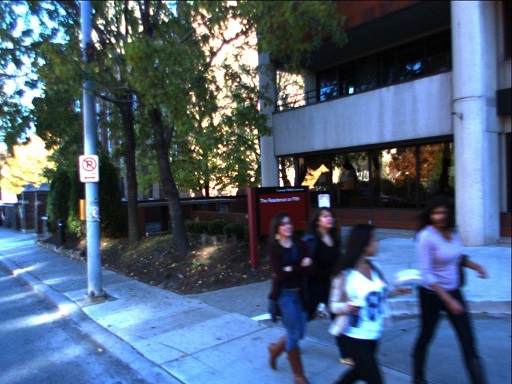}\\
			\vspace{0.05cm}
			\includegraphics[width=\linewidth]{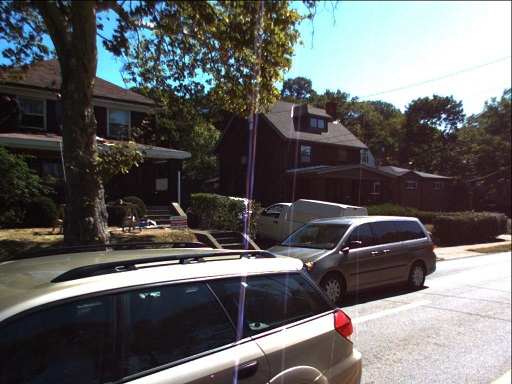}\\
			\vspace{0.05cm}
			\includegraphics[width=\linewidth]{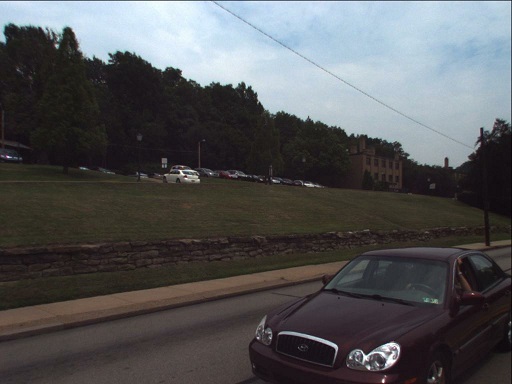}\\
			\vspace{0.05cm}
			\includegraphics[width=\linewidth]{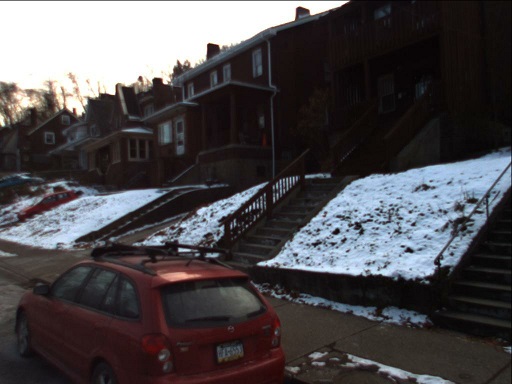}\\
			\vspace{0.05cm}
			\includegraphics[width=\linewidth]{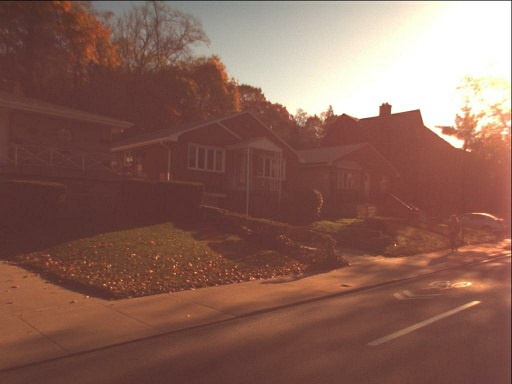}\\
			\vspace{0.05cm}
		\end{minipage}%
		\begin{minipage}[t]{0.19\linewidth}
			\centering
			\includegraphics[width=\linewidth]{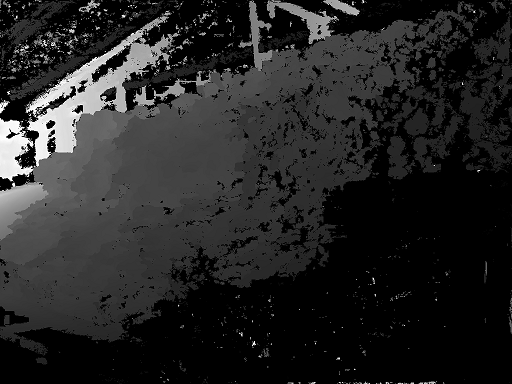}\\
			\vspace{0.05cm}
			\includegraphics[width=\linewidth]{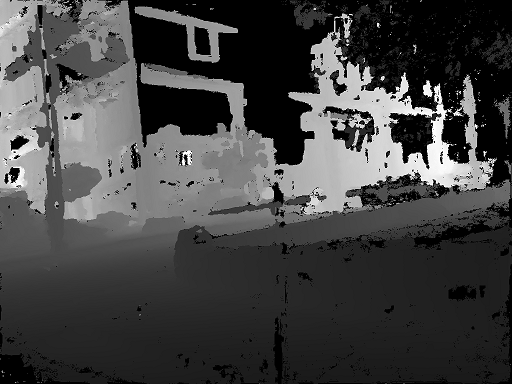}\\
			\vspace{0.05cm}
			\includegraphics[width=\linewidth]{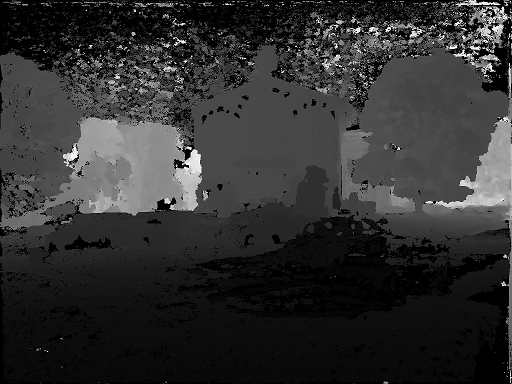}\\
			\vspace{0.05cm}
			\includegraphics[width=\linewidth]{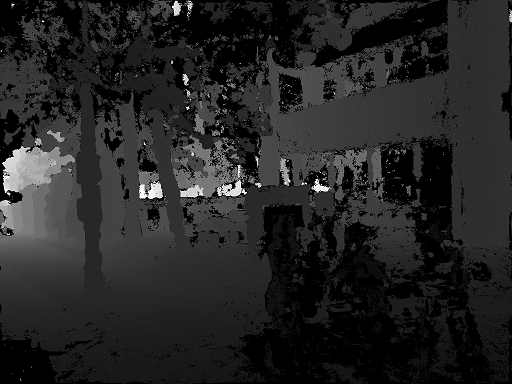}\\
			\vspace{0.05cm}
			\includegraphics[width=\linewidth]{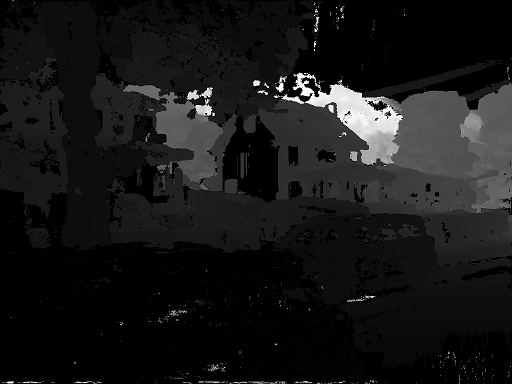}\\
			\vspace{0.05cm}
			\includegraphics[width=\linewidth]{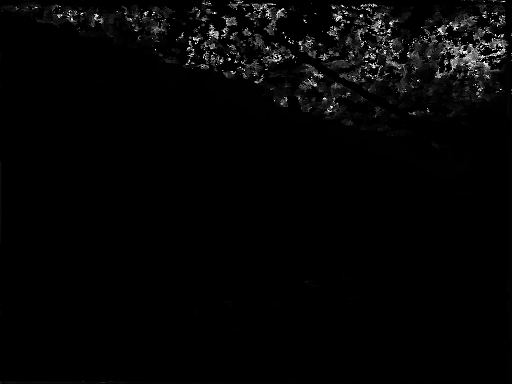}\\
			\vspace{0.05cm}
			\includegraphics[width=\linewidth]{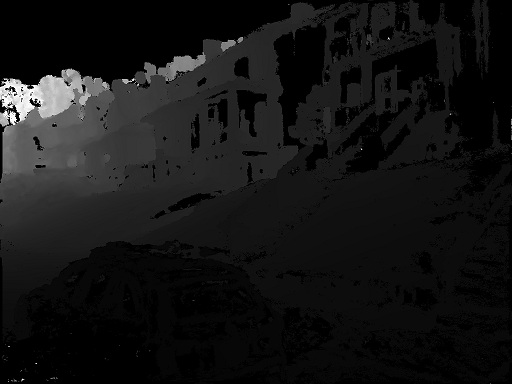}\\
			\vspace{0.05cm}
			\includegraphics[width=\linewidth]{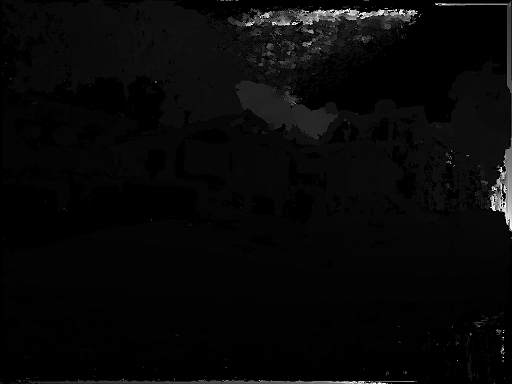}\\
			\vspace{0.05cm}
		\end{minipage}%
		\begin{minipage}[t]{0.19\linewidth}
			\centering
			\includegraphics[width=\linewidth]{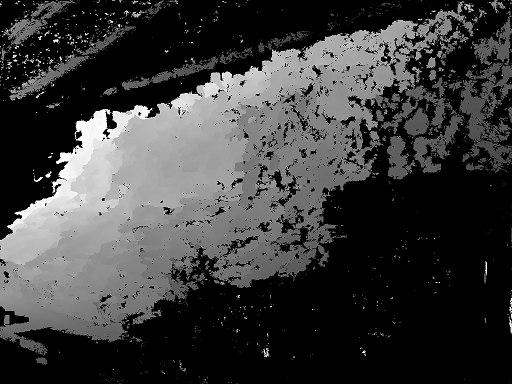}\\
			\vspace{0.05cm}
			\includegraphics[width=\linewidth]{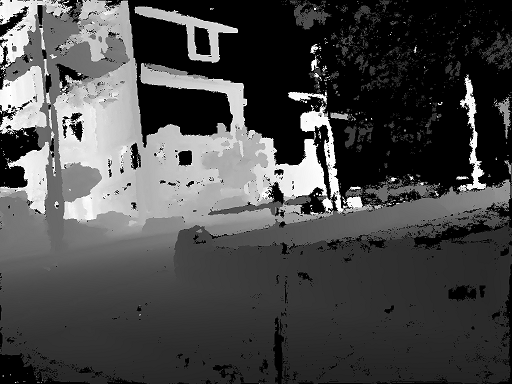}\\
			\vspace{0.05cm}
			\includegraphics[width=\linewidth]{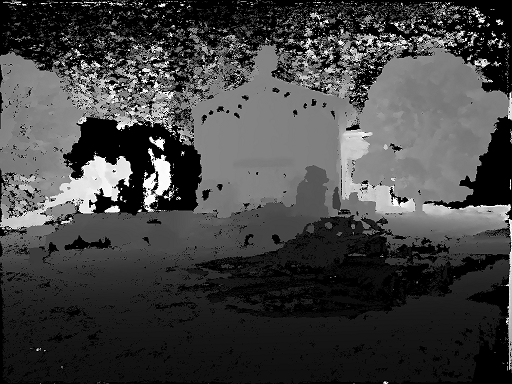}\\
			\vspace{0.05cm}
			\includegraphics[width=\linewidth]{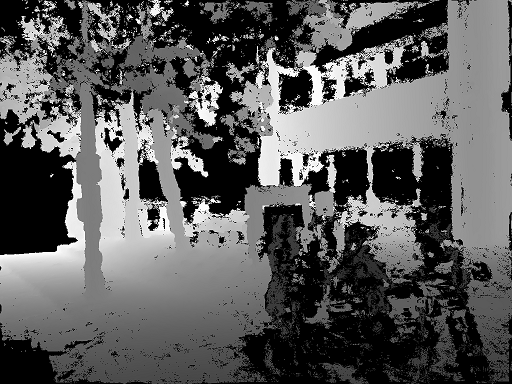}\\
			\vspace{0.05cm}
			\includegraphics[width=\linewidth]{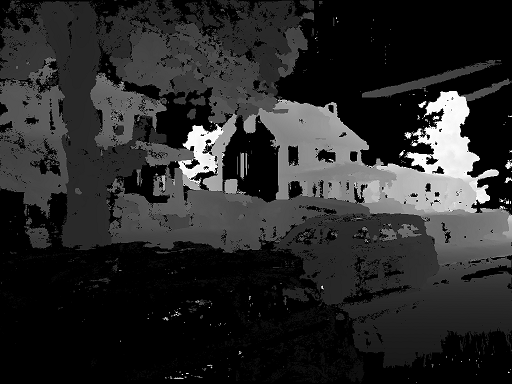}\\
			\vspace{0.05cm}
			\includegraphics[width=\linewidth]{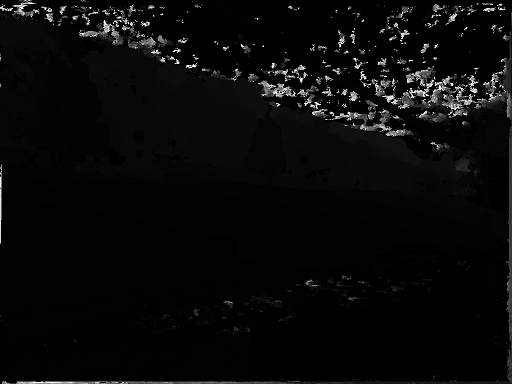}\\
			\vspace{0.05cm}
			\includegraphics[width=\linewidth]{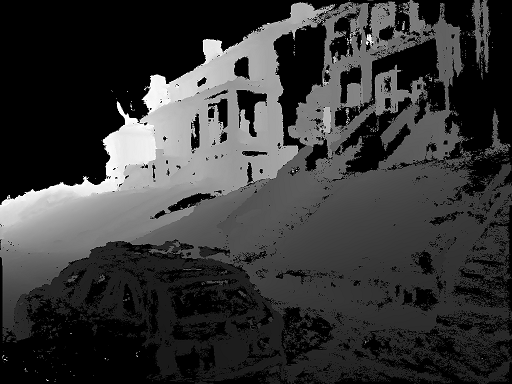}\\
			\vspace{0.05cm}
			\includegraphics[width=\linewidth]{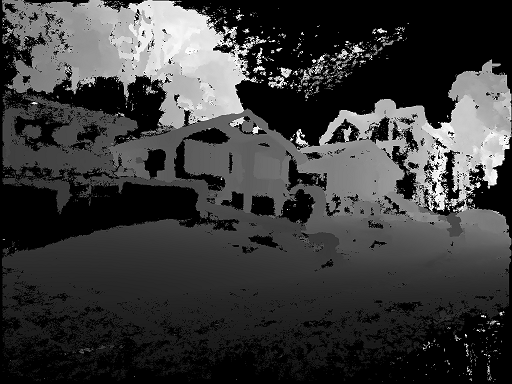}\\
			\vspace{0.05cm}
		\end{minipage}%
		\begin{minipage}[t]{0.19\linewidth}
			\centering
			\includegraphics[width=\linewidth]{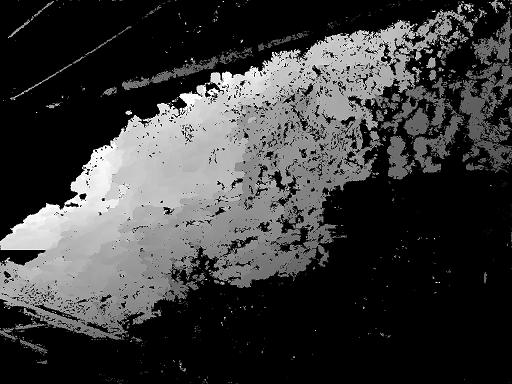}\\
			\vspace{0.05cm}
			\includegraphics[width=\linewidth]{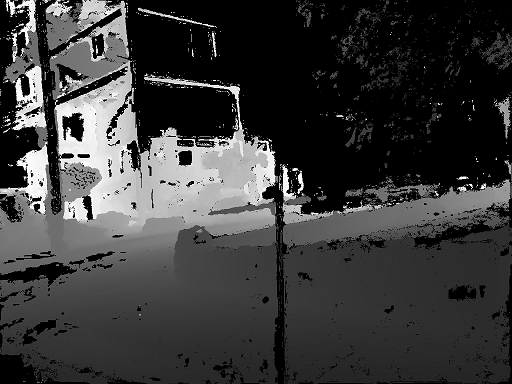}\\
			\vspace{0.05cm}
			\includegraphics[width=\linewidth]{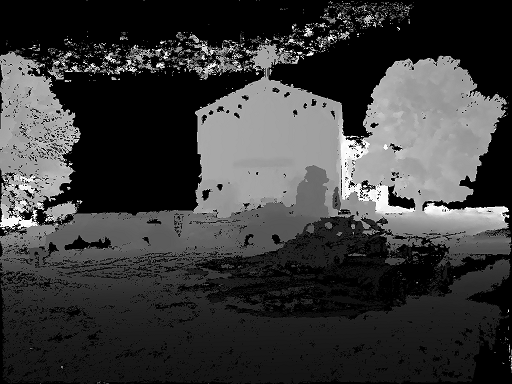}\\
			\vspace{0.05cm}
			\includegraphics[width=\linewidth]{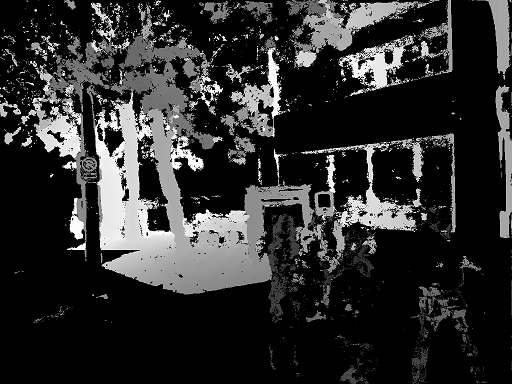}\\
			\vspace{0.05cm}
			\includegraphics[width=\linewidth]{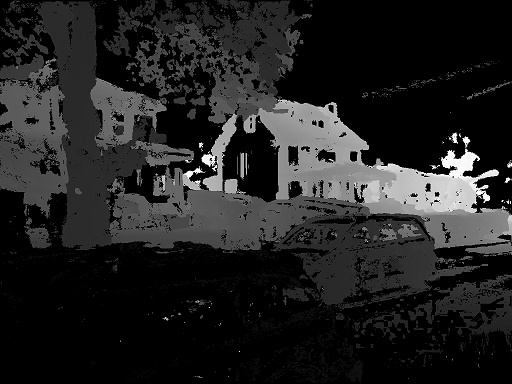}\\
			\vspace{0.05cm}
			\includegraphics[width=\linewidth]{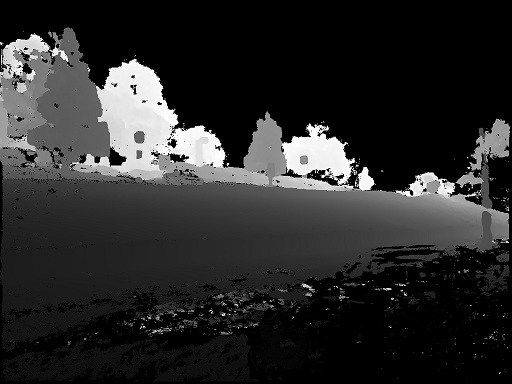}\\
			\vspace{0.05cm}
			\includegraphics[width=\linewidth]{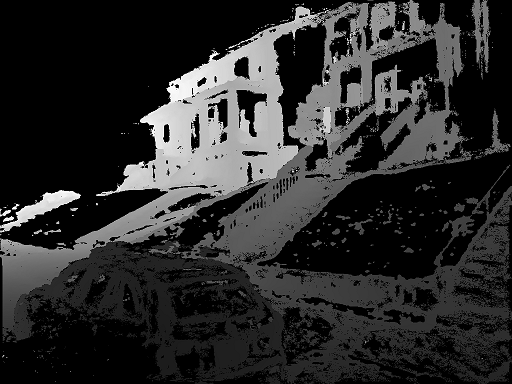}\\
			\vspace{0.05cm}
			\includegraphics[width=\linewidth]{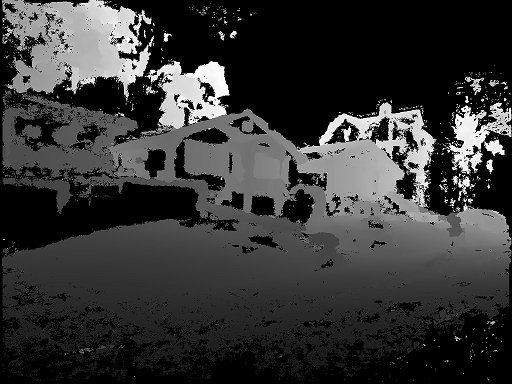}\\
			\vspace{0.05cm}
		\end{minipage}%
		\begin{minipage}[t]{0.19\linewidth}
			\centering
			\includegraphics[width=\linewidth]{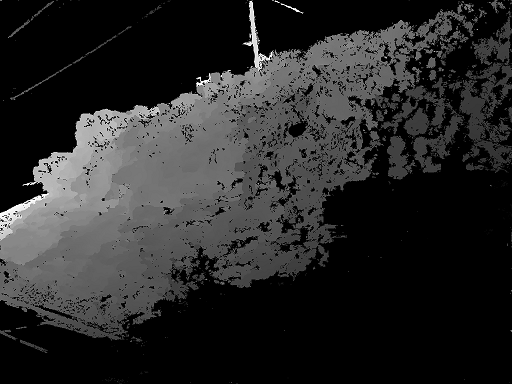}\\
			\vspace{0.05cm}
			\includegraphics[width=\linewidth]{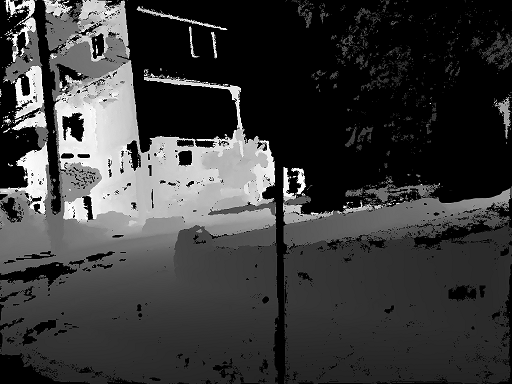}\\
			\vspace{0.05cm}
			\includegraphics[width=\linewidth]{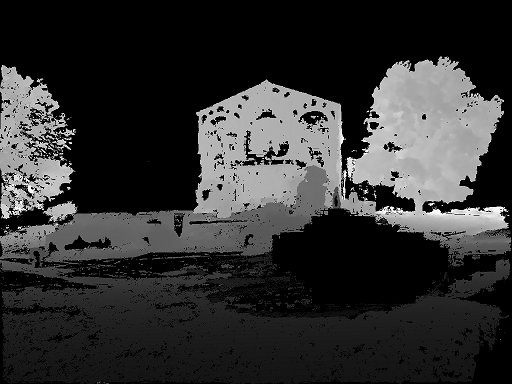}\\
			\vspace{0.05cm}
			\includegraphics[width=\linewidth]{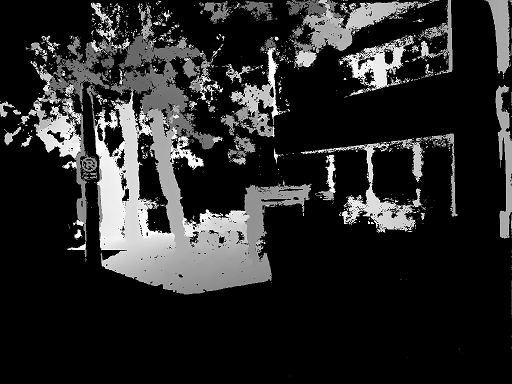}\\
			\vspace{0.05cm}
			\includegraphics[width=\linewidth]{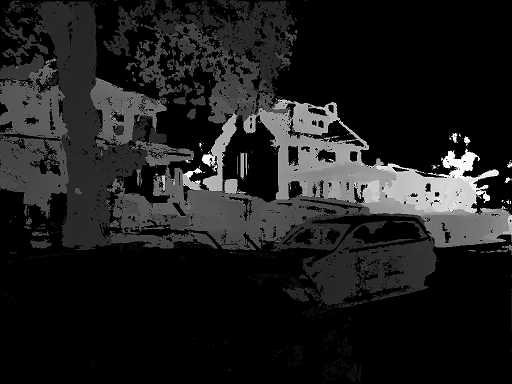}\\
			\vspace{0.05cm}
			\includegraphics[width=\linewidth]{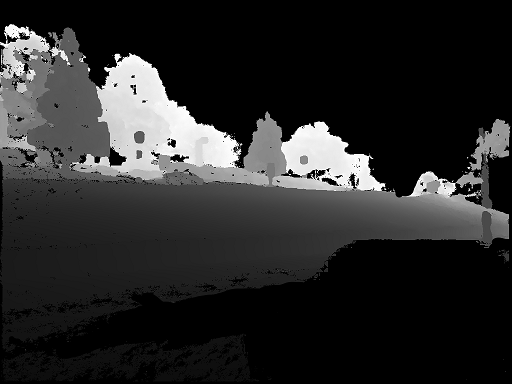}\\
			\vspace{0.05cm}
			\includegraphics[width=\linewidth]{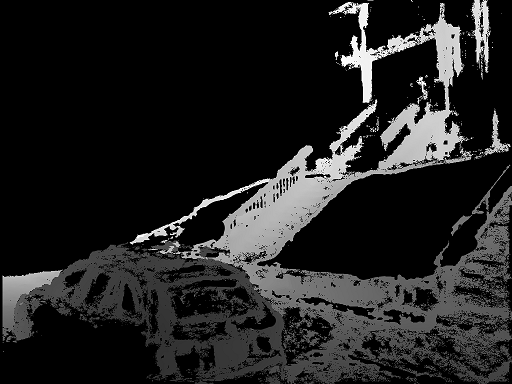}\\
			\vspace{0.05cm}
			\includegraphics[width=\linewidth]{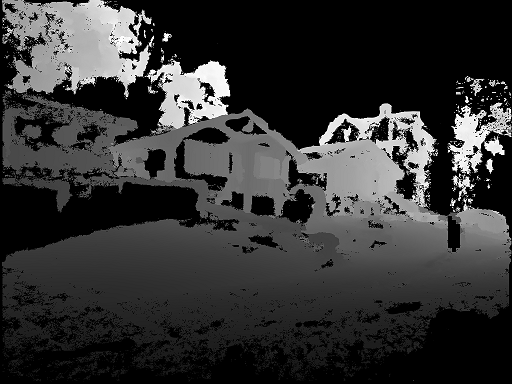}\\
			\vspace{0.05cm}
		\end{minipage}%
	\caption{The processing samples given RGB image followed by normalized depth maps for clear visualization  of (a) dense reconstruction, (b) range filtering, (c) HSV-based refinement and (d) manual post-processing.
	}
	\label{ransac_filter_sup}
\end{figure*}

	We adopt COLMAP's MVS pipeline \cite{schonberger2016structure,schonberger2016pixelwise} to find the 3D structure and depth map. We follow the instruction on \url{https://colmap.github.io/} with sequential SIFT matching with RANSAC, sparse reconstruction, and dense reconstruction.
	Some important detailed hyperparameters can be found in Tab. \ref{colmap_para}, while others are with the default configuration. To make full use of the image sequences, we adjust the sequential matching overlap to be 15 instead of the whole sequence, improving the local optimization with less noise. 
	 During each iteration of RANSAC algorithm in triangulation, the minimum inlier ratio for SIFT matching is set to be 0.65 for the consideration that most pixels of a single image are static in most cases. The maximum SIFT matching distance is 0.55 to adapt the distance of dynamic objects and improve efficiency.
	The image samples after SfM can be found in Fig. \ref{ransac_filter_sup}-(b)
	
	The valid pixels of the original depth map are between the lower threshold and upper threshold to filter most noise pixels.
	For one thing, since the fields, forests, and cloud in the far distance away from the camera matter little to the depth prediction applications for autonomous driving, we truncate the depth values over {92\%} ({80\%} in some cases) of the whole image to focus more on the near roads, vehicles, buildings, vegetation, \textit{etc.} For another, due to the camera placement on both sides of the car, the very near descriptors of the road cannot be correctly matched during SfM and reconstructed for a dense depth map, which should be removed by filtering the pixel values less than 5\% of the whole depth map. Besides, in the special cases where all the near-road noises appear on the bottom of the images, we directly filter the pixels with depth values greater than a threshold in that rectangular bottom area of the images. The samples after depth range truncation can be seen in Fig. \ref{ransac_filter_sup}-(c).

	Although depth range truncation removes some pixels with too large depth values, there are still misrecontructed pixels of sky, cloud or shadow with normal depth values. We use \texttt{PowerToys} from \url{https://github.com/microsoft/PowerToys} to pick up typical HSV values for further refinement and denoising. As Tab. \ref{hsv_para} shows, the minimal and maximal HSV values are given for some typical noises, including sky, cloud, reflections and shadows. For the clear or cloudy sky, Value tends to be high around 200 and Hue is usually blue or white. However, for those areas in the shadow of low sun, Saturation and Value are extremely low to be about 10\% so the depth map pixels are too hard to be correctly reconstructed, which need to be filtered. 
	The samples after HSV refinement are shown in Fig. \ref{ransac_filter_sup}-(d).
	
	Though RANSAC algorithm inside the SfM and MVS pipeline largely removes pixels of the dynamic objects to ensure the accuracy of overall depth values, the dynamic pixels cannot be fully eliminated and the contours of objects are not clear as well. Therefore, we employ MaskRCNN \cite{he2017mask} with pre-trained models from Detectron2 on \url{https://github.com/facebookresearch/detectron2}. 
	We adopt the pre-trained model with configuration file of \textit{COCO-InstanceSegmentation/mask\_rcnn\_R50\_FPN\_3x.yaml} and modify the \texttt{MODEL.ROI\_HEADS.SCORE\_THRESH\_TEST} to be \texttt{0.5} to find the instance segmentation with the class of \texttt{car}, \texttt{person} and \texttt{bus}. To process the image directly, we modify the visualization part in the official colab notebook, omitting boxes, keypoints and labels and letting $\alpha=1$ in \texttt{draw\_polygon} function to set the pixels of the target objects to be black.	
	But semantic or instance segmentation cannot distinguish dynamic objects that need to be removed, we use human annotation to check whether segmented vehicles or pedestrians are moving or not, relabeling the missing dynamic objects and correcting the mislabeled objects. The depth map samples after all the post-processing can be found in Fig. \ref{ransac_filter_sup}-(e). \textcolor{revision}{Note that since there are often more mis-reconstructed depth pixels around thin objects like branches and poles, we manually filter some of them in the processing for accuracy and reliable evaluation.}
	\begin{table*}[htbp]
		\begin{center}
			\caption{Some Important Hyperparameters for COLMAP}
			\label{colmap_para}
		\begin{tabular}{ccc}
			\toprule
			Process                                   & Hyperparameter                 & Value \\ \midrule
			\multirow{4}{*}{Sequential SIFT Matching} & \texttt{min\_inlier\_ratio}             & \texttt{0.65}  \\
			& \texttt{max\_distance}                  & \texttt{0.55}  \\
			& \texttt{min\_num\_inliers}              & \texttt{50}    \\
			& \texttt{overlap\_num}                   & \texttt{15}    \\ \midrule
			\multirow{3}{*}{RANSAC}                   & \texttt{dyn\_num\_trials\_multiplier}   & \texttt{3.0}   \\
			& \texttt{confidence}                     & \texttt{0.99}  \\
			& \texttt{min\_inlier\_ratio}             & \texttt{0.1}   \\ \midrule
			\multirow{3}{*}{Sparse Reconstucion}      & \texttt{abs\_pose\_min\_inlier\_ratio}  & \texttt{0.25}  \\
			& \texttt{filter\_max\_reproj\_error}     & \texttt{4.0}   \\
			& \texttt{filter\_min\_tri\_angle}        & \texttt{1.5}   \\ \midrule
			\multirow{2}{*}{Dense Reconstucion}       & \texttt{geom\_consistency\_max\_cost}   & \texttt{3.0}   \\
			& \texttt{geom\_consistency\_regularizer} & \texttt{0.3}   \\ \bottomrule
		\end{tabular}
	\end{center}
	\end{table*}
\begin{table*}[htbp]
		\begin{center}
			\caption{Some Typical Noises and HSV Thresholds}
			\label{hsv_para}
			\resizebox{\textwidth}{!}{
			\begin{tabular}{ccc}
				\toprule
				Noise Source and Type                                      & \begin{tabular}[c]{@{}c@{}}minimal threshold\\ \texttt{(H, S, V)}\end{tabular} & \begin{tabular}[c]{@{}c@{}}maximal threshold\\ \texttt{(H, S, V)}\end{tabular} \\ \midrule
				Blue Sky                                        & \texttt{(172, 5\%, 40\%)}       & \texttt{(240, 90\%, 100\%)}    \\
				White Cloud and Bright Reflections from Windows & \texttt{(0, 0\%, 100\%)}       & \texttt{(360, 100\%, 100\%)}   \\
				Dark and Black Shadows                          & \texttt{(0,0\%,0\%)}           & \texttt{(0,0\%,0)\%}           \\
				Dusk Cloud and Refections from Roads and Cars   & \texttt{(0,0\%,70\%)}          & \texttt{(90,20\%,100\%)}       \\
				Dusk Sky                                        & \texttt{(140, 11\%, 40\%)}     & \texttt{(160, 50\%, 100\%)}    \\ \bottomrule
			\end{tabular}
			}
		\end{center}
	\end{table*}

\subsection{Statistics and Analysis of Depth Map for Each Environment}
\label{stat_dataset}
Here we give the statistical analysis of the proposed \textit{SeasonDepth} dataset for each environment. Since all the depth values are scale-free and not absolute for distance, it is not applicable to directly find the pixel value distribution for the dataset as  \cite{guizilini20203d,vasiljevic2019diode} do.
However, the depth values of sequential frames in similar urban scenes under the same environment are similarly distributed, \textit{i.e.} the depth values of images along similar streets and blocks are consistent. Then the key point is to align the distribution of each environment to the mean of all environments, obtaining the normalized whole distribution map and dismissing the scale discrepancy.

\begin{figure*}[]
\centering
\includegraphics[width=\linewidth]{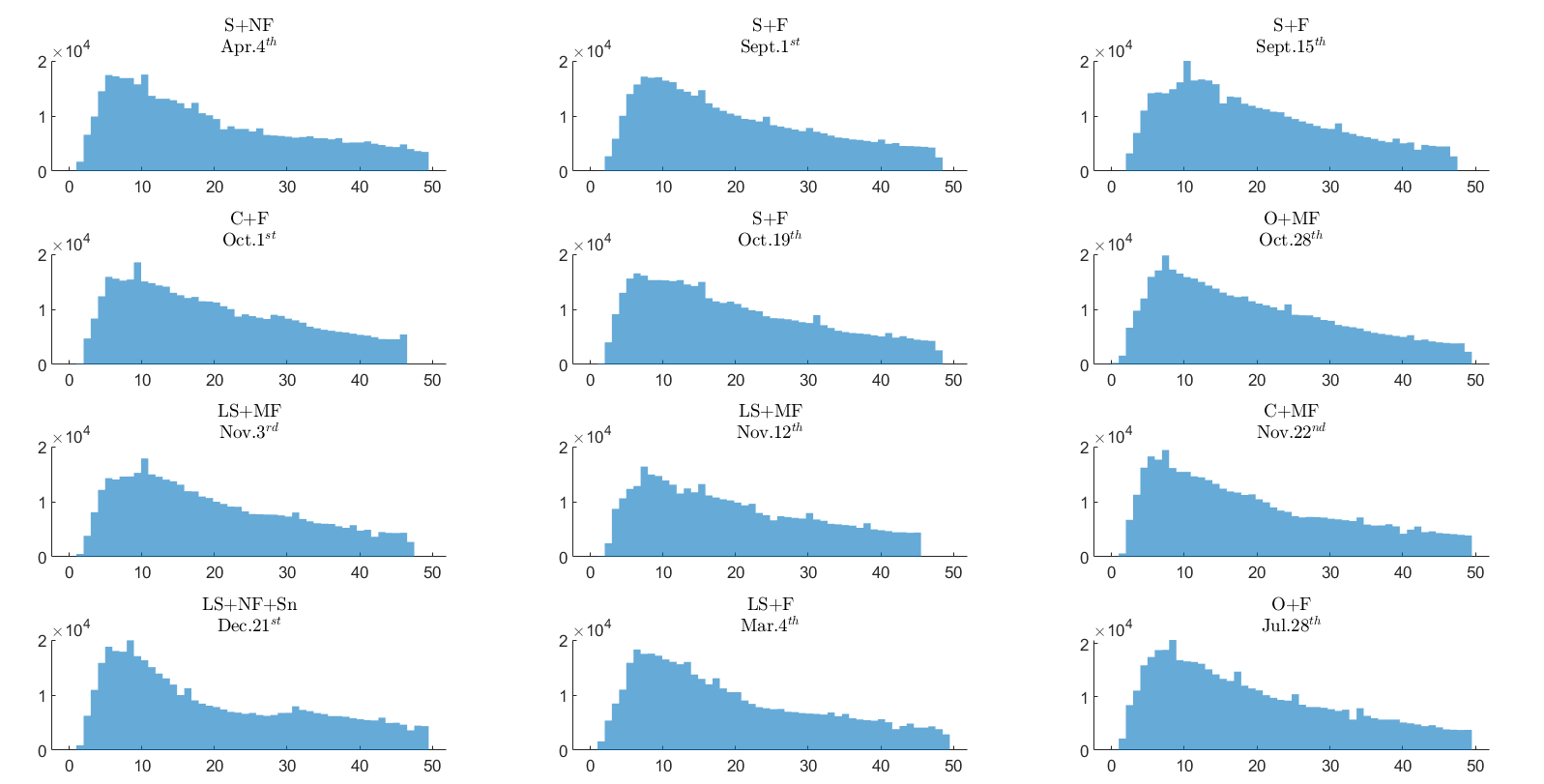}
\caption{The normalized depth map distribution under all environments. The values of y-axes are the number of pixels with the value of abscissa on each image on average.}
\label{all_envs}
\end{figure*}

Therefore, we first find the original depth value distribution $ p_{D_i}(x) $ for all the slices under each environment $ i $. Then lower quartile $ Q_1 $ (25\%), median $ Q_2 $ (50\%) and upper quartile $ Q_3 $ (75\%) are calculated for the original distribution of every environment and the mean value of quartiles can be found as reference quartiles$ Q_{{1_{ref}}},Q_{{2_{ref}}},Q_{{3_{ref}}} $ for all $n$ environments, \[{Q_{{1_{ref}}}} = \frac{1}{n}\sum\limits_{i = 1}^n {{Q_{{1_i}}}}, {Q_{{2_{ref}}}} = \frac{1}{n}\sum\limits_{i = 1}^n {{Q_{{2_i}}}}, {Q_{{3_{ref}}}} = \frac{1}{n}\sum\limits_{i = 1}^n {{Q_{{3_i}}}}\]
To find the scale normalization ratio $ r_{i} $, we use arithmetic mean to measure the ratio of reference quartiles $ Q_{{1_{ref}}},Q_{{2_{ref}}},Q_{{3_{ref}}} $ and other quartiles $ Q_{{1_{i}}},Q_{{2_{i}}},Q_{{3_{i}}} $,
\begin{align}
\label{ratio_normalization}
r_{i} = \frac{1}{3}(\frac{{{Q_{{1_{ref}}}}}}{{{Q_{{1_i}}}}} + \frac{{{Q_{{2_{ref}}}}}}{{{Q_{{2_i}}}}} + \frac{{{Q_{{3_{ref}}}}}}{{{Q_{{3_i}}}}})
\end{align}
Then the distribution $ p_{D_i}(x) $ can be normalized to mean reference environment to obtain $ p_{D\_norm_i}(x) $,
\begin{align}
\label{distribution_normalization}
 p_{D\_norm_i}(x) = r_i p_{D_i}(x)
\end{align}
	After that, the normalized distribution of all the environments can be added directly to get the whole distribution. The distribution map of each environment can be found in Fig. \ref{all_envs}. It can be seen that all the pixels follow a similar long-tail distribution, and the average y-axis numbers of per-image pixels overcome the bias caused by unbalanced image quantities across different environments. The normalization makes each distribution aligned on the x-axis, which can be directly added to obtain the total distribution map, as Fig. \ref{distribution_map}  shows.

\section{SeasonDepth Benchmark}

\subsection{Details about Evaluated Models}
\label{benchmark_details}
For fairness in evaluating the performance algorithms under changing environments, we present the \textit{SeasonDepth} benchmark with the well-tuned models on our training data and with no limit to the pretrained state-of-the-art models for the best results. Since there are only monocular videos with depth maps in our dataset, we report the results of some supervised learning methods and monocular video based self-supervised learning methods and leave other categories in the cross-dataset generalization benchmark. Specifically, the following supervised learning models are evaluated, DepthFormer \cite{li2022depthformer} with implementation of \url{https://github.com/zhyever/Monocular-Depth-Estimation-Toolbox/tree/main/configs/depthformer}, BTS \cite{lee2019big} with the implementation of \url{https://github.com/cleinc/bts} and DPT \cite{ranftl2021vision} with pretrained models on \url{https://github.com/intel-isl/DPT/releases/download/1_0/dpt_hybrid-midas-501f0c75.pt} from \url{https://github.com/isl-org/DPT} after fine-tuning over 60 epochs with our training set.

For the well-tuned self-supervised models on \textit{SeasonDepth} benchmakr, we evaluate SUB-Depth	 \cite{zhou2021sub} with ResNet18 as the backbone for 5 epochs using learning rate 0.0001, VADepth \cite{xiang2022visual} from \url{https://github.com/xjixzz/vadepth-net},  Monodepth2 \cite{godard2019digging} from \url{https://github.com/nianticlabs/monodepth2}, SfMLearner \cite{zhou2017unsupervised} from \url{https://github.com/ClementPinard/SfmLearner-Pytorch} and ManyDepth   \cite{watson2021temporal} from \url{https://github.com/nianticlabs/manydepth} as baselines.

For the cross-dataset evaluation for the generalization of depth prediction, we further benchmark the representative  supervised, self-supervised, and domain adaptation models from the well-known \textit{KITTI} leaderboard \cite{Uhrig2017THREEDV}, which are
 with open-source codes and pre-trained models for a fair comparison. Here are some important details for all the evaluated baselines. Our experiments are conducted on two NVIDIA 2080Ti cards with 64G RAM on Ubuntu 18.04 system. The evaluation metrics are modified based on \texttt{development kit} \cite{Uhrig2017THREEDV} on \url{http://www.cvlibs.net/datasets/kitti/eval_depth.php?benchmark=depth_prediction}. 

For the supervised methods, we evaluate four representative methods, Eigen \textit{et al.} \cite{eigen2014depth}, \textit{BTS} \cite{lee2019big}, \textit{MegaDepth} \cite{li2018megadepth} \textit{VNL} \cite{yin2019enforcing}.
Eigen \textit{et al.} propose the first CNNs-based depth prediction method and introduce the famous Eigen split of \textit{KITTI} dataset for depth prediction benchmark. We hence evaluate this representative method using the PyTorch implementation through \url{https://github.com/DhruvJawalkar/Depth-Map-Prediction-from-a-Single-Image-using-a-Multi-Scale-Deep-Network} 
with the improved image gradient component in the newer loss to see the performance across multiple environments. 
Supervised work \textit{BTS} ranks $ 4th $ on the \textit{KITTI} benchmark and we test it on \url{https://github.com/cogaplex-bts/bts} using the pre-trained model \texttt{DenseNet161} on Eigen split. \textcolor{revision}{We further fine-tune this pre-trained model of \textit{BTS} on our training set for 20 epochs with a batch size of 16. The best performance of $Average$ metric is obtained from epoch 20. Due to the scaleless and partially validated ground truth, we only calculate the non-zero pixels and conduct alignment using the mean value for loss when fine-tuning.}
Note that \texttt{focal value} does not influence the experimental results due to the relative scale of the depth metrics.
We test \textit{the MegaDepth} method according to \url{https://www.cs.cornell.edu/projects/megadepth/} with the \texttt{MegaDepth} pre-trained models as described in the paper and all the hyperparameters are set as default. \textit{VNL} is evaluated using \url{https://github.com/YvanYin/VNL_Monocular_Depth_Prediction} with the pre-trained model of \texttt{ResNext101\_32x4d} backbone and trained on \textit{KITTI} dataset. 

\begin{table*}[]
	\caption{\bm{$ AbsRel $} Results (\textbf{Lower Better}) under Each Environment: Mean(\textcolor{blue}{Standard Deviation}) }
	\label{absrel_env_results}
	\centering
	\resizebox{\textwidth}{!}{
	\begin{tabular}{ccccccccccccc}
		\toprule
		\textbf{Method}         & \begin{tabular}[c]{@{}c@{}}\textbf{S+NF}\\ \textbf{Apr. 4th}\end{tabular} & \begin{tabular}[c]{@{}c@{}}\textbf{S+F}\\ \textbf{Sept. 1st}\end{tabular} & \begin{tabular}[c]{@{}c@{}}\textbf{S+F}\\ \textbf{Sept. 15th}\end{tabular} & \begin{tabular}[c]{@{}c@{}}\textbf{C+F}\\ \textbf{Oct. 1st}\end{tabular} & \begin{tabular}[c]{@{}c@{}}\textbf{S+F}\\ \textbf{Oct. 19th}\end{tabular} & \begin{tabular}[c]{@{}c@{}}\textbf{O+MF}\\ \textbf{Oct. 28th}\end{tabular} & \begin{tabular}[c]{@{}c@{}}\textbf{LS+MF}\\ \textbf{Nov. 3rd}\end{tabular} & \begin{tabular}[c]{@{}c@{}}\textbf{LS+MF}\\ \textbf{Nov. 12th}\end{tabular} & \begin{tabular}[c]{@{}c@{}}\textbf{C+MF}\\ \textbf{Nov. 22nd}\end{tabular} & \begin{tabular}[c]{@{}c@{}}\textbf{LS+NF+Sn}\\ \textbf{Dec. 21st}\end{tabular} & \begin{tabular}[c]{@{}c@{}}\textbf{LS+F}\\ \textbf{Mar. 4th}\end{tabular} & \begin{tabular}[c]{@{}c@{}}\textbf{O+F}\\ \textbf{Jul. 28th}\end{tabular} \\ \midrule
		Eigen \textit{et al.} \cite{eigen2014depth}   & 1.080(\textcolor{blue}{0.39})                                             & 1.111(\textcolor{blue}{0.40})                                             & 1.034(\textcolor{blue}{0.43})                                              & 1.061(\textcolor{blue}{0.40})                                            & 1.043(\textcolor{blue}{0.40})                                             & 1.072(\textcolor{blue}{0.38})                                              & 1.233(\textcolor{blue}{0.43})                                              & 1.125(\textcolor{blue}{0.37})                                               & 1.008(\textcolor{blue}{0.32})                                              & 1.067(\textcolor{blue}{0.42})                                                  & 1.136(\textcolor{blue}{0.54})                                             & 1.150(\textcolor{blue}{0.55})                                             \\
		BTS \cite{lee2019big}            & 0.697(\textcolor{blue}{0.29})                                             & 0.652(\textcolor{blue}{0.24})                                             & 0.605(\textcolor{blue}{0.24})                                              & 0.641(\textcolor{blue}{0.29})                                            & 0.647(\textcolor{blue}{0.27})                                             & 0.646(\textcolor{blue}{0.28})                                              & 0.758(\textcolor{blue}{0.35})                                              & 0.574(\textcolor{blue}{0.27})                                               & 0.637(\textcolor{blue}{0.27})                                              & 0.848(\textcolor{blue}{0.36})                                                  & 0.761(\textcolor{blue}{0.38})                                             & 0.657(\textcolor{blue}{0.28})                                             \\
		MegaDepth \cite{li2018megadepth}      & 0.514(\textcolor{blue}{0.20})                                             & 0.494(\textcolor{blue}{0.16})                                             & 0.471(\textcolor{blue}{0.17})                                              & 0.494(\textcolor{blue}{0.18})                                            & 0.486(\textcolor{blue}{0.18})                                             & 0.510(\textcolor{blue}{0.18})                                              & 0.574(\textcolor{blue}{0.21})                                              & 0.512(\textcolor{blue}{0.18})                                               & 0.489(\textcolor{blue}{0.19})                                              & 0.553(\textcolor{blue}{0.26})                                                  & 0.547(\textcolor{blue}{0.25})                                             & 0.530(\textcolor{blue}{0.24})                                             \\
		VNL \cite{yin2019enforcing}            & 0.321(\textcolor{blue}{0.16})                                             & 0.294(\textcolor{blue}{0.13})                                             & 0.257(\textcolor{blue}{0.11})                                              & 0.281(\textcolor{blue}{0.14})                                            & 0.281(\textcolor{blue}{0.13})                                             & 0.302(\textcolor{blue}{0.16})                                              & 0.357(\textcolor{blue}{0.20})                                              & 0.271(\textcolor{blue}{0.14})                                               & 0.282(\textcolor{blue}{0.14})                                              & 0.380(\textcolor{blue}{0.21})                                                  & 0.342(\textcolor{blue}{0.21})                                             & 0.306(\textcolor{blue}{0.15})                                             \\ \hline
		Monodepth \cite{godard2017unsupervised}      & 0.450(\textcolor{blue}{0.19})                                             & 0.437(\textcolor{blue}{0.16})                                             & 0.389(\textcolor{blue}{0.14})                                              & 0.424(\textcolor{blue}{0.18})                                            & 0.434(\textcolor{blue}{0.18})                                             & 0.432(\textcolor{blue}{0.16})                                              & 0.475(\textcolor{blue}{0.20})                                              & 0.418(\textcolor{blue}{0.17})                                               & 0.421(\textcolor{blue}{0.16})                                              & 0.465(\textcolor{blue}{0.21})                                                  & 0.441(\textcolor{blue}{0.20})                                             & 0.449(\textcolor{blue}{0.20})                                             \\
		adareg \cite{wong2019bilateral}         & 0.553(\textcolor{blue}{0.22})                                             & 0.515(\textcolor{blue}{0.16})                                             & 0.473(\textcolor{blue}{0.18})                                              & 0.489(\textcolor{blue}{0.20})                                            & 0.509(\textcolor{blue}{0.19})                                             & 0.493(\textcolor{blue}{0.19})                                              & 0.515(\textcolor{blue}{0.17})                                              & 0.463(\textcolor{blue}{0.18})                                               & 0.498(\textcolor{blue}{0.20})                                              & 0.523(\textcolor{blue}{0.20})                                                  & 0.543(\textcolor{blue}{0.29})                                             & 0.515(\textcolor{blue}{0.25})                                             \\
		monoResMatch \cite{tosi2019learning}   & 0.536(\textcolor{blue}{0.31})                                             & 0.466(\textcolor{blue}{0.24})                                             & 0.398(\textcolor{blue}{0.19})                                              & 0.444(\textcolor{blue}{0.27})                                            & 0.463(\textcolor{blue}{0.25})                                             & 0.479(\textcolor{blue}{0.31})                                              & 0.526(\textcolor{blue}{0.28})                                              & 0.428(\textcolor{blue}{0.25})                                               & 0.486(\textcolor{blue}{0.28})                                              & 0.600(\textcolor{blue}{0.40})                                                  & 0.544(\textcolor{blue}{0.39})                                             & 0.475(\textcolor{blue}{0.26})                                             \\ \hline
		SfMLearner \cite{zhou2017unsupervised}     & 0.745(\textcolor{blue}{0.29})                                             & 0.682(\textcolor{blue}{0.26})                                             & 0.644(\textcolor{blue}{0.27})                                              & 0.657(\textcolor{blue}{0.28})                                            & 0.684(\textcolor{blue}{0.29})                                             & 0.671(\textcolor{blue}{0.28})                                              & 0.718(\textcolor{blue}{0.35})                                              & 0.627(\textcolor{blue}{0.27})                                               & 0.698(\textcolor{blue}{0.27})                                              & 0.765(\textcolor{blue}{0.32})                                                  & 0.714(\textcolor{blue}{0.29})                                             & 0.713(\textcolor{blue}{0.31})                                             \\
		PackNet \cite{guizilini20203d}        & 0.715(\textcolor{blue}{0.27})                                             & 0.740(\textcolor{blue}{0.23})                                             & 0.680(\textcolor{blue}{0.26})                                              & 0.692(\textcolor{blue}{0.26})                                            & 0.672(\textcolor{blue}{0.24})                                             & 0.728(\textcolor{blue}{0.27})                                              & 0.806(\textcolor{blue}{0.27})                                              & 0.732(\textcolor{blue}{0.22})                                               & 0.682(\textcolor{blue}{0.25})                                              & 0.684(\textcolor{blue}{0.22})                                                  & 0.727(\textcolor{blue}{0.36})                                             & 0.803(\textcolor{blue}{0.43})                                             \\
		Monodepth2 \cite{godard2019digging}     & 0.476(\textcolor{blue}{0.18})                                             & 0.414(\textcolor{blue}{0.15})                                             & 0.383(\textcolor{blue}{0.17})                                              & 0.412(\textcolor{blue}{0.17})                                            & 0.396(\textcolor{blue}{0.17})                                             & 0.412(\textcolor{blue}{0.17})                                              & 0.441(\textcolor{blue}{0.23})                                              & 0.380(\textcolor{blue}{0.16})                                               & 0.414(\textcolor{blue}{0.16})                                              & 0.452(\textcolor{blue}{0.20})                                                  & 0.459(\textcolor{blue}{0.20})                                             & 0.402(\textcolor{blue}{0.16})                                             \\
		CC \cite{ranjan2019competitive}             & 0.613(\textcolor{blue}{0.23})                                             & 0.633(\textcolor{blue}{0.23})                                             & 0.587(\textcolor{blue}{0.25})                                              & 0.640(\textcolor{blue}{0.24})                                            & 0.627(\textcolor{blue}{0.27})                                             & 0.652(\textcolor{blue}{0.24})                                              & 0.768(\textcolor{blue}{0.25})                                              & 0.649(\textcolor{blue}{0.23})                                               & 0.593(\textcolor{blue}{0.24})                                              & 0.644(\textcolor{blue}{0.28})                                                  & 0.673(\textcolor{blue}{0.34})                                             & 0.703(\textcolor{blue}{0.39})                                             \\
		SGDepth \cite{klingner2020self}        & 0.635(\textcolor{blue}{0.24})                                             & 0.650(\textcolor{blue}{0.21})                                             & 0.605(\textcolor{blue}{0.23})                                              & 0.640(\textcolor{blue}{0.23})                                            & 0.628(\textcolor{blue}{0.23})                                             & 0.649(\textcolor{blue}{0.24})                                              & 0.726(\textcolor{blue}{0.26})                                              & 0.659(\textcolor{blue}{0.20})                                               & 0.599(\textcolor{blue}{0.19})                                              & 0.651(\textcolor{blue}{0.23})                                                  & 0.661(\textcolor{blue}{0.31})                                             & 0.671(\textcolor{blue}{0.29})                                             \\ \hline
		Atapour \textit{et al.} \cite{atapour2018real} & 0.741(\textcolor{blue}{0.27})                                             & 0.658(\textcolor{blue}{0.22})                                             & 0.619(\textcolor{blue}{0.24})                                              & 0.643(\textcolor{blue}{0.27})                                            & 0.667(\textcolor{blue}{0.27})                                             & 0.686(\textcolor{blue}{0.29})                                              & 0.658(\textcolor{blue}{0.28})                                              & 0.627(\textcolor{blue}{0.29})                                               & 0.708(\textcolor{blue}{0.27})                                              & 0.778(\textcolor{blue}{0.32})                                                  & 0.728(\textcolor{blue}{0.29})                                             & 0.724(\textcolor{blue}{0.30})                                             \\
		T2Net \cite{zheng2018t2net}          & 0.809(\textcolor{blue}{0.39})                                             & 0.830(\textcolor{blue}{0.29})                                             & 0.732(\textcolor{blue}{0.34})                                              & 0.796(\textcolor{blue}{0.35})                                            & 0.760(\textcolor{blue}{0.33})                                             & 0.831(\textcolor{blue}{0.35})                                              & 0.968(\textcolor{blue}{0.33})                                              & 0.797(\textcolor{blue}{0.29})                                               & 0.776(\textcolor{blue}{0.33})                                              & 0.869(\textcolor{blue}{0.37})                                                  & 0.912(\textcolor{blue}{0.48})                                             & 0.849(\textcolor{blue}{0.45})                                             \\
		GASDA \cite{zhao2019geometry}          & 0.443(\textcolor{blue}{0.24})                                             & 0.414(\textcolor{blue}{0.20})                                             & 0.402(\textcolor{blue}{0.21})                                              & 0.420(\textcolor{blue}{0.26})                                            & 0.426(\textcolor{blue}{0.24})                                             & 0.412(\textcolor{blue}{0.22})                                              & 0.495(\textcolor{blue}{0.26})                                              & 0.416(\textcolor{blue}{0.24})                                               & 0.429(\textcolor{blue}{0.24})                                              & 0.521(\textcolor{blue}{0.29})                                                  & 0.460(\textcolor{blue}{0.26})                                             & 0.423(\textcolor{blue}{0.26})                                             \\ \bottomrule
	\end{tabular}
}

~\\
~\\

\caption{\bm{$ a_1 $} Results (\textbf{Higher Better}) under Each Environment: Mean(\textcolor{blue}{Standard Deviation}) }
\label{a1_env_results}
\centering
\resizebox{\textwidth}{!}{
	\begin{tabular}{ccccccccccccc}
		\toprule
		\textbf{Method} & \textbf{\begin{tabular}[c]{@{}c@{}}S+NF\\ Apr. 4th\end{tabular}} & \textbf{\begin{tabular}[c]{@{}c@{}}S+F\\ Sept. 1st\end{tabular}} & \textbf{\begin{tabular}[c]{@{}c@{}}S+F\\ Sept. 15th\end{tabular}} & \textbf{\begin{tabular}[c]{@{}c@{}}C+F\\ Oct. 1st\end{tabular}} & \textbf{\begin{tabular}[c]{@{}c@{}}S+F\\ Oct. 19th\end{tabular}} & \textbf{\begin{tabular}[c]{@{}c@{}}O+MF\\ Oct. 28th\end{tabular}} & \textbf{\begin{tabular}[c]{@{}c@{}}LS+MF\\ Nov. 3rd\end{tabular}} & \textbf{\begin{tabular}[c]{@{}c@{}}LS+MF\\ Nov. 12th\end{tabular}} & \textbf{\begin{tabular}[c]{@{}c@{}}C+MF\\ Nov. 22nd\end{tabular}} & \textbf{\begin{tabular}[c]{@{}c@{}}LS+NF+Sn\\ Dec. 21st\end{tabular}} & \textbf{\begin{tabular}[c]{@{}c@{}}LS+F\\ Mar. 4th\end{tabular}} & \textbf{\begin{tabular}[c]{@{}c@{}}O+F\\ Jul. 28th\end{tabular}} \\ \midrule
Eigen \textit{et al.} \cite{eigen2014depth}    & 0.336(\textcolor{blue}{0.14})                                                      & 0.335(\textcolor{blue}{0.12})                                                      & 0.337(\textcolor{blue}{0.14})                                                       & 0.352(\textcolor{blue}{0.14})                                                     & 0.348(\textcolor{blue}{0.13})                                                      & 0.345(\textcolor{blue}{0.13})                                                       & 0.311(\textcolor{blue}{0.12})                                                       & 0.338(\textcolor{blue}{0.13})                                                        & 0.360(\textcolor{blue}{0.12})                                                       & 0.351(\textcolor{blue}{0.13})                                                           & 0.341(\textcolor{blue}{0.13})                                                      & 0.321(\textcolor{blue}{0.13})                                                      \\
BTS \cite{lee2019big}             & 0.200(\textcolor{blue}{0.11})                                                      & 0.201(\textcolor{blue}{0.10})                                                      & 0.233(\textcolor{blue}{0.10})                                                       & 0.218(\textcolor{blue}{0.11})                                                     & 0.225(\textcolor{blue}{0.12})                                                      & 0.217(\textcolor{blue}{0.12})                                                       & 0.183(\textcolor{blue}{0.12})                                                       & 0.263(\textcolor{blue}{0.15})                                                        & 0.221(\textcolor{blue}{0.11})                                                       & 0.161(\textcolor{blue}{0.10})                                                          & 0.185(\textcolor{blue}{0.10})                                                     & 0.201(\textcolor{blue}{0.11})                                                      \\
MegaDepth \cite{li2018megadepth}       & 0.417(\textcolor{blue}{0.14})                                                      & 0.430(\textcolor{blue}{0.13})                                                      & 0.439(\textcolor{blue}{0.15})                                                       & 0.422(\textcolor{blue}{0.16})                                                     & 0.427(\textcolor{blue}{0.13})                                                      & 0.420(\textcolor{blue}{0.15})                                                       & 0.377(\textcolor{blue}{0.13})                                                       & 0.408(\textcolor{blue}{0.15})                                                        & 0.436(\textcolor{blue}{0.15})                                                       & 0.399(\textcolor{blue}{0.17})                                                           & 0.402(\textcolor{blue}{0.17})                                                      & 0.421(\textcolor{blue}{0.15})                                                      \\
VNL \cite{yin2019enforcing}             & 0.513(\textcolor{blue}{0.21})                                                      & 0.532(\textcolor{blue}{0.18})                                                      & 0.579(\textcolor{blue}{0.18})                                                       & 0.554(\textcolor{blue}{0.20})                                                     & 0.550(\textcolor{blue}{0.19})                                                      & 0.535(\textcolor{blue}{0.20})                                                       & 0.463(\textcolor{blue}{0.20})                                                       & 0.579(\textcolor{blue}{0.19})                                                        & 0.557(\textcolor{blue}{0.21})                                                       & 0.442(\textcolor{blue}{0.19})                                                           & 0.499(\textcolor{blue}{0.23})                                                      & 0.528(\textcolor{blue}{0.21})                                                      \\ \hline
Monodepth \cite{godard2017unsupervised}       & 0.456(\textcolor{blue}{0.17})                                                      & 0.446(\textcolor{blue}{0.15})                                                      & 0.485(\textcolor{blue}{0.13})                                                       & 0.463(\textcolor{blue}{0.15})                                                     & 0.453(\textcolor{blue}{0.14})                                                      & 0.460(\textcolor{blue}{0.15})                                                       & 0.434(\textcolor{blue}{0.14})                                                       & 0.463(\textcolor{blue}{0.14})                                                        & 0.463(\textcolor{blue}{0.14})                                                       & 0.428(\textcolor{blue}{0.17})                                                           & 0.464(\textcolor{blue}{0.16})                                                      & 0.445(\textcolor{blue}{0.15})                                                      \\
adareg \cite{wong2019bilateral}          & 0.363(\textcolor{blue}{0.18})                                                      & 0.387(\textcolor{blue}{0.14})                                                      & 0.419(\textcolor{blue}{0.15})                                                       & 0.422(\textcolor{blue}{0.17})                                                     & 0.389(\textcolor{blue}{0.14})                                                      & 0.417(\textcolor{blue}{0.15})                                                       & 0.389(\textcolor{blue}{0.15})                                                       & 0.444(\textcolor{blue}{0.16})                                                        & 0.405(\textcolor{blue}{0.17})                                                       & 0.393(\textcolor{blue}{0.15})                                                           & 0.398(\textcolor{blue}{0.16})                                                      & 0.431(\textcolor{blue}{0.18})                                                      \\
monoResMatch \cite{tosi2019learning}    & 0.363(\textcolor{blue}{0.21})                                                      & 0.386(\textcolor{blue}{0.18})                                                      & 0.439(\textcolor{blue}{0.18})                                                       & 0.428(\textcolor{blue}{0.20})                                                     & 0.391(\textcolor{blue}{0.17})                                                      & 0.400(\textcolor{blue}{0.19})                                                       & 0.354(\textcolor{blue}{0.18})                                                       & 0.429(\textcolor{blue}{0.20})                                                        & 0.385(\textcolor{blue}{0.19})                                                       & 0.342(\textcolor{blue}{0.19})                                                           & 0.368(\textcolor{blue}{0.20})                                                      & 0.386(\textcolor{blue}{0.17})                                                      \\ \hline
SfMLearner \cite{zhou2017unsupervised}      & 0.251(\textcolor{blue}{0.10})                                                      & 0.268(\textcolor{blue}{0.09})                                                      & 0.270(\textcolor{blue}{0.09})                                                       & 0.284(\textcolor{blue}{0.11})                                                     & 0.268(\textcolor{blue}{0.11})                                                      & 0.271(\textcolor{blue}{0.10})                                                       & 0.271(\textcolor{blue}{0.11})                                                       & 0.292(\textcolor{blue}{0.12})                                                        & 0.258(\textcolor{blue}{0.09})                                                      & 0.245(\textcolor{blue}{0.09})                                                          & 0.253(\textcolor{blue}{0.09})                                                     & 0.254(\textcolor{blue}{0.09})                                                     \\
PackNet \cite{guizilini20203d}         & 0.436(\textcolor{blue}{0.13})                                                      & 0.394(\textcolor{blue}{0.13})                                                      & 0.422(\textcolor{blue}{0.15})                                                       & 0.435(\textcolor{blue}{0.15})                                                     & 0.430(\textcolor{blue}{0.14})                                                      & 0.429(\textcolor{blue}{0.14})                                                       & 0.368(\textcolor{blue}{0.13})                                                       & 0.403(\textcolor{blue}{0.12})                                                        & 0.458(\textcolor{blue}{0.13})                                                       & 0.450(\textcolor{blue}{0.13})                                                           & 0.444(\textcolor{blue}{0.14})                                                      & 0.386(\textcolor{blue}{0.17})                                                      \\
Monodepth2 \cite{godard2019digging}      & 0.366(\textcolor{blue}{0.17})                                                      & 0.423(\textcolor{blue}{0.16})                                                      & 0.465(\textcolor{blue}{0.19})                                                       & 0.438(\textcolor{blue}{0.17})                                                     & 0.454(\textcolor{blue}{0.18})                                                      & 0.442(\textcolor{blue}{0.16})                                                       & 0.418(\textcolor{blue}{0.19})                                                       & 0.473(\textcolor{blue}{0.18})                                                        & 0.426(\textcolor{blue}{0.17})                                                       & 0.403(\textcolor{blue}{0.17})                                                           & 0.391(\textcolor{blue}{0.18})                                                      & 0.452(\textcolor{blue}{0.16})                                                      \\
CC \cite{ranjan2019competitive}              & 0.493(\textcolor{blue}{0.19})                                                      & 0.478(\textcolor{blue}{0.18})                                                      & 0.501(\textcolor{blue}{0.21})                                                       & 0.480(\textcolor{blue}{0.20})                                                     & 0.494(\textcolor{blue}{0.19})                                                      & 0.479(\textcolor{blue}{0.19})                                                       & 0.400(\textcolor{blue}{0.15})                                                       & 0.480(\textcolor{blue}{0.18})                                                        & 0.525(\textcolor{blue}{0.18})                                                       & 0.488(\textcolor{blue}{0.19})                                                           & 0.483(\textcolor{blue}{0.20})                                                      & 0.445(\textcolor{blue}{0.21})                                                      \\
SGDepth \cite{klingner2020self}         & 0.497(\textcolor{blue}{0.17})                                                      & 0.459(\textcolor{blue}{0.16})                                                      & 0.487(\textcolor{blue}{0.19})                                                       & 0.475(\textcolor{blue}{0.18})                                                     & 0.487(\textcolor{blue}{0.17})                                                      & 0.487(\textcolor{blue}{0.18})                                                       & 0.437(\textcolor{blue}{0.14})                                                       & 0.475(\textcolor{blue}{0.15})                                                        & 0.525(\textcolor{blue}{0.15})                                                       & 0.483(\textcolor{blue}{0.16})                                                           & 0.495(\textcolor{blue}{0.18})                                                      & 0.449(\textcolor{blue}{0.19})                                                      \\ \hline
Atapour \textit{et al.} \cite{atapour2018real}  & 0.281(\textcolor{blue}{0.12})                                                      & 0.304(\textcolor{blue}{0.12})                                                      & 0.313(\textcolor{blue}{0.12})                                                       & 0.320(\textcolor{blue}{0.13})                                                     & 0.309(\textcolor{blue}{0.13})                                                      & 0.301(\textcolor{blue}{0.11})                                                       & 0.309(\textcolor{blue}{0.13})                                                       & 0.325(\textcolor{blue}{0.15})                                                        & 0.287(\textcolor{blue}{0.11})                                                       & 0.287(\textcolor{blue}{0.11})                                                           & 0.282(\textcolor{blue}{0.11})                                                      & 0.284(\textcolor{blue}{0.12})                                                      \\
T2Net \cite{zheng2018t2net}           & 0.421(\textcolor{blue}{0.17})                                                      & 0.367(\textcolor{blue}{0.15})                                                      & 0.416(\textcolor{blue}{0.17})                                                       & 0.403(\textcolor{blue}{0.17})                                                     & 0.416(\textcolor{blue}{0.16})                                                      & 0.390(\textcolor{blue}{0.16})                                                       & 0.340(\textcolor{blue}{0.13})                                                       & 0.404(\textcolor{blue}{0.15})                                                        & 0.429(\textcolor{blue}{0.17})                                                       & 0.349(\textcolor{blue}{0.14})                                                           & 0.363(\textcolor{blue}{0.16})                                                      & 0.393(\textcolor{blue}{0.17})                                                      \\
GASDA \cite{zhao2019geometry}           & 0.414(\textcolor{blue}{0.18})                                                      & 0.418(\textcolor{blue}{0.16})                                                      & 0.426(\textcolor{blue}{0.14})                                                       & 0.429(\textcolor{blue}{0.17})                                                     & 0.428(\textcolor{blue}{0.16})                                                      & 0.427(\textcolor{blue}{0.15})                                                       & 0.377(\textcolor{blue}{0.16})                                                       & 0.433(\textcolor{blue}{0.18})                                                        & 0.420(\textcolor{blue}{0.17})                                                       & 0.347(\textcolor{blue}{0.19})                                                           & 0.383(\textcolor{blue}{0.19})                                                      & 0.427(\textcolor{blue}{0.16})                                                      \\ \bottomrule
	\end{tabular}
}

\end{table*}

For self-supervised methods, we further categorize them and choose baselines respectively, \textit{i.e.} \textit{Monodepth} \cite{godard2017unsupervised}, \textit{adareg} \cite{wong2019bilateral} and \textit{monoResMatch} \cite{tosi2019learning} for stereo geometry based methods, \textit{SfMLearner} \cite{zhou2017unsupervised}, \textit{Monodepth2} \cite{godard2019digging} and \textit{PackNet} \cite{guizilini20203d} for monocular video SfM based methods, and \textit{CC} \cite{ranjan2019competitive} and \textit{SGDepth} \cite{klingner2020self} for multi-task learning with monocular SfM unsupervised pipeline. We also report the results  of \textit{FSRE-Depth} \cite{jung2021fine} \textit{CADepth-Net} \cite{yan2021channel}  \textit{VADepth} \cite{xiang2022visual} from recent work \cite{xiang2022visual}, which uses our validation set for the generalization evaluation from KITTI pretrained models.
For stereo geometry based unsupervised methods, \textit{Monodepth} method is evaluated using \url{https://github.com/OniroAI/MonoDepth-PyTorch}
, which is able to reproduce similar results to those in the paper on Eigen split.
We test the model of \textit{adareg} from \url{https://github.com/alexklwong/adareg-monodispnet} pre-trained with Eigen split. \textit{monoResMatch} is tested through \url{https://github.com/fabiotosi92/monoResMatch-Tensorflow} with \textit{KITTI} pretrined model with default hyperparameters. 
For sequence SfM based unsupervised methods, we adopt \url{https://github.com/ClementPinard/SfmLearner-Pytorch} to benchmark \textit{SfMLearner} for better performance than original repo with slight modification. \textcolor{revision}{We further fine-tune the pre-trained models of \texttt{dispnet\_model\_best} and \texttt{exp\_pose\_model\_best} on our training set  using default configuration file with sequence length of 5 for 20 epochs to get the best performance on $Average$ metric at epoch 20.} We use the model of \texttt{ResNet18} pre-trained on \textit{ImageNet} and fine-tuned on \textit{KITTI} with the resolution of $ 640 \times 192 $ to test \textit{PackNet} on \url{https://github.com/TRI-ML/packnet-sfm}. Similarly, in order to incorporate stereo geometric constraint into the monocular SfM framework, we use the  model of \texttt{mono+stereo} pre-trained on \textit{ImageNet} and \textit{KITTI} with the resolution of $ 640 \times 192 $ to evaluate the performance of \textit{Monodepth2} on \url{https://github.com/nianticlabs/monodepth2}.
For the multi-task SfM unsupervised learning methods, \textit{CC} is evaluated with \texttt{DispNet, PoseNet, MaskNet and FlowNet} pre-trained model on \textit{KITTI} through \url{https://github.com/anuragranj/cc}. We also test another work \textit{SGDepth} on \url{https://github.com/ifnspaml/SGDepth} with the \texttt{full} model of semantic segmentation and depth prediction with the resolution of $ 640 \times 192 $.

Since synthetic datasets like \textit{V-KITTI} include multiple environments in spite of existing domain gap, we additionally evaluate the performance of three domain adaptation methods from \textit{KITTI} benchmark, Atapour \textit{et al.} \cite{atapour2018real}, \textit{T2Net} \cite{zheng2018t2net} and \textit{GASDA} \cite{zhao2019geometry}. We follow the instruction on \url{https://github.com/atapour/monocularDepth-Inference} to evaluate the method proposed by Atapour \textit{et al.} with the model pre-trained on \texttt{KITTI} and \texttt{DeepGTAV} \cite{gtav}. \textit{T2Net} is tested on \url{https://github.com/lyndonzheng/Synthetic2Realistic} with the  weakly-supervised pre-trained model for outdoor scenes of \textit{KITTI} and \textit{V-KITTI}. We then evaluate the performance of \textit{GASDA} on \url{https://github.com/sshan-zhao/GASDA} with the model pre-trained on \textit{V-KITTI} and \textit{KITTI} using self-supervised stereo geometric information.

\subsection{Detailed Evaluation Results and Analysis}
\label{more_qualitative_results}
\subsubsection{Cross-Dataset Performance under Different Environments}
In this section, the detailed results with mean values and standard deviations across multiple environments are shown in Tab. \ref{absrel_env_results} and Tab. \ref{a1_env_results}, it can be seen that models with larger mean values tend to have more significant deviation for each environment. However, though there are some large standard deviations in Tab. \ref{absrel_env_results} and Tab. \ref{a1_env_results},
the quality of depth map ground truths is assured. So we attribute it to the poor generalization ability of those algorithms since not all the methods present such poor results with too large variances,  which cannot be correctly analyzed.

\begin{figure*}[]
	\centering
		\begin{minipage}[t]{0.12\linewidth}
			\centering
			\includegraphics[width=\linewidth]{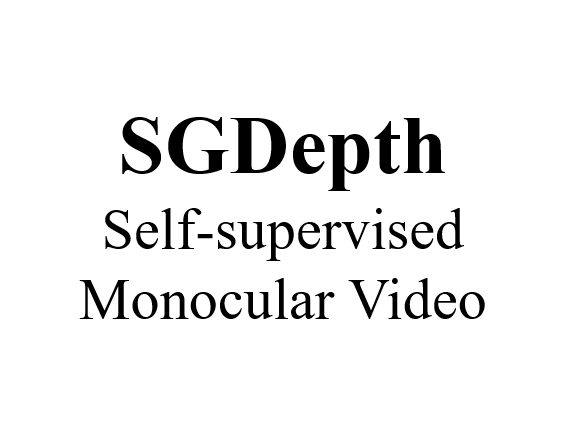}\\
			\includegraphics[width=\linewidth]{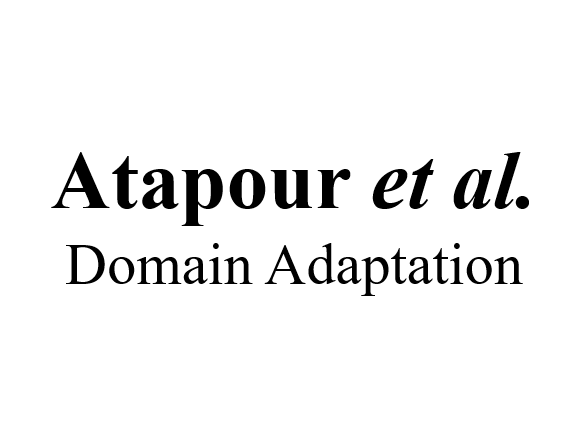}\\
			\includegraphics[width=\linewidth]{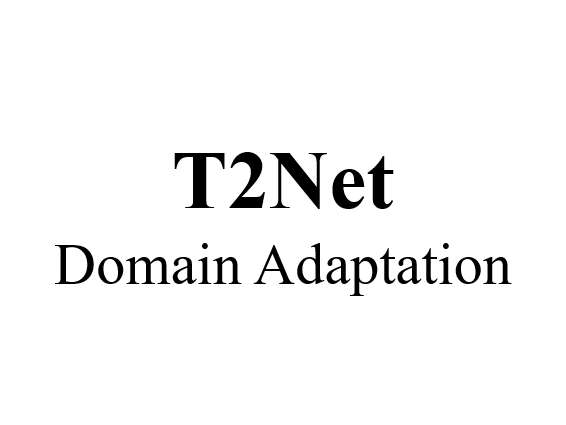}\\
			\includegraphics[width=\linewidth]{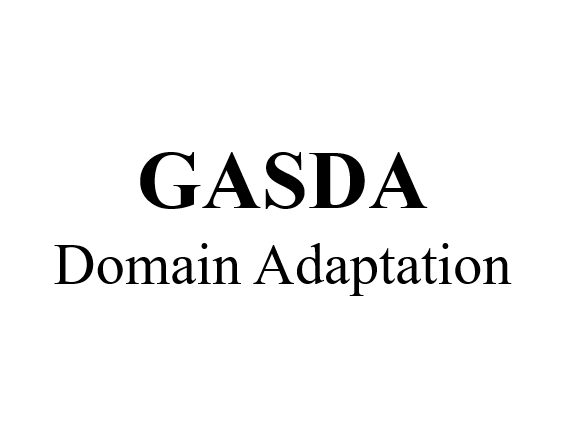}\\
		\end{minipage}%
		\begin{minipage}[t]{0.12\linewidth}
			\centering
			\includegraphics[width=\linewidth]{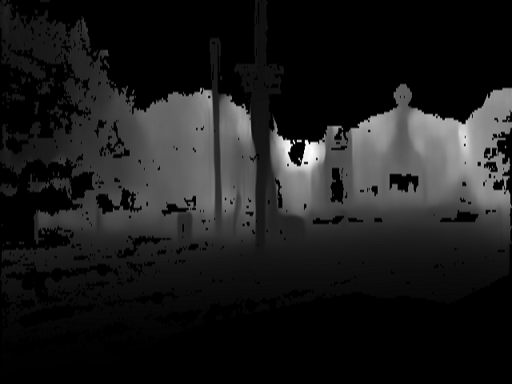}\\
			\includegraphics[width=\linewidth]{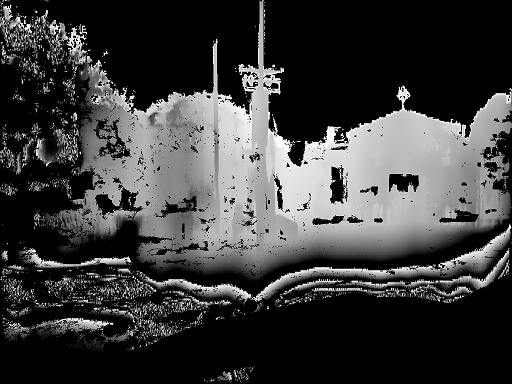}\\
			\includegraphics[width=\linewidth]{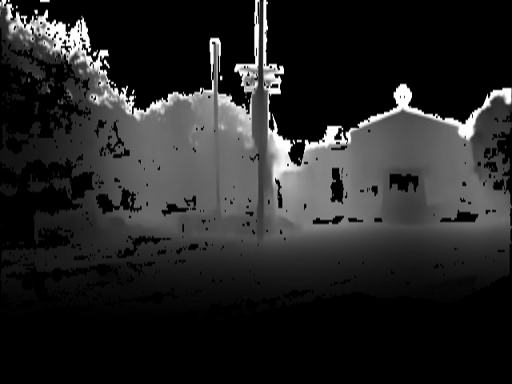}\\
			\includegraphics[width=\linewidth]{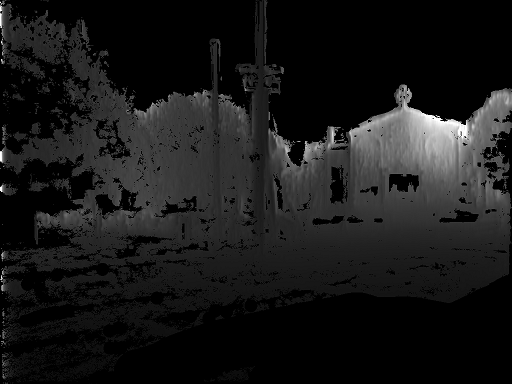}\\
			\begin{center}
			O+MF \\ Oct. 28th
			\end{center}
		\end{minipage}%
		\begin{minipage}[t]{0.12\linewidth}
			\centering
			\includegraphics[width=\linewidth]{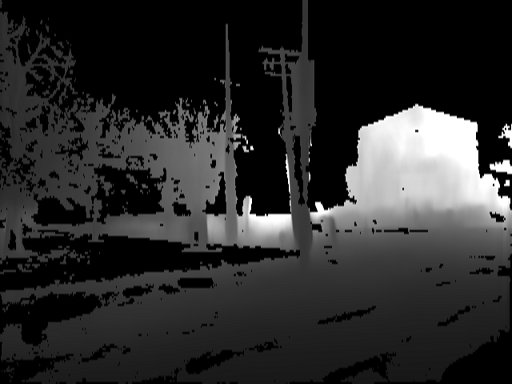}\\
			\includegraphics[width=\linewidth]{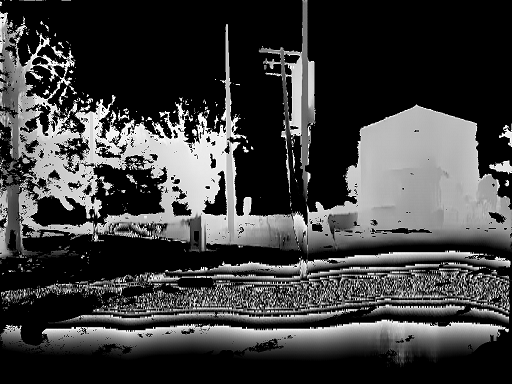}\\
			\includegraphics[width=\linewidth]{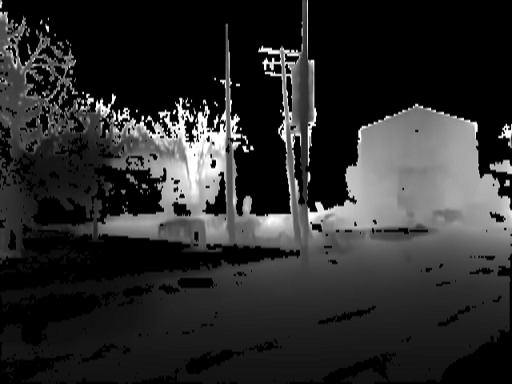}\\
			\includegraphics[width=\linewidth]{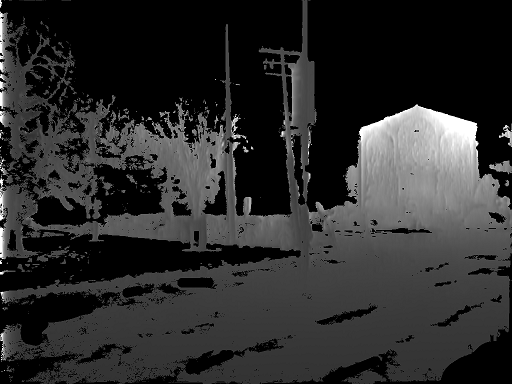}\\
			\begin{center}
			LS+NF+Sn \\ Dec. 21st
			\end{center}
		\end{minipage}%
		\begin{minipage}[t]{0.12\linewidth}
			\centering
			\includegraphics[width=\linewidth]{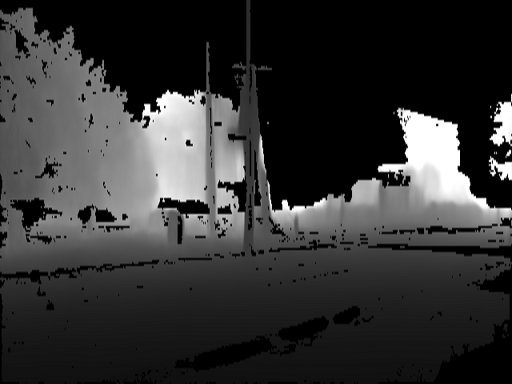}\\
			\includegraphics[width=\linewidth]{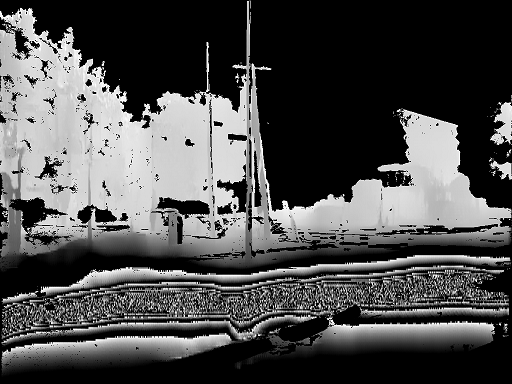}\\
			\includegraphics[width=\linewidth]{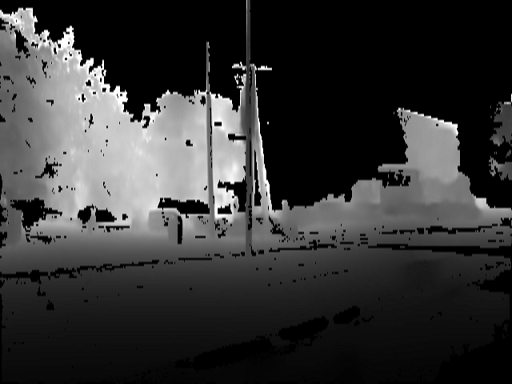}\\
			\includegraphics[width=\linewidth]{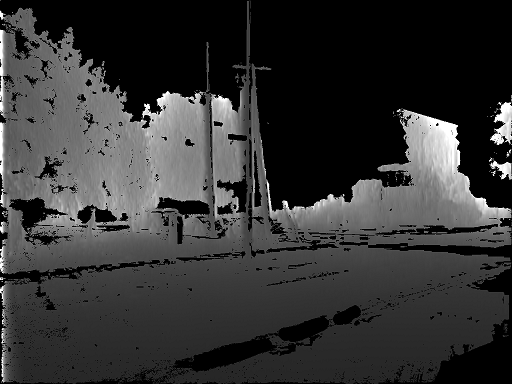}\\
			\begin{center}
			S+F \\ Sept. 15th
			\end{center}
		\end{minipage}%
		\begin{minipage}[t]{0.12\linewidth}
			\centering
			\includegraphics[width=\linewidth]{img/results2/adjusted/SGDepth/img_00811_c0_1303398537646371us.png}\\
			\includegraphics[width=\linewidth]{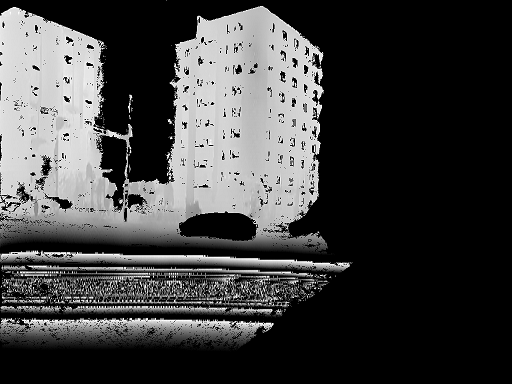}\\
			\includegraphics[width=\linewidth]{img/results2/adjusted/Synthetic2Realistic/img_00811_c0_1303398537646371us.png}\\
			\includegraphics[width=\linewidth]{img/results2/adjusted/GASDA/img_00811_c0_1303398537646371us.png}\\
			\begin{center}
			S+NF \\ Apr. 4th
			\end{center}
		\end{minipage}%
		\begin{minipage}[t]{0.12\linewidth}
			\centering
			\includegraphics[width=\linewidth]{img/results2/adjusted/SGDepth/img_01349_c0_1288792431498268us.png}\\
			\includegraphics[width=\linewidth]{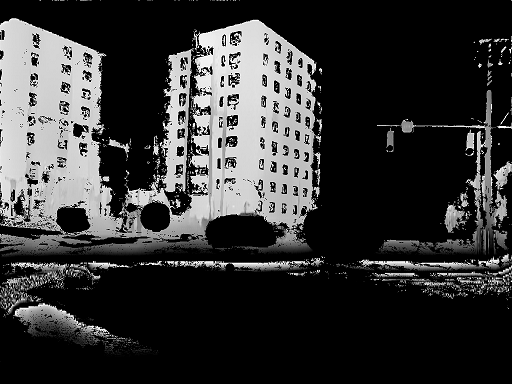}\\
			\includegraphics[width=\linewidth]{img/results2/adjusted/Synthetic2Realistic/img_01349_c0_1288792431498268us.png}\\
			\includegraphics[width=\linewidth]{img/results2/adjusted/GASDA/img_01349_c0_1288792431498268us.png}\\
			\begin{center}
			LS+MF \\ Nov. 3rd
			\end{center}
		\end{minipage}%
		\begin{minipage}[t]{0.12\linewidth}
			\centering
			\includegraphics[width=\linewidth]{img/results2/adjusted/SGDepth/img_01781_c0_1289589794783019us.png}\\
			\includegraphics[width=\linewidth]{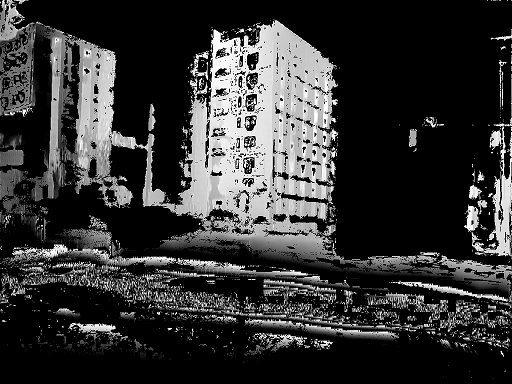}\\
			\includegraphics[width=\linewidth]{img/results2/adjusted/Synthetic2Realistic/img_01781_c0_1289589794783019us.png}\\
			\includegraphics[width=\linewidth]{img/results2/adjusted/GASDA/img_01781_c0_1289589794783019us.png}\\
			\begin{center}
			LS+MF \\ Nov. 12th
			\end{center}
		\end{minipage}%
	\caption{Qualitative results for all the baselines with multiple illuminations, vegetation and weather conditions.  
	}
	\label{more_results}
\end{figure*}

Moreover, all the evaluated baselines are visualized after adjustment under typical challenging environments, including dark illumination, snowy scene, and complex vegetation. See Fig. \ref{more_results} for more details. From the results of supervised methods, it can be seen that the patterns of predicted depth maps are similar, especially for \textit{BTS} \cite{lee2019big} and \textit{VNL} \cite{yin2019enforcing}, where the top and bottom areas are dark while the middle areas are bright due to overfitting, see buildings as examples. But \textit{VNL} \cite{yin2019enforcing} shows the advantage in depth details (\textit{e.g.} telephone poles and vegetation) in the middle areas which accounts for the best average performance. 

Stereo training involved self-supervised methods (including \textit{Monodepth2} \cite{godard2019digging} and \textit{GASDA} \cite{zhao2019geometry}) perform best continuous depth results for the same entity under all environments, \textit{e.g.} depth values of buildings. 
Monocular video-based self-supervised methods do better in distinguishing relative depth from far and near areas, \textit{e.g.} depth values for objects along different directions of roads, especially for multi-task learning ones \textit{CC} \cite{ranjan2019competitive} and \textit{SGDepth} \cite{klingner2020self}. Besides, domain adaptation methods still suffer from domain gaps, which shows that synthetic multi-environment images help little to improve performance under real-world changing environments.

\begin{figure*}[]
	\centering
		\begin{minipage}[t]{0.12\linewidth}
			\centering
			\includegraphics[width=\linewidth]{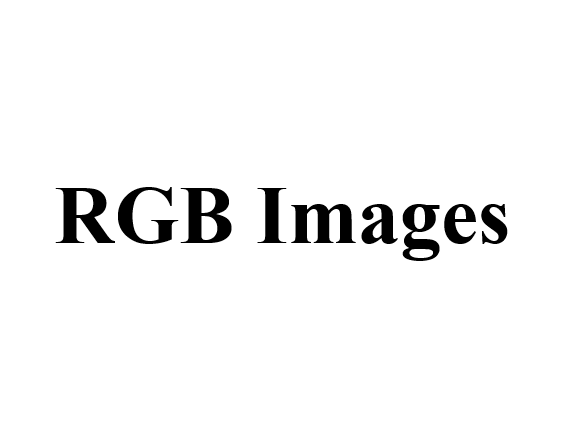}\\
			\includegraphics[width=\linewidth]{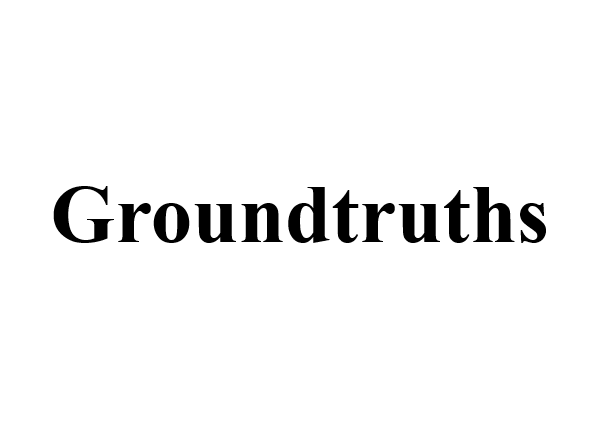}\\
			\includegraphics[width=\linewidth]{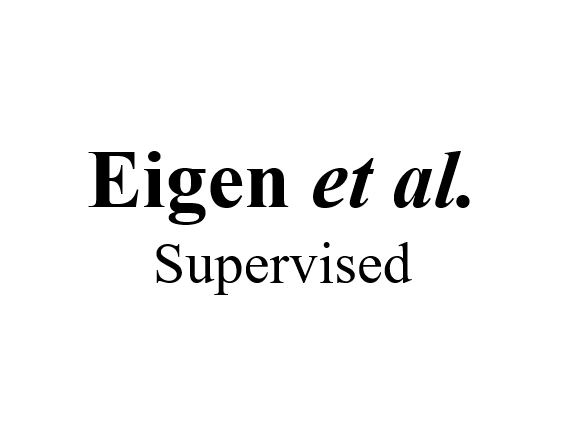}\\
			\includegraphics[width=\linewidth]{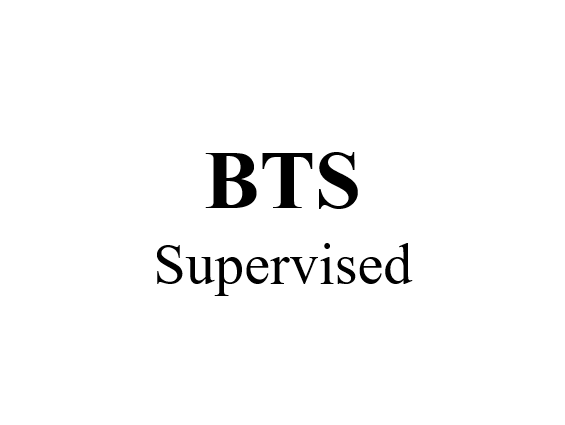}\\
			\includegraphics[width=\linewidth]{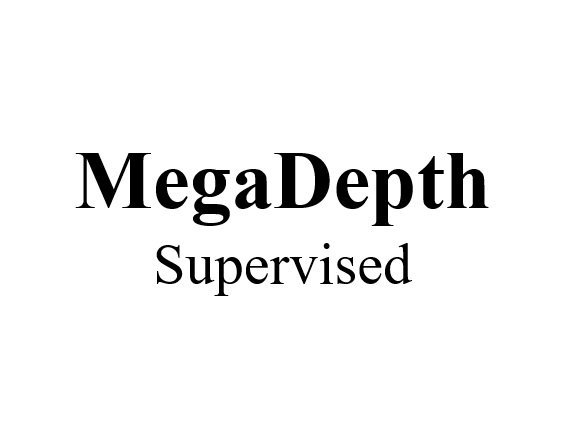}\\
			\includegraphics[width=\linewidth]{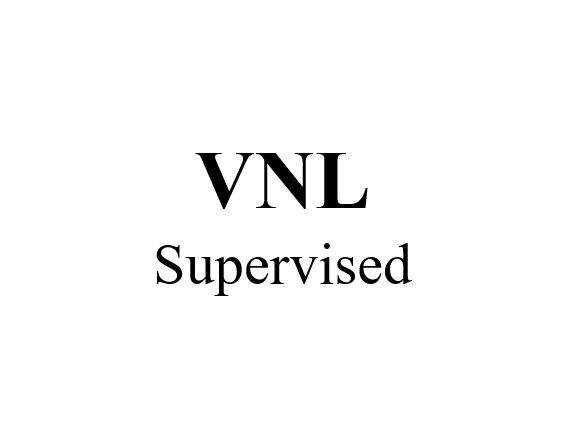}\\
			\includegraphics[width=\linewidth]{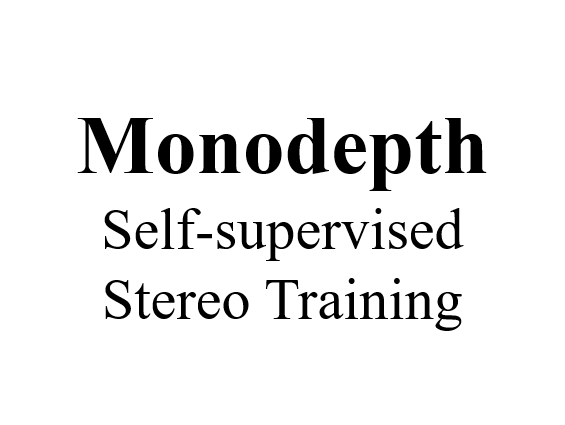}\\
			\includegraphics[width=\linewidth]{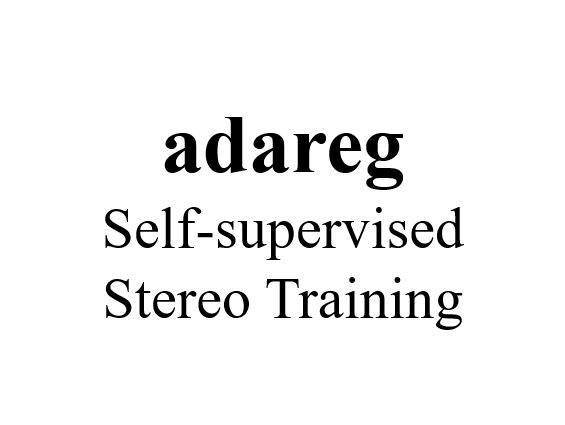}\\
			\includegraphics[width=\linewidth]{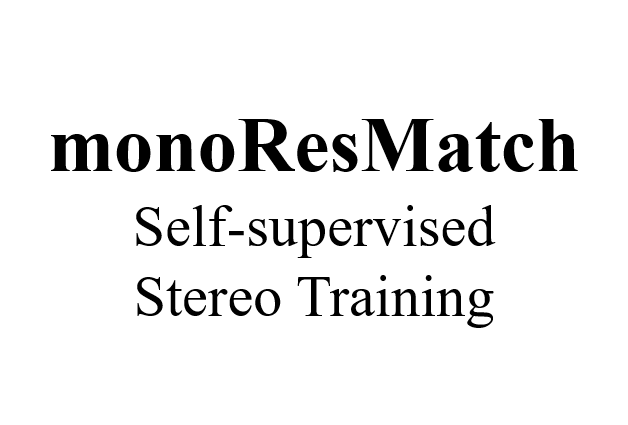}\\
			\includegraphics[width=\linewidth]{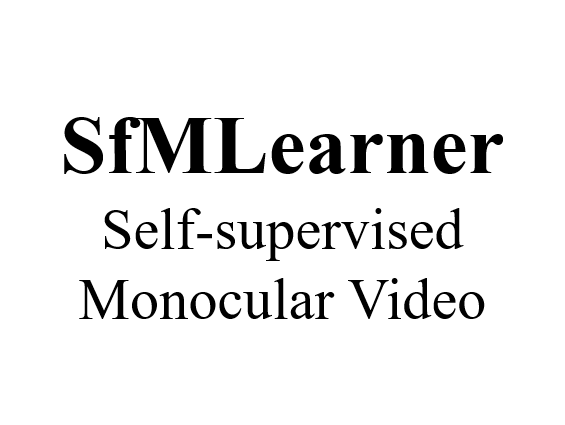}\\
			\includegraphics[width=\linewidth]{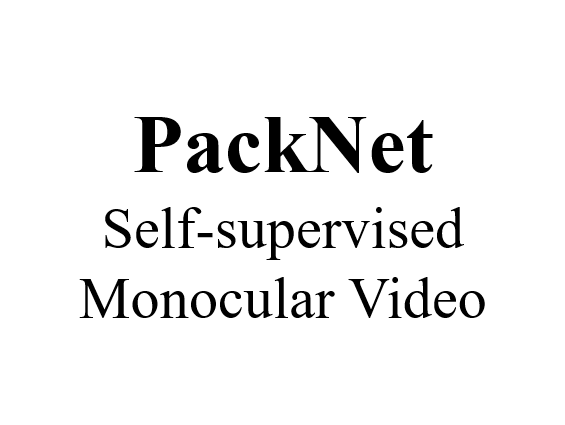}\\
			\includegraphics[width=\linewidth]{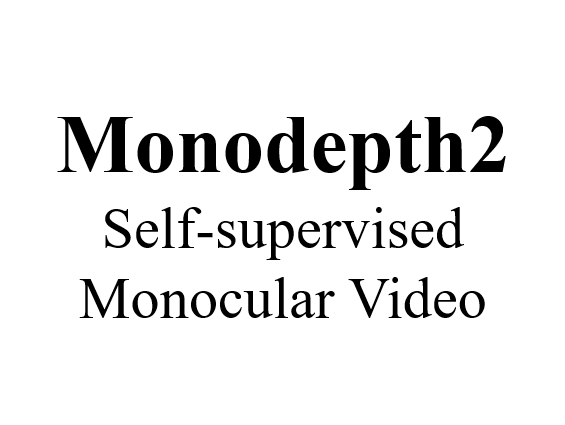}\\
			\includegraphics[width=\linewidth]{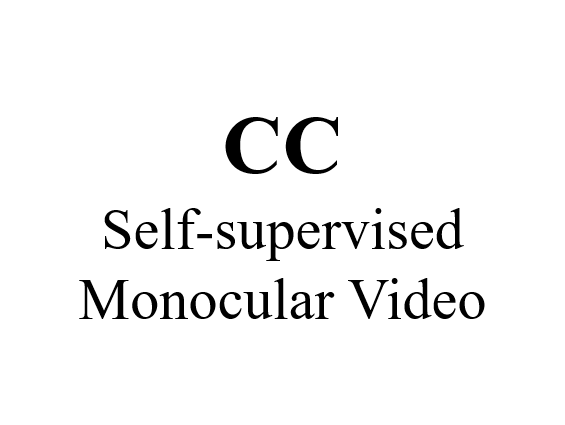}\\
		\end{minipage}%
		\begin{minipage}[t]{0.12\linewidth}
			\centering
			\includegraphics[width=\linewidth]{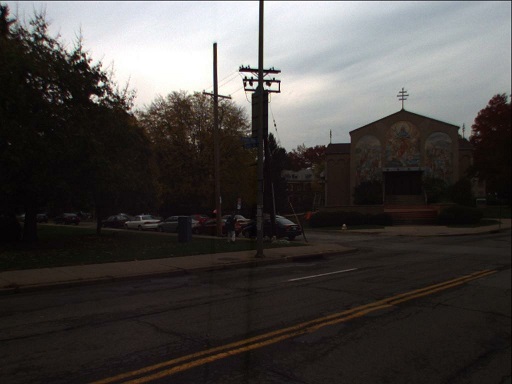}\\
			\includegraphics[width=\linewidth]{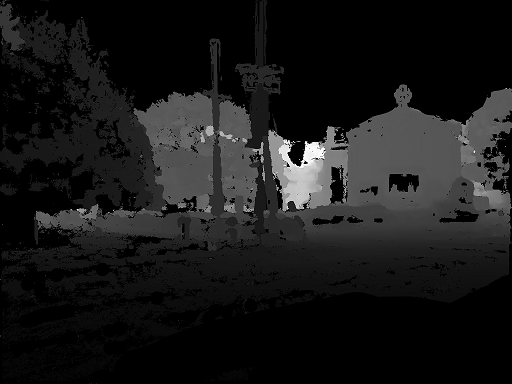}\\
			\includegraphics[width=\linewidth]{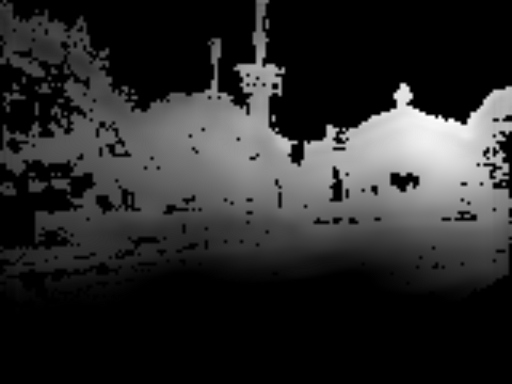}\\
			\includegraphics[width=\linewidth]{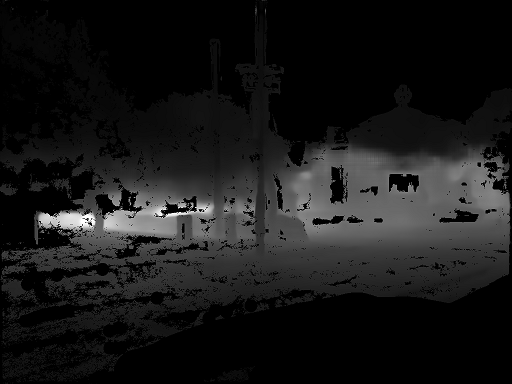}\\
			\includegraphics[width=\linewidth]{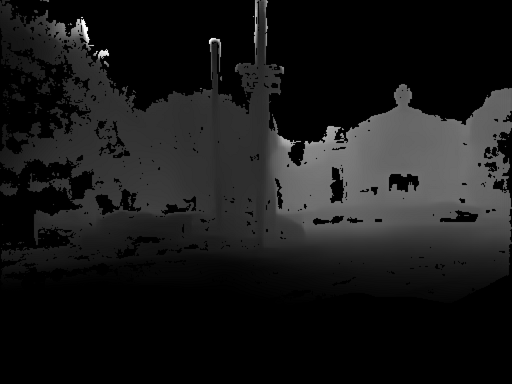}\\
			\includegraphics[width=\linewidth]{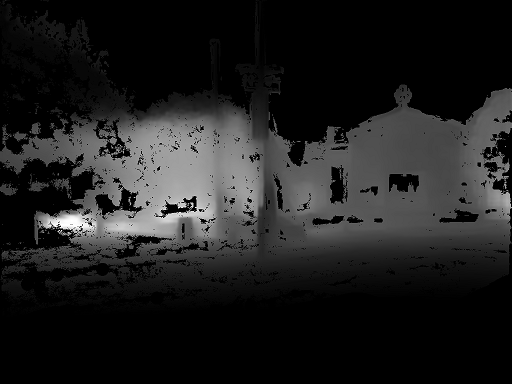}\\
			\includegraphics[width=\linewidth]{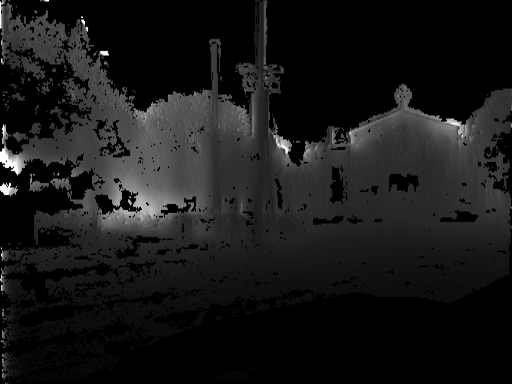}\\
			\includegraphics[width=\linewidth]{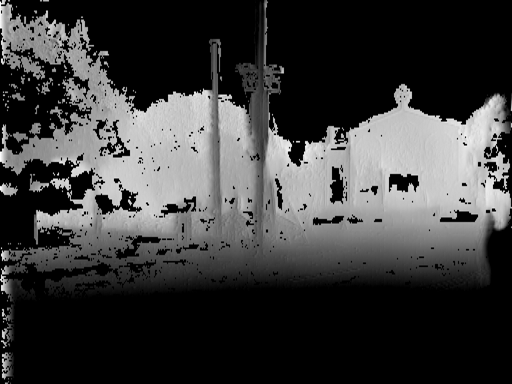}\\
			\includegraphics[width=\linewidth]{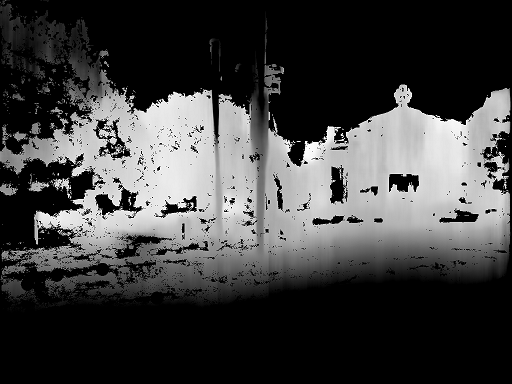}\\
			\includegraphics[width=\linewidth]{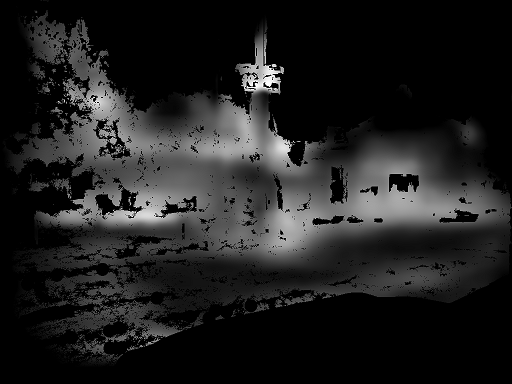}\\
			\includegraphics[width=\linewidth]{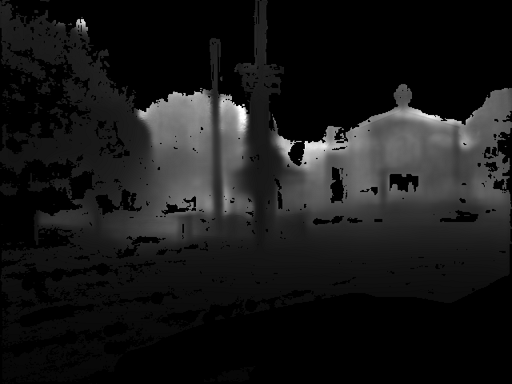}\\
			\includegraphics[width=\linewidth]{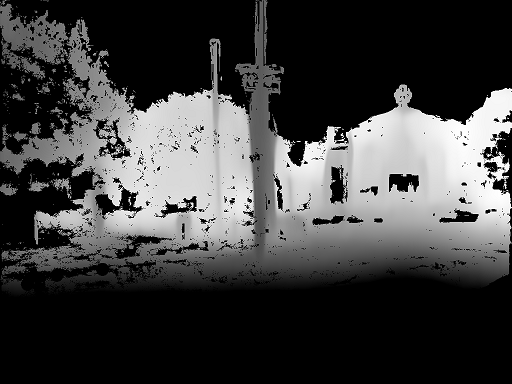}\\
			\includegraphics[width=\linewidth]{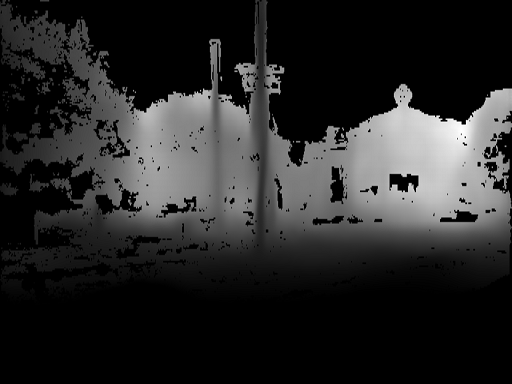}\\
			\begin{center}
			O+MF \\ Oct. 28th
			\end{center}
		\end{minipage}%
		\begin{minipage}[t]{0.12\linewidth}
			\centering
			\includegraphics[width=\linewidth]{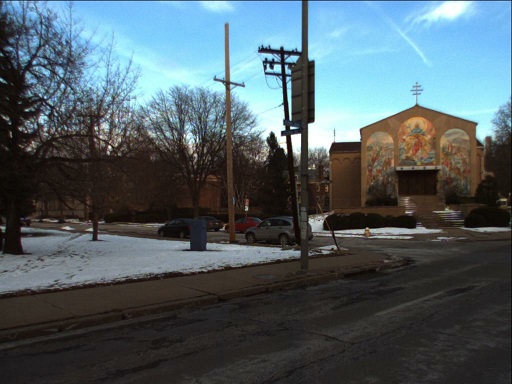}\\
			\includegraphics[width=\linewidth]{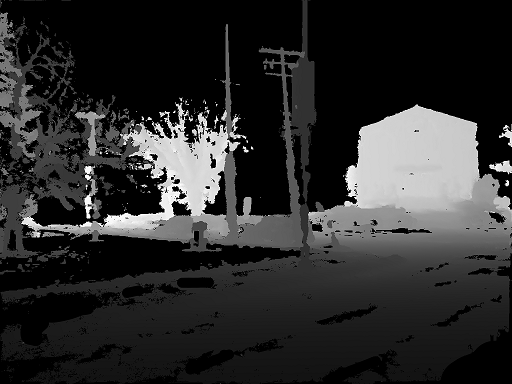}\\
			\includegraphics[width=\linewidth]{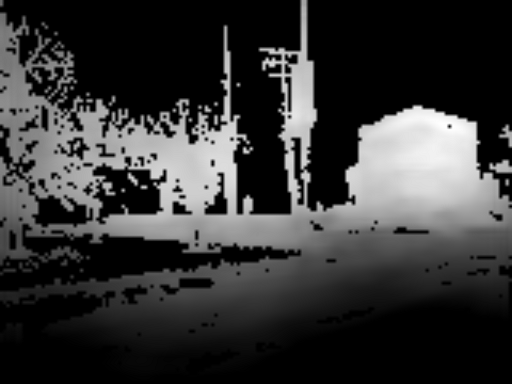}\\
			\includegraphics[width=\linewidth]{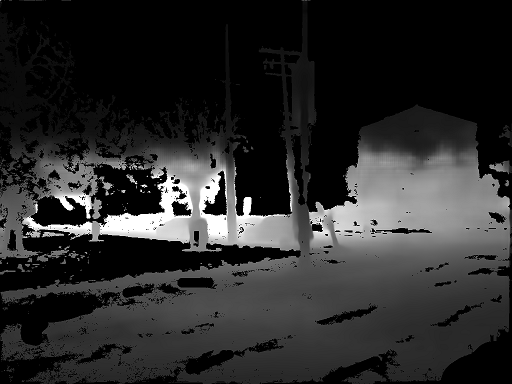}\\
			\includegraphics[width=\linewidth]{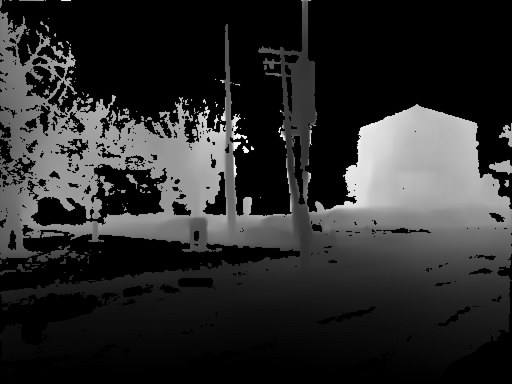}\\
			\includegraphics[width=\linewidth]{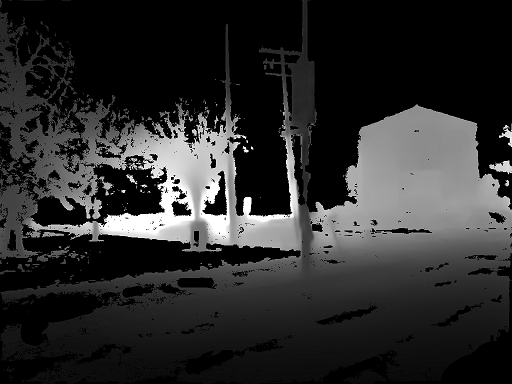}\\
			\includegraphics[width=\linewidth]{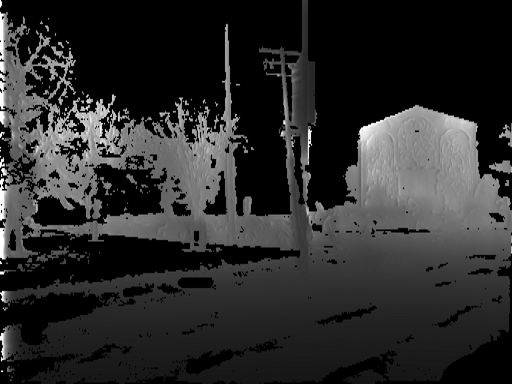}\\
			\includegraphics[width=\linewidth]{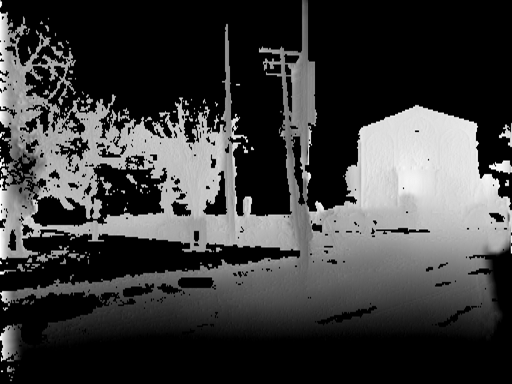}\\
			\includegraphics[width=\linewidth]{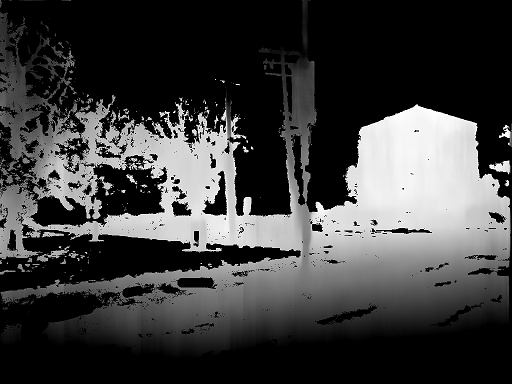}\\
			\includegraphics[width=\linewidth]{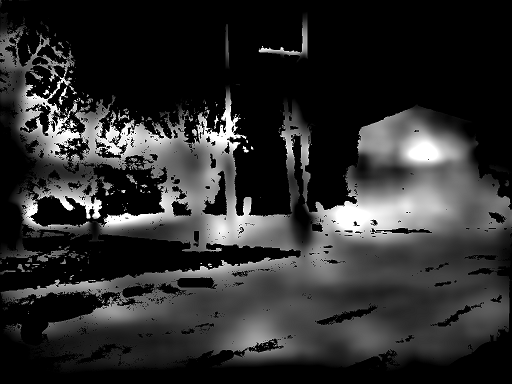}\\
			\includegraphics[width=\linewidth]{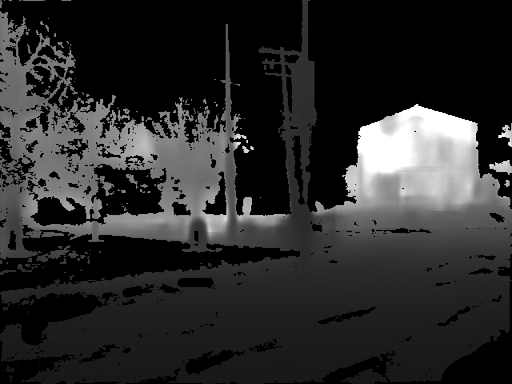}\\
			\includegraphics[width=\linewidth]{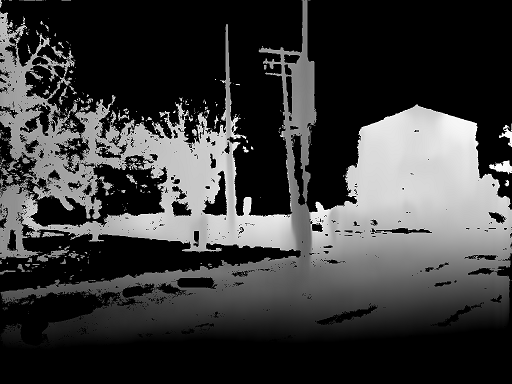}\\
			\includegraphics[width=\linewidth]{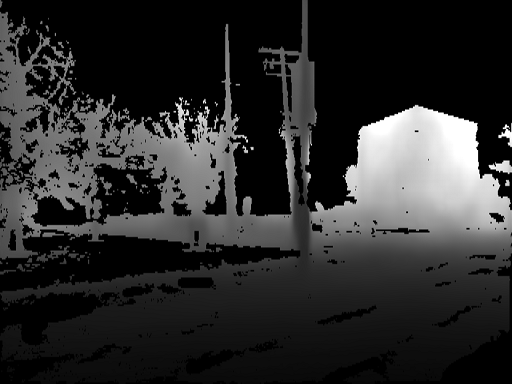}\\
			\begin{center}
			LS+NF+Sn \\ Dec. 21st
			\end{center}
		\end{minipage}%
		\begin{minipage}[t]{0.12\linewidth}
			\centering
			\includegraphics[width=\linewidth]{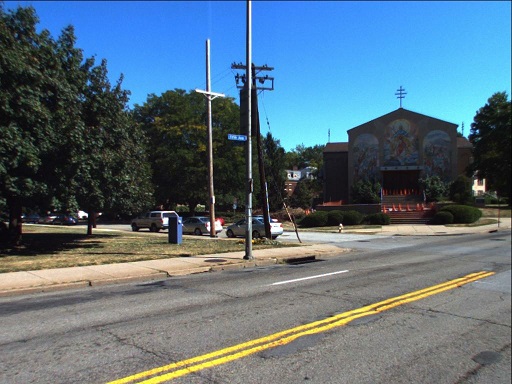}\\
			\includegraphics[width=\linewidth]{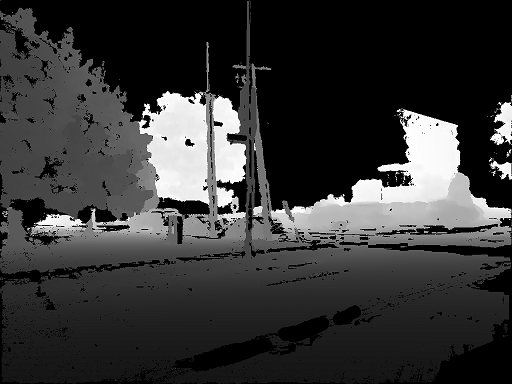}\\
			\includegraphics[width=\linewidth]{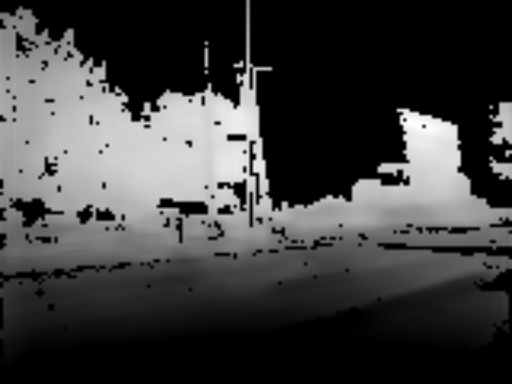}\\
			\includegraphics[width=\linewidth]{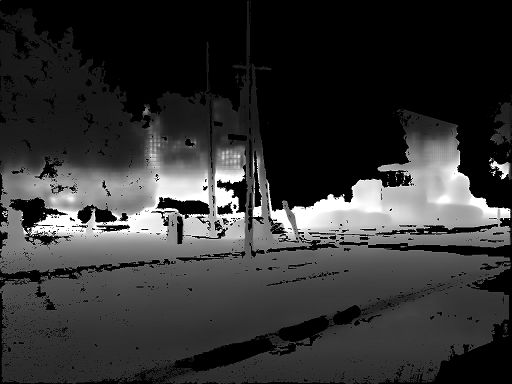}\\
			\includegraphics[width=\linewidth]{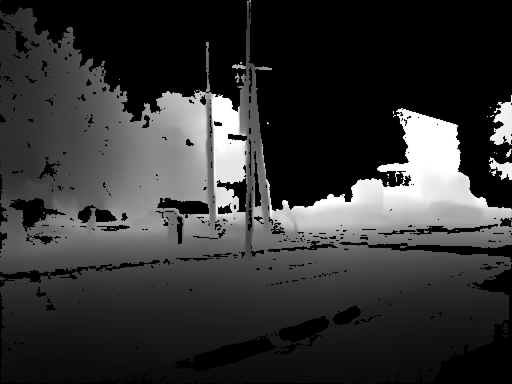}\\
			\includegraphics[width=\linewidth]{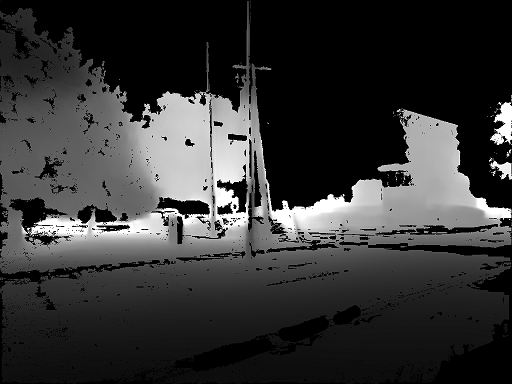}\\
			\includegraphics[width=\linewidth]{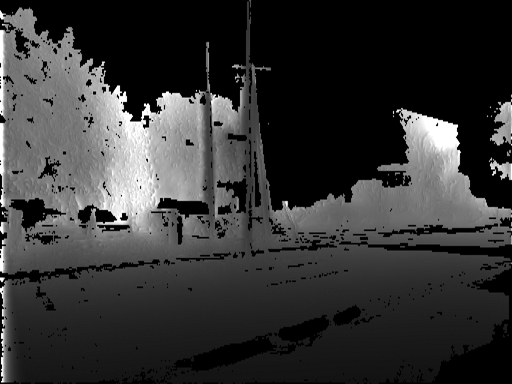}\\
			\includegraphics[width=\linewidth]{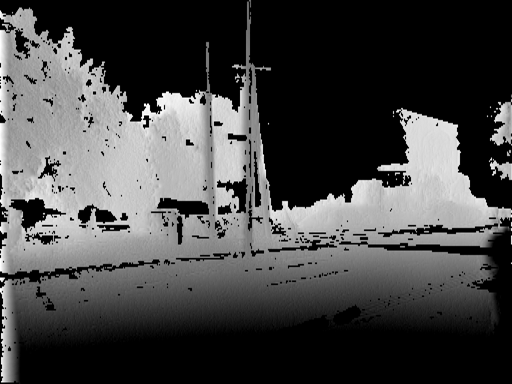}\\
			\includegraphics[width=\linewidth]{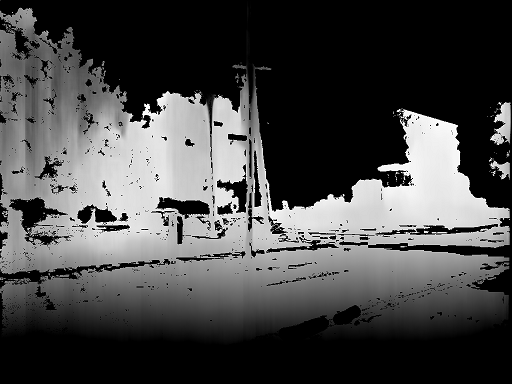}\\
			\includegraphics[width=\linewidth]{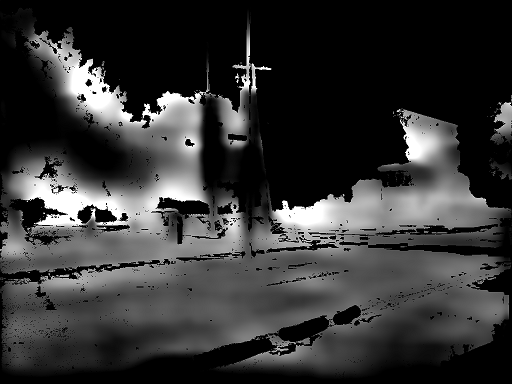}\\
			\includegraphics[width=\linewidth]{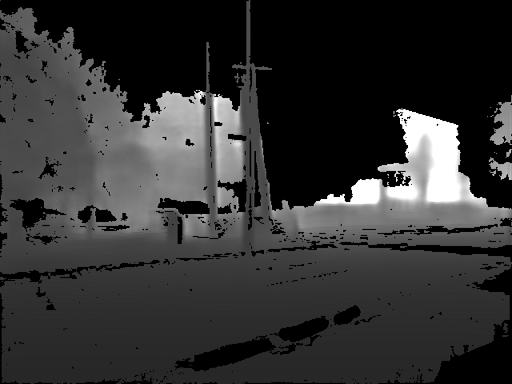}\\
			\includegraphics[width=\linewidth]{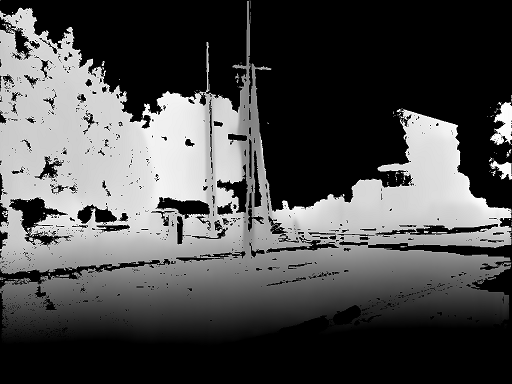}\\
			\includegraphics[width=\linewidth]{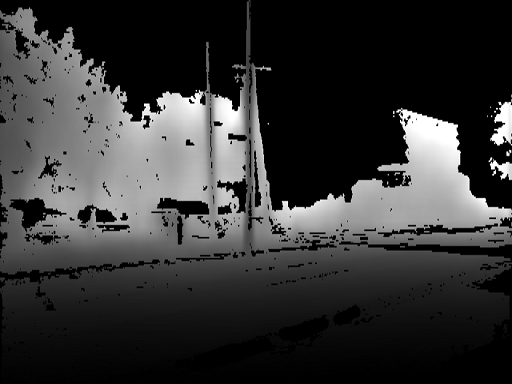}\\
			\begin{center}
			S+F \\ Sept. 15th
			\end{center}
		\end{minipage}%
		\begin{minipage}[t]{0.12\linewidth}
			\centering
			\includegraphics[width=\linewidth]{img/results2/rgb/img_00811_c0_1303398537646371us.jpg}\\
			\includegraphics[width=\linewidth]{img/results2/ground_truth/img_00811_c0_1303398537646371us.png}\\
			\includegraphics[width=\linewidth]{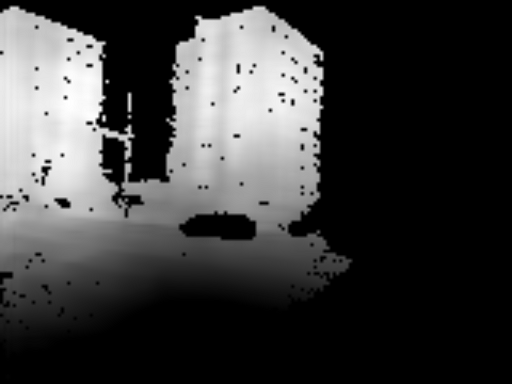}\\
			\includegraphics[width=\linewidth]{img/results2/adjusted/bts/img_00811_c0_1303398537646371us.png}\\
			\includegraphics[width=\linewidth]{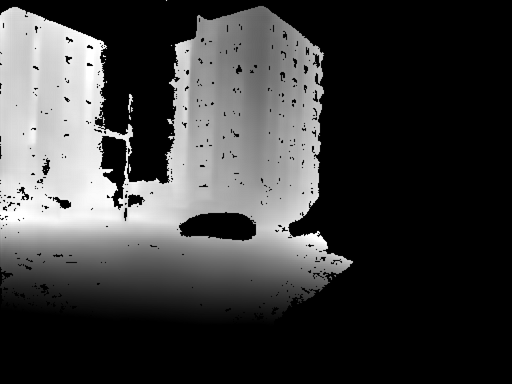}\\
			\includegraphics[width=\linewidth]{img/results2/adjusted/VNL/img_00811_c0_1303398537646371us.png}\\
			\includegraphics[width=\linewidth]{img/results2/adjusted/MonoDepth-Pytorch/img_00811_c0_1303398537646371us.png}\\
			\includegraphics[width=\linewidth]{img/results2/adjusted/adareg-monodispnet/img_00811_c0_1303398537646371us.png}\\
			\includegraphics[width=\linewidth]{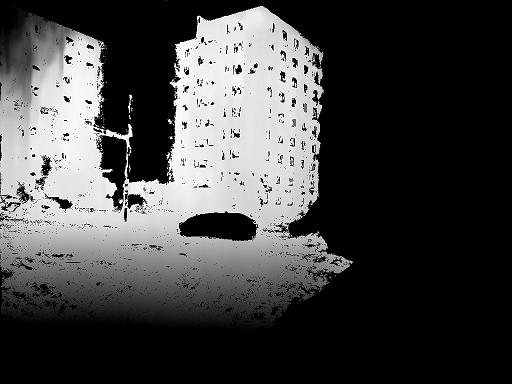}\\
			\includegraphics[width=\linewidth]{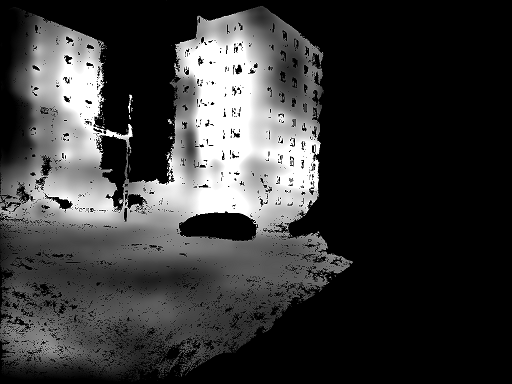}\\
			\includegraphics[width=\linewidth]{img/results2/adjusted/packnet-sfm/img_00811_c0_1303398537646371us.png}\\
			\includegraphics[width=\linewidth]{img/results2/adjusted/monodepth2/img_00811_c0_1303398537646371us.png}\\
			\includegraphics[width=\linewidth]{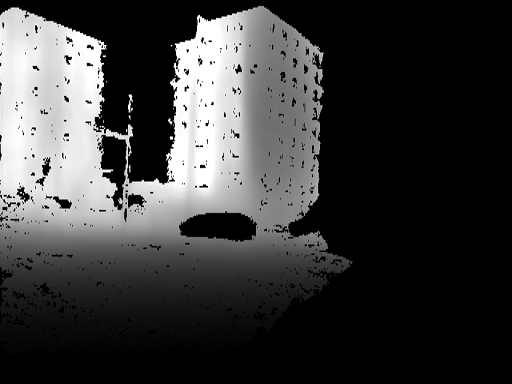}\\
			\begin{center}
			S+NF \\ Apr. 4th
			\end{center}
		\end{minipage}%
		\begin{minipage}[t]{0.12\linewidth}
			\centering
			\includegraphics[width=\linewidth]{img/results2/rgb/img_01349_c0_1288792431498268us.jpg}\\
			\includegraphics[width=\linewidth]{img/results2/ground_truth/img_01349_c0_1288792431498268us.png}\\
			\includegraphics[width=\linewidth]{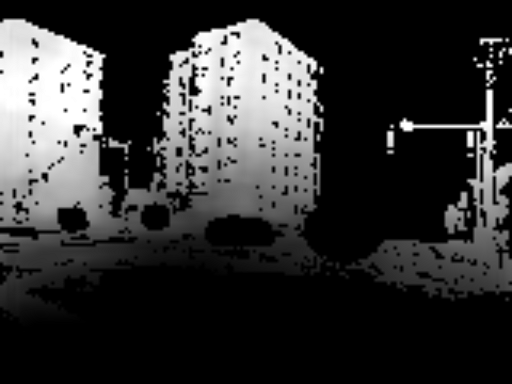}\\
			\includegraphics[width=\linewidth]{img/results2/adjusted/bts/img_01349_c0_1288792431498268us.png}\\
			\includegraphics[width=\linewidth]{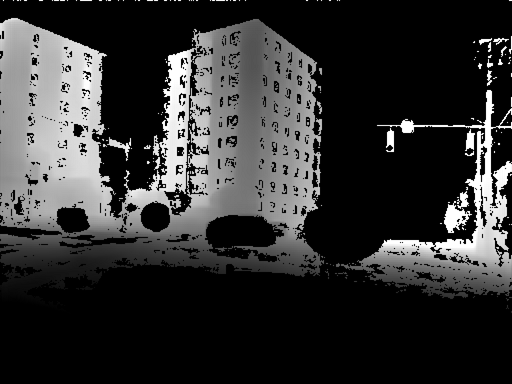}\\
			\includegraphics[width=\linewidth]{img/results2/adjusted/VNL/img_01349_c0_1288792431498268us.png}\\
			\includegraphics[width=\linewidth]{img/results2/adjusted/MonoDepth-Pytorch/img_01349_c0_1288792431498268us.png}\\
			\includegraphics[width=\linewidth]{img/results2/adjusted/adareg-monodispnet/img_01349_c0_1288792431498268us.png}\\
			\includegraphics[width=\linewidth]{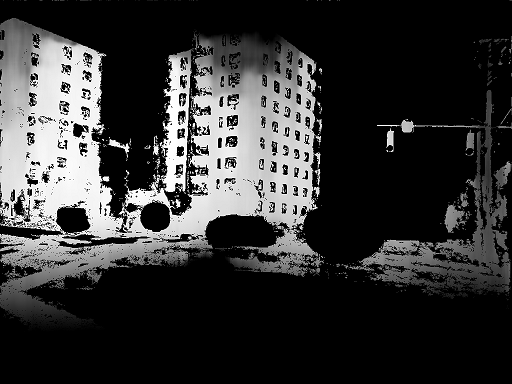}\\
			\includegraphics[width=\linewidth]{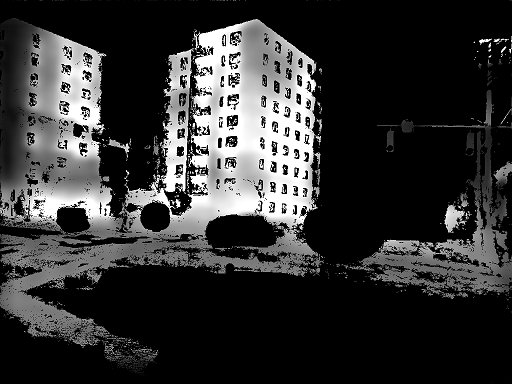}\\
			\includegraphics[width=\linewidth]{img/results2/adjusted/packnet-sfm/img_01349_c0_1288792431498268us.png}\\
			\includegraphics[width=\linewidth]{img/results2/adjusted/monodepth2/img_01349_c0_1288792431498268us.png}\\
			\includegraphics[width=\linewidth]{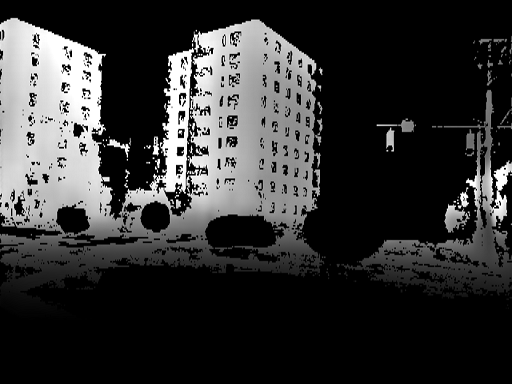}\\
			\begin{center}
			LS+MF \\ Nov. 3rd
			\end{center}
		\end{minipage}%
		\begin{minipage}[t]{0.12\linewidth}
			\centering
			\includegraphics[width=\linewidth]{img/results2/rgb/img_01781_c0_1289589794783019us.jpg}\\
			\includegraphics[width=\linewidth]{img/results2/ground_truth/img_01781_c0_1289589794783019us.png}\\
			\includegraphics[width=\linewidth]{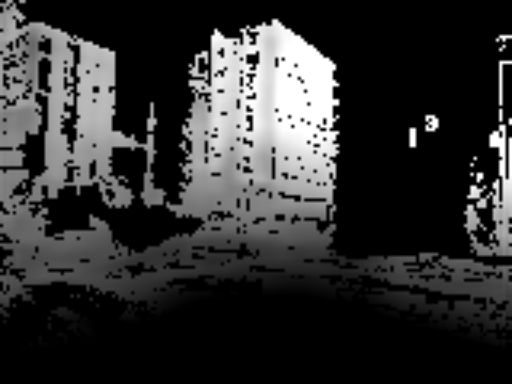}\\
			\includegraphics[width=\linewidth]{img/results2/adjusted/bts/img_01781_c0_1289589794783019us.png}\\
			\includegraphics[width=\linewidth]{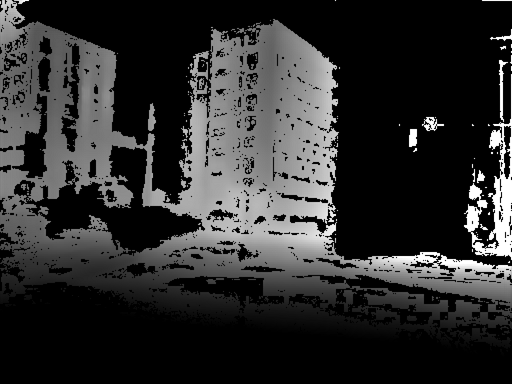}\\
			\includegraphics[width=\linewidth]{img/results2/adjusted/VNL/img_01781_c0_1289589794783019us.png}\\
			\includegraphics[width=\linewidth]{img/results2/adjusted/MonoDepth-Pytorch/img_01781_c0_1289589794783019us.png}\\
			\includegraphics[width=\linewidth]{img/results2/adjusted/adareg-monodispnet/img_01781_c0_1289589794783019us.png}\\
			\includegraphics[width=\linewidth]{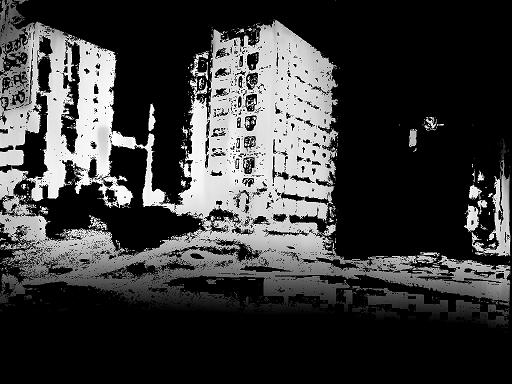}\\
			\includegraphics[width=\linewidth]{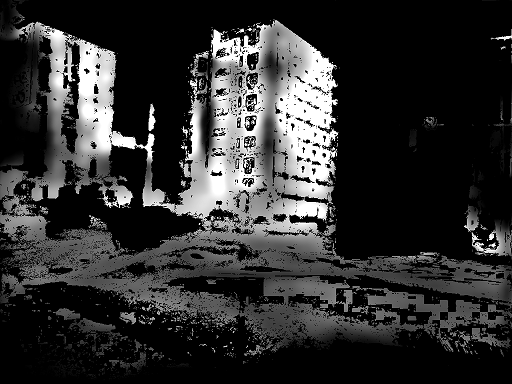}\\
			\includegraphics[width=\linewidth]{img/results2/adjusted/packnet-sfm/img_01781_c0_1289589794783019us.png}\\
			\includegraphics[width=\linewidth]{img/results2/adjusted/monodepth2/img_01781_c0_1289589794783019us.png}\\
			\includegraphics[width=\linewidth]{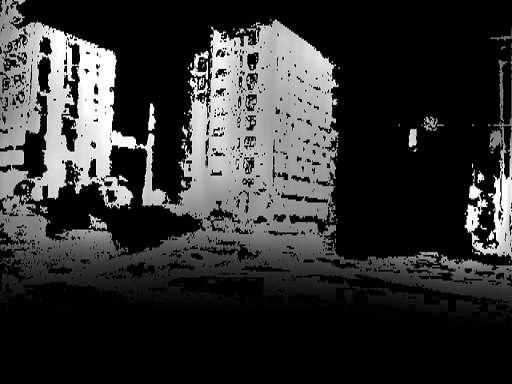}\\
			\begin{center}
			LS+MF \\ Nov. 12th
			\end{center}
			\rightline{ \footnotesize To be continued}
		\end{minipage}%
\end{figure*}

\begin{figure*}[]
	\centering
		\begin{minipage}[t]{0.32\linewidth}
			\centering
			\includegraphics[width=\linewidth]{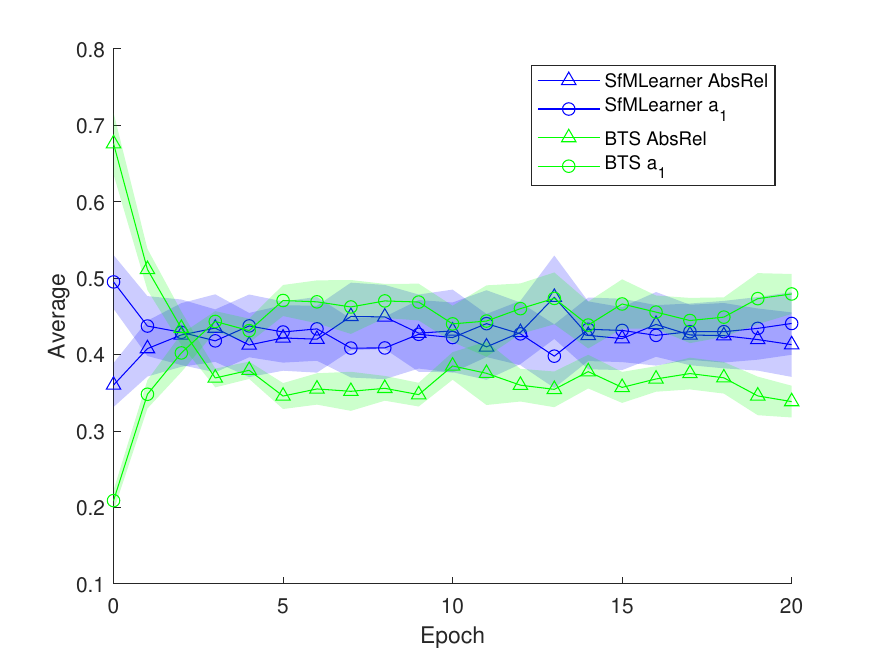}
		\end{minipage}%
		\begin{minipage}[t]{0.32\linewidth}
			\centering
			\includegraphics[width=\linewidth]{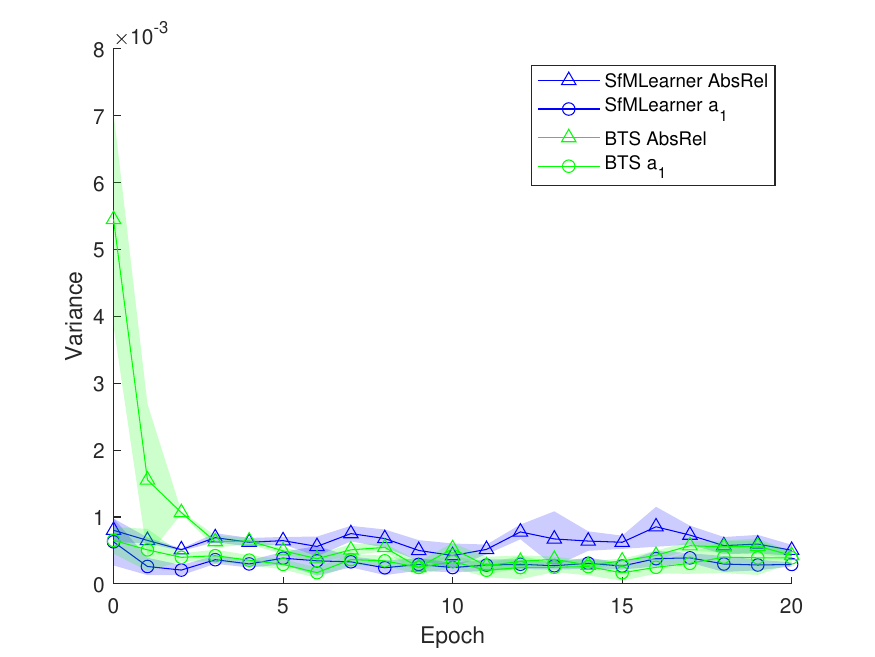}
		\end{minipage}%
		\begin{minipage}[t]{0.32\linewidth}
			\centering
			\includegraphics[width=\linewidth]{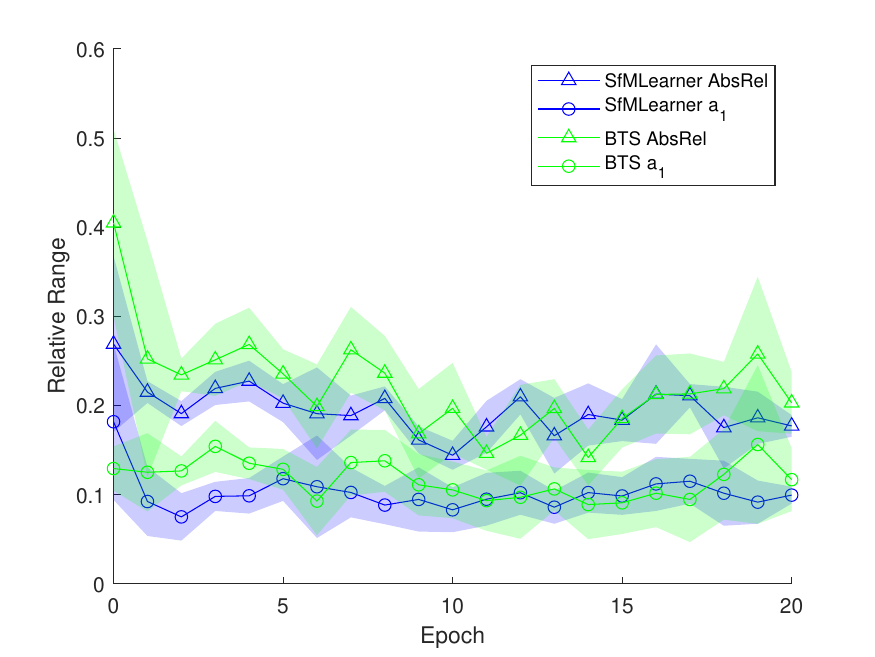}
		\end{minipage}%
	\centering
	\caption{Performance evolution after fine-tuning on \textit{SeasonDepth} training set from \textit{KITTI} pretrained models.  } 
	\label{fine-tuning_results}
\end{figure*}

\subsubsection{Analysis of Performance of Fine-tuned Models}
As the comparison of zero-shot cross-dataset evaluation from KITTI to SeasonDepth on the validation set, we use our  training set to fine-tune one supervised  \cite{lee2019big} and one self-supervised model \cite{zhou2017unsupervised} from KITTI, which initially perform poor zero-shot results, to alleviate the impact of dataset bias between \textit{KITTI} and \textit{SeasonDepth}.  Since our dataset does not contain stereo images, segmentation ground truth, and KITTI-like scenarios, just like in V-KITTI, the stereo training based, semantic segmentation involved multi-task training and domain adaptation models are omitted for the sake of fairness. 

Besides results in Tab. \ref{finetune_evalation_results}, we present the evaluation results along training epochs with shadows of cross-slice standard deviation after zooming 0.5, 0.2, and 0.5 times in Fig. \ref{fine-tuning_results}. It can be seen that after the fine-tuning, overall performance is improved while some $Variance$ and $Relative Range$ results still perform badly, especially for SfMLearner \cite{zhou2017unsupervised}, showing that the naive fine-tuning with more data will not help too much compared to other zero-shot evaluations in Tab. \ref{evalation_results}.


\subsection{Cross-dataset Comparison with Cityscapes}
\label{comparison_cityscape}
\begin{figure*}[]
	\centering
		\begin{minipage}[t]{0.24\linewidth}
			\centering
			\includegraphics[width=\linewidth]{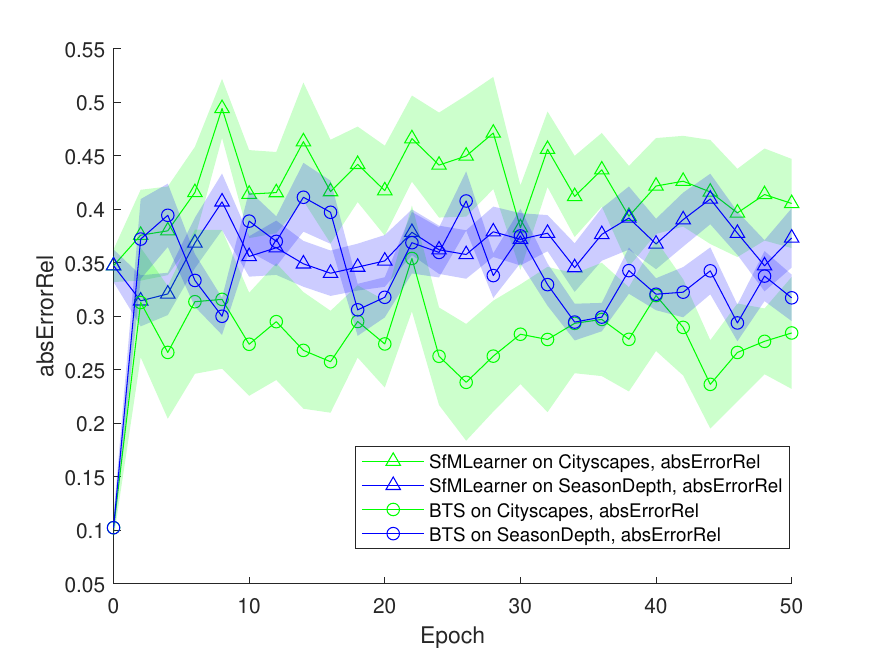}
		\end{minipage}%
		\begin{minipage}[t]{0.24\linewidth}
			\centering
			\includegraphics[width=\linewidth]{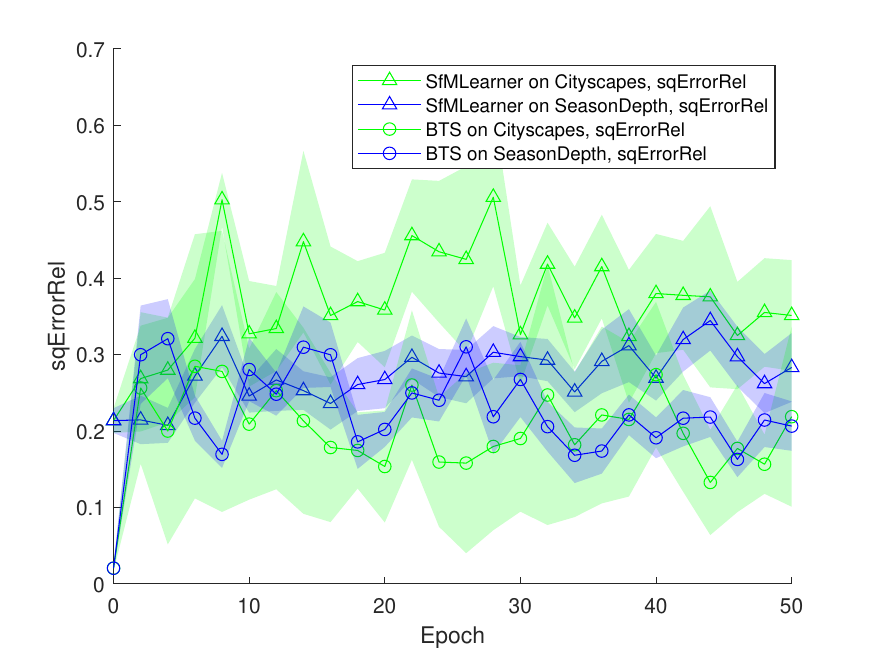}
		\end{minipage}%
		\begin{minipage}[t]{0.24\linewidth}
			\centering
			\includegraphics[width=\linewidth]{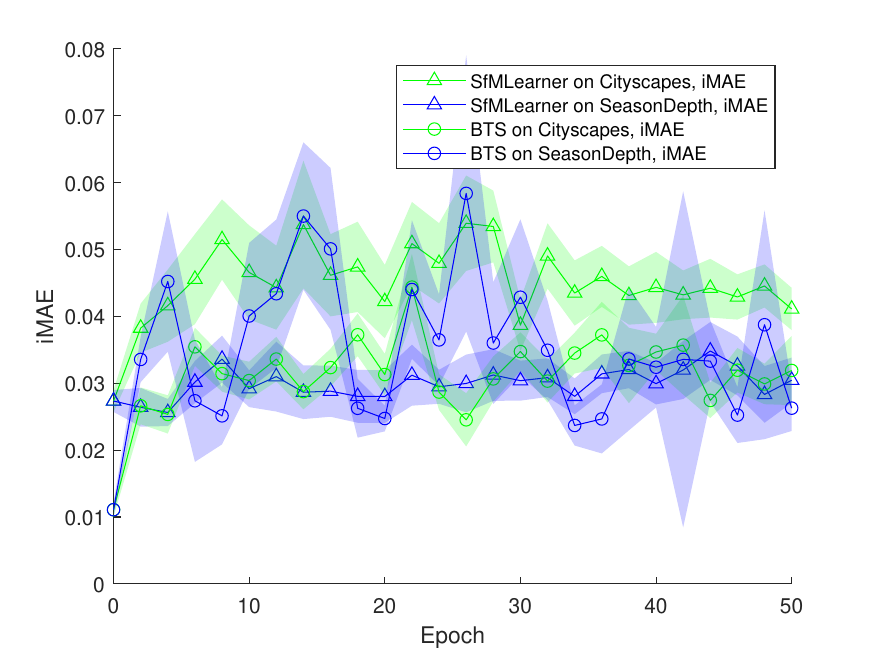}
		\end{minipage}%
		\begin{minipage}[t]{0.24\linewidth}
			\centering
			\includegraphics[width=\linewidth]{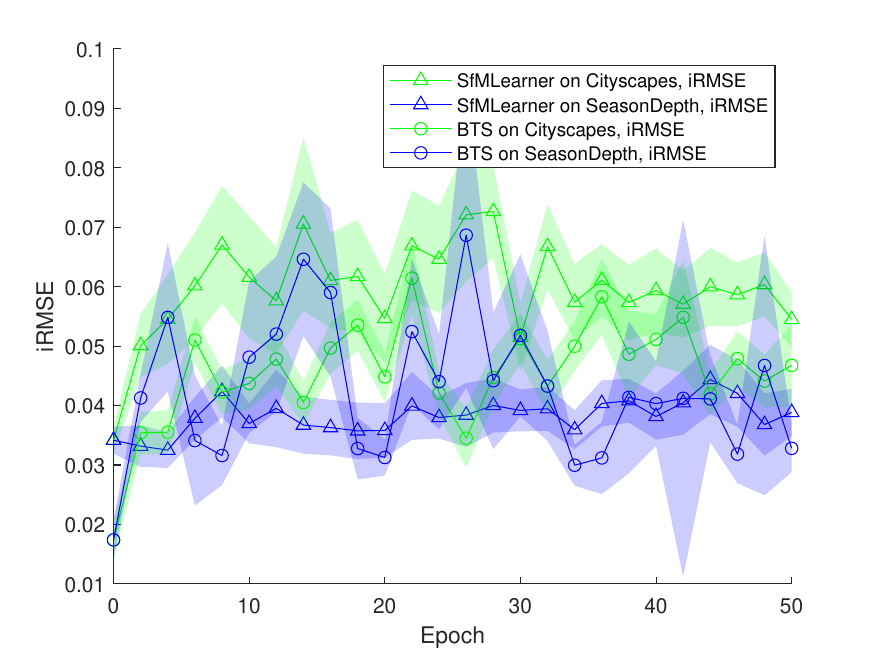}
		\end{minipage}%
	\centering
	\caption{Cross-dataset performance evolution on KITTI validation set \cite{Uhrig2017THREEDV} with models  fine-tuned on \textit{SeasonDepth} and \textit{Cityscapes} \cite{cordts2016cityscapes}. } 
	\label{justification_quan}
\end{figure*}

In this section, we present more details about the cross-dataset comparison experiment with \textit{Cityscapes} to justify our depth quality for model training. As it is introduced in Sec. \ref{benchmark_details}, we choose the KITTI pre-trained models for \textit{BTS} and \textit{SfMLearner} methods, and fine-tune them on our training set and \textit{Cityscapes} \cite{cordts2016cityscapes} for 50 epochs, respectively. Finally, we evaluate the cross-dataset transfer performance on the KITTI validation set \cite{Uhrig2017THREEDV} using \texttt{development kit} from \url{http://www.cvlibs.net/datasets/kitti/eval\_depth.php?benchmark=depth\_prediction}. We choose 11407 images from \texttt{train\_extra} in \textit{Cityscapes} \cite{cordts2016cityscapes} to fine-tune the models, which is exactly the same amount of images in our training set to make the comparison fair. 

To fine-tune the self-supervised model of \textit{SfMLearner}, we set  \texttt{batch\_size} to be 4,  \texttt{epoch\_size} to be 1000 and \texttt{sequence\_length} to be 1000. Along with the instructions to train with own data \url{https://github.com/ClementPinard/SfmLearner-Pytorch/issues/108}, we crop a quarter of the bottom in the image and resize it to be $416\times128$ to remove the car logo in \textit{Cityscapes} dataset. We change the intrinsic parameters accordingly to make them consistent with cropped images. For a fair comparison, we also conduct such cropping for the images from the \textit{SeasonDepth} dataset. When testing the KITTI validation set, we resize the images to $416\times128$ before feeding them into the networks.

When fine-tuning the supervised \textit{BTS} model, we set  \texttt{batch\_size} to be 16,  \texttt{input\_size} to be $256\times192$ for \textit{SeasonDepth} images and $256\times128$ for \textit{Cityscapes} images. For depth ground truth, we directly adopt the depth maps in   \textit{Cityscapes} as supervision signals while for \textit{the SeasonDepth} dataset, we only consider the non-zero pixels and conduct  alignment using mean value  to the ground truth to construct loss when fine-tuning. The experimental results show that such alignment to construct supervised loss is effective using our dataset for supervised model training.

Besides the results in Tab. \ref{justification_quan2}, we present the  \textit{KITTI} performance evolution over epochs with models fine-tuned on \textit{SeasonDepth} and \textit{Cityscapes} in Fig. \ref{justification_quan}. We can see that although the performance will be degraded compared to the KITTI pre-trained models due to domain shift when fine-tuning, the performance fine-tuned on \textit{SeasonDepth} is better than models fine-tuned on \textit{Cityscapes}, especially for \textit{SfMLearner} method and $iMAE$ and $iRMSE$ metrics. Besides, the fluctuation of models  fine-tuned on \textit{SeasonDepth} is much less than those  fine-tuned on \textit{Cityscapes} in terms of $absErrorRel$ and $sqErrorRel$ metrics. 
 Consequently, although the depth maps of \textit{SeasonDepth} are reconstructed from structure from motion and do not contain dynamic objects,  the ground truth accuracy is eligible  to be used for model training compared to the stereo depth dataset \textit{Cityscapes}, justifying our ground truth accuracy is adequate though it is  not perfect.

\section{Limitation and Discussion}
\label{limitation}

In this section, we discuss the limitation of our work. As mentioned before, our \textit{SeasonDepth} dataset is built based on the CMU Visual Localization dataset, \textcolor{revision}{which was initially collected for visual localization and contained multiple scenes but without challenging night scenes. Although it is different from the dataset for autonomous driving like \textit{KITTI}, which causes concern about the evaluation due to the domain gap.  However, based on the experimental evidence, it is acceptable that fine-tuned models only provide limited help in terms of $Variance$ and $Relative Range$.
Although dynamic objects are not included in the dataset to ensure accuracy and reliability, it does not affect the evaluation for real driving applications because it cannot be distinguished whether the objects are dynamic or static given a single monocular image when testing. And the cross-dataset justification experiment also shows that missing dynamic objects do not influence the model training too much. Consequently, the evaluation of the depth prediction of static objects can reveal the performance of dynamic objects, although they are not involved in the ground truth. }

Besides, though normalizing the scale of evaluation metrics through alignment of mean and variance can also be done through quantile alignment shown in Sec \ref{stat_dataset}, it is more sensitive to noise to adopt quantile-based alignment of every single image for evaluation.
Although we try our best to survey and test the open-source representative models as much as possible, it is impossible to involve all the monocular depth prediction methods in our benchmark. So we  release the training, validation and test set, and benchmark toolkit to make up for it. 
\textcolor{revision}{Another limitation is that it is not straightforward to train models on the dataset because of the ground truths of scaleless relative values, but it can be trained after the mean value alignment to the ground truth just as the fine-tuned BTS does. It can also reflect how environmental changes affect depth prediction models and give hints of what kind of method is more promising to this problem. } 

\section{Discussion on Societal Impacts}
\label{impacts}
To our best knowledge, we are the first work focusing on changing environments on monocular depth prediction tasks, which has great significance to long-term or lifelong autonomous driving and outdoor mobile robotics. The robustness of the depth prediction algorithm is important to the safety of vehicles and pedestrians from the long-run perspective. 

However, there are also some potential negative societal impacts. First, our dataset is not that general because the original dataset CMU Visual Localization dataset is only collected in one city, which may mislead the algorithm to overfit on similar scenes, leading to instability and risks when used in complex scenes for applications. Second, privacy is another problem. Although the dataset is secondarily derived and there are many licenses on it, malicious and unintended uses may still happen, \textit{e.g.} collect the human faces or properties of the locals, which may violate the privacy right and cause other problems.

Dismissing such concerns needs the efforts of research, industry, and other social organizations. For example, researchers and engineers should thoroughly evaluate the performance and robustness of algorithms with environmental changes despite using our dataset to ensure the safety of autonomous driving. Social organizations should also keep an eye on such open-source real-world datasets to avoid illegal use.

\end{document}